\documentclass{article}

\PassOptionsToPackage{numbers, compress}{natbib}
\usepackage{wrapfig}
\usepackage{tikz}
\usepackage[final]{neurips_2025}

\usepackage[utf8]{inputenc} % allow utf-8 input
\usepackage[T1]{fontenc}    % use 8-bit T1 fonts
\usepackage{amssymb}    
\usepackage{bm}
\usepackage{makecell}
\usepackage{derivative}
\usepackage{graphicx}
\usepackage{hyperref}       % hyperlinks
\usepackage{url}            % simple URL typesetting
\usepackage{booktabs}       % professional-quality tables
\usepackage{amsfonts}       % blackboard math symbols
\usepackage{nicefrac}       % compact symbols for 1/2, etc.
\usepackage{paralist}
\usepackage{enumitem}
\usepackage{microtype}      % microtypography
\usepackage{xcolor}         % colors
\usepackage{amsmath}
\usepackage{xcolor}
\usepackage{verbatim}
\usepackage{comment}
\usepackage{multirow}
\usepackage{arydshln}
\usepackage{lipsum}
\usepackage{cancel}
\usepackage[most]{tcolorbox}
\usepackage[ruled, noend, vlined]{algorithm2e}

\definecolor{appleblue}{RGB}{84, 151, 193}
\definecolor{appleorange}{RGB}{250, 151, 92}
\definecolor{applegreen}{RGB}{83, 172, 121}
\definecolor{applered}{RGB}{227, 94, 105}
\definecolor{applepurple}{RGB}{161, 150, 204}

\definecolor{google_blue}{RGB}{66, 133, 244}
\definecolor{google_green}{RGB}{52, 168, 83}
\definecolor{google_red}{RGB}{234, 67, 53}
\definecolor{google_orange}{RGB}{251, 188, 5}

\tcbset{
  questionbox/.style={
    enhanced,
    colback=gray!5,    
    colframe=white,    % no frame visible (same color as background)
    boxrule=0pt,       % no border line
    arc=2mm,           % small rounded corners
    left=2mm, right=2mm, top=1mm, bottom=1mm, % internal padding
    sharp corners,
    fonttitle=\bfseries,
    before skip=10pt, after skip=10pt, % vertical space
  }
}

\DeclareMathOperator*{\argmin}{arg\,min}
\def\v#1{\mathbf{#1}}

\newcommand{\logebm}[1]{\ensuremath{\textcolor{google_blue}{\mathbf{ #1_{E_{\theta}}}}}}
\newcommand{\invebm}[1]{\ensuremath{\textcolor{google_green}{\mathbf{ #1_{1/p_{\theta}}}}}}
\newcommand{\rbf}[1]{\ensuremath{\textcolor{google_red}{\mathbf{ #1_{RBF}}}}}
\newcommand{\landm}[1]{\ensuremath{\textcolor{google_orange}{\mathbf{ #1_{LAND}}}}}

\newcommand{\rev}[1]{\textcolor{black}{#1}}

\title{Follow the Energy, Find the Path: Riemannian Metrics from Energy-Based Models}

\author{%
  Louis Bethune\thanks{Equal contribution.} \\
  Apple \\
  \And
  David Vigouroux \\
  IRT Saint Exupéry, ANITI, IMT Atlantique \\
  \AND
  Yilun Du \\
  Harvard University \\
  \And
  Rufin VanRullen \\
  CNRS \\
  \And
  Thomas Serre \\
  Brown University \\
  \And
  Victor Boutin\footnotemark[1] \\
  CNRS \\
}

\begin{document}

\maketitle

\begin{abstract}
% Context
What is the shortest path between two data points lying in a high-dimensional space? While the answer is trivial in Euclidean geometry, it becomes significantly more complex when the data lies on a curved manifold—requiring a Riemannian metric to describe the space's local curvature.
% The challenge
Estimating such a metric, however, remains a major challenge in high dimensions.

% Our work
In this work, we propose a method for deriving Riemannian metrics directly from pretrained Energy-Based Models (EBMs)—a class of generative models that assign low energy to high-density regions.
These metrics define spatially varying distances, enabling the computation of geodesics—shortest paths that follow the data manifold’s intrinsic geometry.
%plan
We introduce two novel metrics derived from EBMs and show that they produce geodesics that remain closer to the data manifold and exhibit lower curvature distortion, as measured by alignment with ground-truth trajectories.
We evaluate our approach on increasingly complex datasets: synthetic datasets with known data density, rotated character images with interpretable geometry, and high-resolution natural images embedded in a pretrained VAE latent space.
Our results show that EBM-derived metrics consistently outperform established baselines, especially in high-dimensional settings. 

% opening 
Our work is the first to derive Riemannian metrics from EBMs, enabling data-aware geodesics and unlocking scalable, geometry-driven learning for generative modeling and simulation.
\end{abstract}

\section{Introduction}

\begin{wrapfigure}{r}{0.5\textwidth}
  \centering
  \vspace{-5mm}
  \includegraphics[width=0.5\textwidth]{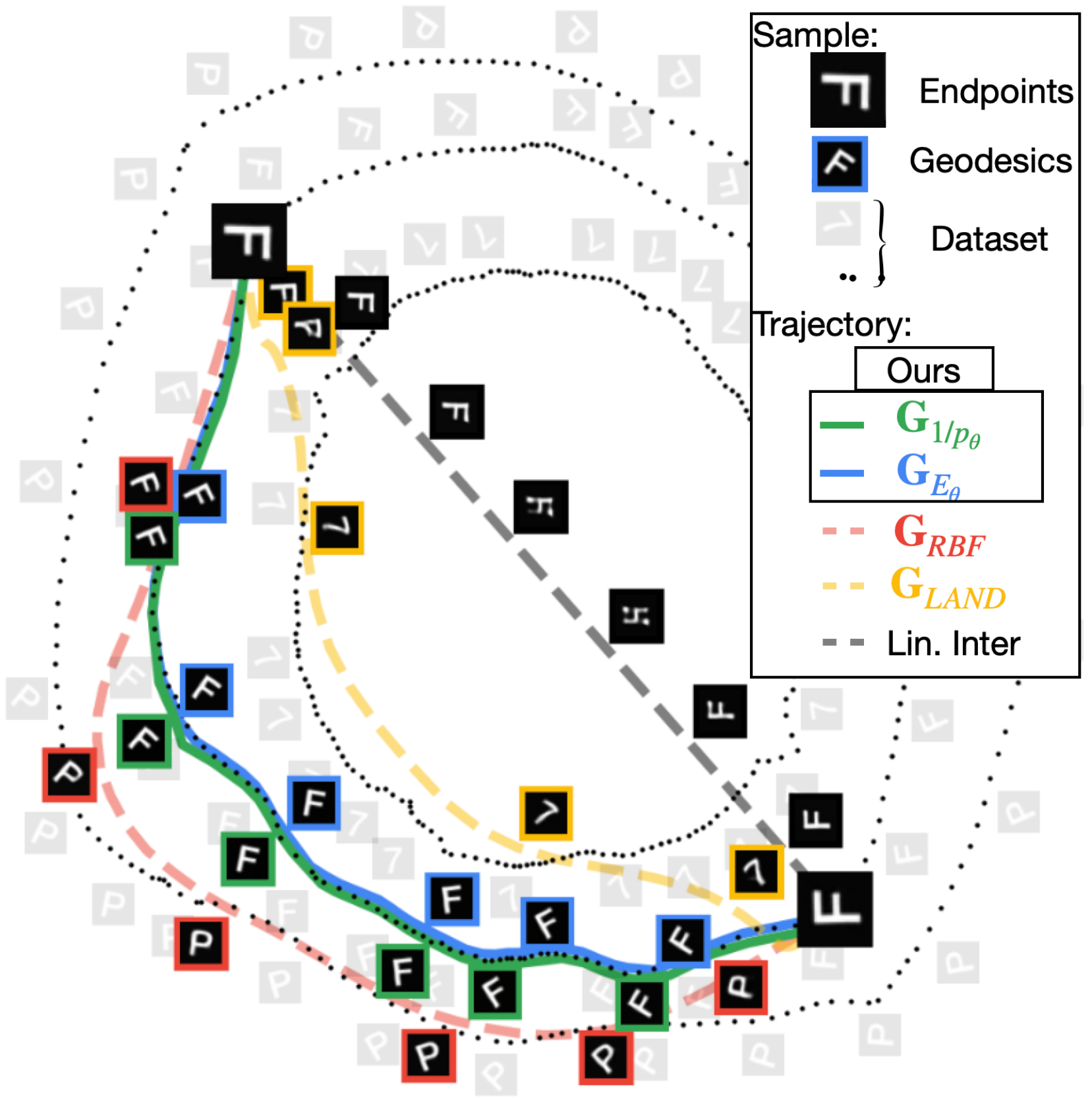}
  \caption{{\bf Geodesics visualization for the URC dataset}. Trajectories and samples are projected in the PCA space for visualization. }.
  \label{fig:fig0}
  %\vspace{-7mm}
\end{wrapfigure}

%\begin{tcolorbox}[questionbox]
%\textit{What does it mean for two data points to be "close" in a high-dimensional space?} %% LOOK AT THIS SUPER STYLE (optional though no pressure) %% pas sur d'aimer, ca prend de la place ... 
%\end{tcolorbox}
% Stating the problem
\textit{What is the shortest path between two data points in a high-dimensional space?} 
In Euclidean geometry, the answer is a straight line. But in modern machine learning, where data often lies on unknown curved manifolds within a high-dimensional space, straight lines slice through regions without data (see linear interp. in Fig.~\ref{fig:fig0}). Capturing the true geometry of data is therefore critical in fields where distance-based analyses depend on underlying structure,
 %in fields where meaningful relationships rely on distance measures that reflect the underlying structure, 
such as vision~\citep{vemulapalli2014human,harandi2014manifold,tuzel2006region}, language~\citep{nickel2017poincare,tifrea2018poincar}, biology~\citep{feng2024multiscale}, and cognitive science~\citep{horibe2023geodesic,neilson2018riemannian}. Riemannian geometry offers a principled way to navigate these spaces by introducing a smoothly varying local metric, the Riemannian metric, which encodes how space bends and stretches~\cite{do1992riemannian}.  Within this framework, the shortest path between two points is no longer a straight line, but a geodesic—a curve that follows the intrinsic curvature of the manifold. Computing geodesics requires knowing the underlying Riemannian metric, but estimating such a metric for complex, high-dimensional data remains a major challenge in machine learning.

A promising strategy for deriving Riemannian metrics is to take a data-driven approach—learning the metric directly from the data itself. This approach estimates the data density and turns it into a Riemannian metric that contracts high-density regions and dilates low-density ones, aligning the geometry with the data manifold~\citep{hauberg2012geometric} (see \S~\ref{sec:related} for more details).
%This line of work first estimates the data’s probability density $p(\v{x})$ and then converts that density into a Riemannian metric: regions where $p(\v{x})$ is high are pulled closer together, while low-density zones are stretched apart, so the geometry conforms to the data manifold~\citep{hauberg2012geometric} (see \S~\ref{sec:related} for alternatives). %This data-driven approach seeks to estimate the geometry of data from its underlying probability distribution~\citep{hauberg2012geometric} (see \S~\ref{sec:related} for a review of alternative approaches). 
Existing methods, such as kernel-based estimators~\citep{arvanitidis2016locally}, normalizing flows~\citep{brehmer2020flows}, and density-based constructions~\citep{arvanitidis2020geometrically}, have succeeded in low-dimensional settings. However, their performance often degrades in high dimensions, where sparse local sampling makes it hard to capture reliable geometric structure~\citep{gruffaz2025riemannian,lebanon2012learning}. Meanwhile, recent advances in generative AI~\citep{du2019implicit,ho2020denoising,rombach2022high} have produced models capable of capturing complex data distributions in high-dimensional spaces with remarkable accuracy. \textit{If these models can learn the data distribution, can they also reveal its underlying geometry?}

In this article, we answer affirmatively by proposing to derive Riemannian metrics from pretrained Energy-Based Models (EBMs)~\citep{du2019implicit,salakhutdinov2009deep,song2021train}.
EBMs are a flexible class of generative models that define an energy function $E_{\theta}$, parameterized by a neural network, assigning low energy to likely data points (i.e., $p_{\theta}(\v{x})\propto \exp(-E_{\theta}(\v{x}))$).%The corresponding probability distribution is akin to a Boltzmann distribution (i.e., $p_{\theta}(\v{x})\propto \exp(-E_{\theta}(\v{x}))$.
We show that the energy landscape of an EBM encodes a rich geometric structure and can be leveraged to derive effective Riemannian metrics. Specifically, we introduce two novel conformal Riemannian metrics—metrics that scale the identity by a positive scalar function: \logebm{G} proportional to the energy itself, and \invebm{G}, proportional to the inverse unnormalized density. We evaluate both against established alternatives (\rbf{G}~\cite{arvanitidis2020geometrically} and \landm{G}~\cite{arvanitidis2016locally}) across datasets of increasing complexity—from toy distributions with known geodesics (see \S~\ref{sec:CircularGaussians}), to rotated character images where the manifold structure is partially known (see \S~\ref{sec:RotatedCharacters}), and finally to high-dimensional natural images where no ground truth geometry is available (see \S~\ref{sec:AFHQ}). \rev{Throughout this work, we adopt the common choice of equipping the data space with a density-based Riemannian metric, thereby defining the geometry of the manifold in terms of data concentration.} We show that EBM-based metrics yield geodesics that (i) remain closer to the data manifold and (ii) better reflect its intrinsic curvature (see Fig.~\ref{fig:fig0} for a visualization of the geodesics). 

Overall, our contributions are summarized as follows:
%\begin{compactitem}
\begin{itemize}[nosep,leftmargin=2em,topsep=0pt]
    \item  We propose a novel framework based on pretrained Energy-Based Models (EBMs) to derive Riemannian metrics. In particular, we introduce two novel conformal metrics \logebm{G} and \invebm{G}, based on the data log-likelihood and data density, respectively.
    \item We demonstrate that these EBM-derived metrics yield geodesics that remain closer to the data manifold and better reflect its curvature.
    \item We show that the proposed EBM-based metrics scale more robustly than prior approaches.
\end{itemize}
%\end{compactitem}

By grounding Riemannian metrics in generative AI, we hope to initiate a new paradigm for understanding and navigating the hidden geometry of high-dimensional data spaces.

\section{Related Work} \label{sec:related}
\paragraph{The many facets of data geometry:} A variety of approaches have been proposed to study the geometry of data:
\begin{itemize}[nosep,leftmargin=2em]
    \item \textit{Information Geometry}: This historical approach is rooted in the work of~\citet{rao1992information} and~\citet{amari1983foundation}. It connects statistics and differential geometry by interpreting the Fisher information~\citep{fisher1922mathematical} as a Riemannian metric on the manifold of parameters of a statistical model. In contrast, our work derives  Riemannian metrics directly from the data space using the energy or the likelihood of an EBM. 
    %I am removing EBM because EBM is nearly just a model used to compute energy, which only depends on the pdf. The metric exists independently of a model. 
    \item \textit{Data-Space induced metrics}:
    Closer to our work, this approach estimates Riemannian metrics directly from samples. %For instance, the LAND metric~\citep{arvanitidis2016locally} uses local covariances of nearby points, while the RBF metric~\citep{arvanitidis2020geometrically} defines a conformal metric using a kernel density estimator with learnable weights. 
    The LAND metric~\citep{arvanitidis2016locally} derives a local metric tensor from the empirical covariance of nearby points.
    The RBF metric~\citep{arvanitidis2020geometrically} defines a conformal metric using an RBF network trained as a parametric KDE, learning centres, widths, and weights so its output forms an unnormalised data density. Both serve as baselines in our study (see \S~\ref{sec:baseline}) and have recently been used for geodesic fitting via flow matching~\citep{kapusniak2025metric}. The (unpublished) work of~\citet{perone2024geometry} was also a key inspiration, proposing to build metrics from the score function of a generative model—an idea also explored by~\citet{diepeveen2024score}. 
    \item \textit{Latent-Space induced metrics}: Another line of work uses pullback geometry~\citep{arvanitidis2021pulling,beik2021learning,kalatzis2020variational,sungeometry,arvanitidis2019fast,chen2019fast}, mapping the Euclidean metric from a network's latent space to the data space—typically through the Jacobian of a VAE encoder~\citep{arvanitidis2018latent}. While our method operates in the latent space of a VAE in high-dimensional settings, the metric is derived from the energy of the EBM and remains independent of the VAE encoder.%the metric is derived directly from the energy of an EBM and is independ of the VAE encoder. %from the energy or the density. % It does not depend on the geometry induced by the VAE. --> a bit, for some XPs so let's not write it
    \item \textit{Generative modeling on a pre-defined manifold:}  Recent approaches such as flow-based models~\citep{lipman2022flow,de2023augmented} and Schrödinger bridges~\citep{wang2021deep,shi2023diffusion} learn transport paths between distributions, sometimes defined over Riemannian manifolds~\citep{chenflow2024,de2024pullback,de2022riemannian,thornton2022riemannian}. These methods assume a known, fixed manifold geometry (e.g., a hypersphere) and design generative models to operate within that structure. In contrast, our approach starts from a generative model—an EBM—and derives the Riemannian metric itself from the model, allowing the geometry to emerge from the data.
    %Recently, flow-based generative models~\citep{lipman2022flow,de2023augmented} and Schrödinger bridges~\citep{wang2021deep,shi2023diffusion} learn transport paths between distributions, including on Riemannian manifolds~\citep{chenflow2024,de2024pullback,de2022riemannian,thornton2022riemannian}. These approaches assume a fixed, known Riemannian structure (e.g, a sphere) and build generative models on top of it. In contrast, we start from a generative model—specifically an energy-based model (EBM)—and use it to derive the Riemannian metric that defines the data manifold.
\end{itemize}    
%\end{compactitem}
%

For a more detailed review of the related work (including topological data analysis, symmetries, computer graphics, or metric learning), see Supp.~\ref{app:extended_related} and ~\citep{peyre2010geodesic,gruffaz2025riemannian}.

\paragraph{Energy-Based Models (EBMs):} EBMs, trained via maximum likelihood~\citep{du2019implicit} (see \S~\ref{sec:ebm}), are particularly well-suited for deriving Riemannian metrics. Their contrastive training, combined with Langevin dynamics sampling, encourages learning a \textit{global} energy landscape that assigns meaningful values across the entire ambient space, including regions far from the data manifold. In contrast, normalizing flows~\citep{rezende2015variational} are limited by their invertible architecture~\citep{kong2020expressive,draxler2024free} and tend to perform poorly on out-of-distribution data~\citep{kirichenko2020normalizing}, sometimes leaking probability mass outside the support~\citep{kelly2023variations}. EBMs trained with diffusion losses~\citep{du2023reduce} or distilled from diffusion models~\citep{thorntoncontrolled} generate high-quality samples, but their energy function depends on a time-indexed noise scale, limiting them to local rather than global energy landscapes. This makes them unsuitable for defining a consistent Riemannian metric. Prior work has used the global energy landscape of EBMs trained via maximum likelihood for trajectory planning in robotics~\citep{du2020model}, though not in the context of geodesics.

\section{Method}
{\bf Notation}: Scalars are denoted by plain lowercase (e.g., x), vectors by bold lowercase (e.g., $\v{x}\in\mathbb{R}^{D}$), and matrices by bold uppercase (e.g., $\v{X}$). Let $\v{I}$ be the identity matrix of $\mathbb{R}^{D \times D}$. $\mathcal{S}^{D}_{++}$ is the set of symmetric $D\times D$ positive definite matrices. Let $\mathcal{M}$ be a Riemannian manifold, with tangent space at $\v{x} \in \mathcal{M}$ denoted $\mathcal{T}_{\v{x}}^{\mathcal{M}}$. Herein, we assume that $\mathcal{M}$ is embedded in a $D$-dimensional Euclidian space ($\mathcal{M}\subset\mathbb{R}^{D}$).

\subsection{A primer on Riemannian geometry}

A Riemannian manifold $(\mathcal{M},\v{G})$ is a smooth manifold $\mathcal{M}$ (i.e., a set locally homeomorphic to $\mathbb{R}^D$) equipped with a Riemannian metric $\v{G} : \mathcal{M} \rightarrow \mathcal{S}^{D}_{++}$. $\v{G}$ defines a smoothly changing inner product on the tangent space $\mathcal{T}_{\v{x}}^{\mathcal{M}}$ at each point $\v{x}\in\mathcal{M}$ : $\langle \v{u}, \v{v}\rangle_\v{x} = \v{u}^\top \v{G}(\v{x})\v{v}$, with $\v{u},\v{v} \in \mathcal{T}_{\v{x}}^{\mathcal{M}}$~\citep{do1992riemannian}. The length of a curve $\boldsymbol{\gamma} : [0,1] \rightarrow \mathcal{M}$ linking two points $\v{x_0}=\boldsymbol{\gamma}(0)$ and $\v{x_1}=\boldsymbol{\gamma}(1)$ ($\v{x_0}$, $\v{x_1} \in \mathcal{M}$), is measured as:
\begin{equation}
L(\boldsymbol{\gamma}) = \int_0^1 \sqrt{\langle \dot{\boldsymbol{\gamma}}(t), \dot{\boldsymbol{\gamma}}(t)\rangle_{\boldsymbol{\gamma}(t)}}dt.
\label{Eq:eq_length}
\end{equation}
In Eq.~\ref{Eq:eq_length}, $\dot{\boldsymbol{\gamma}}(t)$ denotes the velocity vector of the curve $\boldsymbol{\gamma}(t)$, which lies in the tangent space at that point (i.e., $\dot{\boldsymbol{\gamma}}(t) \in \mathcal{T}_{{\boldsymbol{\gamma}}(t)}^{\mathcal{M}}$). The minimizer of Eq.~\ref{Eq:eq_length} is called a \textit{geodesic}; it represents the (locally) shortest path between $\v{x_0}$ and $\v{x_1}$. In this work, we minimize the kinetic energy functional instead of the length (see Eq.~\ref{Eq:eq_energy}). 
Although both functionals yield the same geodesics up to a parametrization, %, which means that an arbitrary geodesic \textit{could} be a local maximum or a local minimum. In our case, the manifold covers the whole ambient space, and in practice, only local minima are retrieved by our method.}
%, minimizing energy selects the unique constant Riemannian speed parametrization, which is more convenient for optimization and numerical stability\citep{do1992riemannian,arvanitidis2020geometrically}:
 minimizing the kinetic energy functional results in a constant Riemannian speed parametrization\footnote{%Technically, a geodesic only \textit{extremalizes} the kinetic energy functional; in practice, our method retrieves local minima since the manifold spans the full ambient space. 
With length fixed, the strictly convex energy $E=\tfrac12\int_0^1 v(t)^2dt$ attains its minimum—by Jensen’s inequality—only when the speed $v(t)$ is constant.}. This property simplifies optimization and improves numerical stability~\citep{do1992riemannian,arvanitidis2020geometrically}.
\begin{equation}
%\vspace{-5mm}
\boldsymbol{\gamma}^{\star}(t) = \argmin_{\boldsymbol{\gamma}} \mathcal{E}[\boldsymbol{\gamma}] \text{\;\;s.t.\;\;} \mathcal{E}[\boldsymbol{\gamma}] = \frac{1}{2} \int_0^1 \langle \dot{\boldsymbol{\gamma}}(t), \dot{\boldsymbol{\gamma}}(t)\rangle_{\boldsymbol{\gamma}(t)} \, dt.
\label{Eq:eq_energy}
\end{equation}
In the Euclidean case ($\mathcal{M}=\mathbb{R}^D$, $\v{G}(\v{x}) = \v{I}$), $\mathcal{E}$ is equivalent to the kinetic energy of a unit-mass particle moving along $\boldsymbol{\gamma}(t)$, hence the name kinetic energy functional. 
%In the Euclidean case (i.e., $\mathcal{M}=\mathbb{R}^D$ and $\v{G}(\v{x}) = \v{I}$), the energy functional simplifies to the integral of the squared velocity norm (i.e., $E[\boldsymbol{\gamma}]=\frac{1}{2}\int_0^1||\dot{\boldsymbol{\gamma}}(t)||^{2}dt$), which is mathematically equivalent to the total kinetic energy of a particle of unit mass moving along the path $\boldsymbol{\gamma}(t)$ over time.

To avoid the computational cost of solving Eq.~\ref{Eq:eq_energy} for each new pair $(\v{x}_{0}, \v{x}_{1})$ at inference time, we follow~\cite{kapusniak2025metric} and 
approximate the geodesic with a neural interpolant $\boldsymbol{\varphi}_\eta$ (with parameters $\eta$).%, following the approach of~\cite{kapusniak2025metric}:
\begin{equation} 
\v{x}_{t,\eta} = (1{-}t)\v{x}_0 + t\v{x}_{1} + 2t(1{-}t)\boldsymbol{\varphi}_{\eta}(\v{x}_0, \v{x}_1, t). 
\label{Eq:geodesic_parametrization}
\end{equation} 

This parameterization satisfies the boundary conditions ($\v{x}_{0,\eta}{=}\v{x}_0$, $\v{x}_{1,\eta}{=}\v{x}_1$). In Eq.~\ref{Eq:geodesic_parametrization}, $\boldsymbol{\varphi}_{\eta}$ serves as a nonlinear correction to the linear path, allowing the learned path to bend toward the data manifold. We train a single interpolant network $\boldsymbol{\varphi}_{\eta}$ over batches of random endpoint pairs so it can approximate geodesics between arbitrary points (see Algo.~\ref{algo:geodesic-interpolant}). \rev{Intuitively, our geodesic interpolant begins with a straight line between the endpoints and uses a neural network to compute a smooth curvature relative to this baseline---bending the path toward regions of higher data density, much like pulling a string taut over a curved surface that reflects the geometry of the data.}
%using Algo.~\ref{algo:geodesic-interpolant} over batches of randomly selected endpoint pairs. The network learns a global geodesic interpolator—approximating geodesics between any two points, not just a specific pair. 
Unlike~\citet{kapusniak2025metric}, who use full autodifferentiation to compute $\dot{\v{x}}_{t,\eta}$, we opt for finite difference instead. We found this approach more stable and accurate when using a fine-time discretization.

\begin{wrapfigure}{r}{0.6\textwidth}
\vspace{-5mm}
  \begin{minipage}{0.6\textwidth}
    \begin{algorithm}[H]
    \SetAlgoLined
    \KwIn{Endpoints pairs: ($\{\v{x}_0\}$, $\{\v{x}_1\}$), Interp. net.: $\boldsymbol{\varphi}_{\eta}$, Metric: $\v{G}$, Time steps: T}
    %\KwOut{Processed data}
    dt=$\frac{1}{\textnormal{T}-1}$, t=[0:1:dt]\\
    \While{\textnormal{training}}{
    $\v{x}_0\sim\{\v{x}_0\}$ and $\v{x}_1\sim\{\v{x}_1\}$\textcolor{gray}{\,\,\,\#\#\,sample batch of pairs}\\
      $\v{x}_{t,\eta} = (1-t)\v{x}_0 + t\v{x}_1 +2t(1-t)\boldsymbol{\varphi}_{\eta}(\v{x}_0, \v{x}_1,t)$ \\
      $\displaystyle \dot{\v{x}}_{t,\eta} = \frac{\v{x}_{t+1,\eta} - \v{x}_{t,\eta}}{\textnormal{dt}}$ \textcolor{gray}{\,\,\,\#\#\,finite difference}\\
      $\displaystyle \mathcal{L}(\eta) = \mathbb{E}_{\v{x}_0,\v{x}_1}\bigg[\frac{1}{2}\sum_{t=0}^{1}\big[\dot{\v{x}}_{t,\eta}^{\top} \v{G}(\v{x}_{t,\eta}) \dot{\v{x}}_{t,\eta}\big]\textnormal{dt}\bigg]$ \\
      Update $\eta$ using gradient $\nabla_{\eta}\mathcal{L}$
    }
    \caption{Training geodesic interpolant}
    \label{algo:geodesic-interpolant}
    \end{algorithm}
  \end{minipage}
  \vspace{-10pt} % Adjust if needed
\end{wrapfigure}
Although Algo.~\ref{algo:geodesic-interpolant} approximates geodesics for a given metric $\v{G}$, the trajectories may initially deviate from the data manifold—especially early in training, when they are initialized as straight lines in the ambient space. However, if (i) the eigenvalues of $\v{G}$ are large when off-manifold and (ii) small %$\v{G}(\v{x}) \approx \v{I}$ 
when on-manifold, then the interpolated points $\v{x}_t$ are progressively drawn toward the manifold during optimization~\citep{kapusniak2025metric,arvanitidis2020geometrically}. In other words, an effective $\v{G}$ should penalize off-manifold directions and encourage paths through high-density paths, steering the geodesics along true data geometry. This insight suggests that defining the metric as a decreasing function of the data probability (e.g., $\v{G}(\v{x}) \propto p(\v{x})^{-1}\cdot\v{I} $) can effectively steer trajectories toward high-density regions. In practice, however, the true data distribution is unknown and only observed through samples. In this work, we use an EBM to approximate the data distribution. %show that using an Energy-Based Model (EBM) to approximate the data distribution leads to a metric that more effectively aligns geodesics with the data manifold than alternative approaches.

\subsection{Energy-Based Models}\label{sec:ebm}
Let $p_{\mathcal{M}}$ be the true data distribution supported on the manifold $\mathcal{M}$, such that  $\int_{\v{x}\in\mathcal{M}} p_{\mathcal{M}}(\v{x})d\v{x}=1$. In practice, we do not have access to $p_{\mathcal{M}}$ directly, but only to a finite set of samples $\mathcal{D}=\{ \v{x}_i\}_{i=1}^N$drawn from it. These samples define the empirical distribution $p_{\mathcal{D}}$, which we use to train our models. 

Energy-Based Models (EBMs) provide a flexible framework for modeling complex, unnormalized probability distributions—making them particularly well-suited for data concentrated on low-dimensional manifolds. Here we define the energy function $E_{\theta}(\v{x})\in\mathbb{R}$, parameterized with a neural network with weights $\theta$. This energy induces a probability distribution of the form: \begin{equation}
p_{\theta}(\v{x})=\frac{\exp\big(-E_{\theta}(\v{x})\big)}{Z(\theta)}\,\,\,\,\textnormal{where} \,\,\,Z(\theta)=\int \exp\big(-E_{\theta}(\v{x})\big)d\v{x}.  
\label{eq:EBM_boltzman}
\end{equation}
Our goal is to train the EBM so that $p_{\theta}$ approximates the data distribution $p_{\mathcal{M}}$. To do so, we minimize the negative log-likelihood w.r.t to the empirical distribution: $\mathcal{L}_{ML}(\theta) = \mathbb{E}_{\v{x}~\sim p_D} [-\log p_{\theta}( \v{x})]$. Although the partition function $Z(\theta)$ is intractable, previous works have shown that the gradient of this objective can be estimated without computing 
$Z(\theta)$ explicitly~\citep{hinton2002training,woodford2006notes} (see Supp.~\ref{app:EBM_ML_derivation} for the demonstration), a loss known as \textit{contrastive divergence}:
\begin{equation}
\nabla_{\theta} \mathcal{L}_{ML} \approx \mathbb{E}_{\v{x}^{+} \sim p_{\mathcal{D}}}[E_{\theta}(\v{x}^{+})] - \mathbb{E}_{\v{x}^{-} \sim p_{\theta}}
[E_{\theta}(\v{x}^{-})]
\label{eq:ml_ebm}
\end{equation}
where $\v{x}^{+}$ are data samples and $\v{x}^{-}$ are samples drawn from the model distribution $p_{\theta}$ using Langevin dynamics. %This sampling method approximates $p_{\theta}$ by iteratively following the negative gradient of the energy with added noise. 
We adopt the training procedure of~\cite{du2019implicit}, which is known to scale well (see Supp.~\ref{app:EBM} for the full pseudo-code). %The full training algorithm%, along with architectural details and dataset-specific hyperparameters, 
%is provided in Supp.~\ref{app:EBM}. 
From this point on, we refer to $E_{\theta}$ as a pre-trained energy function. 

EBM can be hard to train in high-dimensional pixel space, especially because of the sampling procedure~\citep{nijkamp2020anatomy,duvenaud2021no,xiao2021vaebm}. For complex tasks, we follow standard practice and operate in the latent space of a pretrained VAE~\citep{pang2020learning}, where all baselines are evaluated for fairness. \rev{To improve the EBM training training stability, we regularize the contrastive divergence loss with a denoising term, which preserves the global structure of the energy landscape while enhancing convergence---a technique we find both effective and broadly applicable.}%Therefore, for the most challenging tasks, we adopt the common approach in generative modelling of working in the latent space of a pretrained VAE~\citep{pang2020learning}. For fairness, concurrent methods are also evaluated in the same latent space.  

%\textbf{Global energy landscape.} EBMs trained with contrastive divergence have the appealing property of being \textit{global}: they assign an energy to every point in the ambient space, including outside the support of observed data. Other generative models like normalizing flows~\citep{rezende2015variational}, VAE~\citep{kingma2013auto}, or even simulation-free EBMs (e.g distilled from diffusion models~\citep{thorntoncontrolled} or trained with energy discrepancy~\citep{schroder2023energy}) do not share this property. We need it to ensure that the metric is well-defined over the whole space.

\subsection{EBM-derived Riemannian Metrics}\label{sec:Riemann_metric}

Here, we describe the EBM-derived metrics \logebm{G}, \invebm{G}. For details on the baseline Riemannian metrics \landm{G},\rbf{G}, see \S~\ref{sec:baseline}.
%We evaluate four Riemannian metrics in our experiments, denoted \logebm{G}, \invebm{G}, \landm{G}, \rbf{G}. The first two, \logebm{G} and \invebm{G}, are derived from EBMs, while \rbf{G}~\citep{arvanitidis2020geometrically,kapusniak2025metric} and \landm{G}~\citep{arvanitidis2016locally} are established parametric metrics from the Riemannian geometry literature. 
To ensure a fair comparison —and following standard practice in the field~\citep{peyre2010geodesic,hauberg2018only, arvanitidis2021prior}— all metrics are cast using a shared parametric form:
\[
\mathbf{G}(\boldsymbol{x}) = 
\begin{cases}
\alpha \, \mathbf{h}(\boldsymbol{x}) + \beta & \text{for } \logebm{G}, \\
\left( \alpha \, \mathbf{h}(\boldsymbol{x}) + \beta \right)^{-1} & \text{for } \invebm{G}, \landm{G}, \rbf{G},
\end{cases}
\]
where \( \v{h}(\v{x}) \) is a metric-specific, positive-definite function (either scalar, diagonal, or matrix), and \( \alpha \), \( \beta \) are calibration constants. These constants are chosen so that the metric scale to \( \v{I} \) on the data manifold and to \( 10^3 \cdot \v{I} \) in low-density regions\footnote{Note that this multiplicative factor amounts to a change of unit, to ensure reasonable scaling of the lengths, but the induced geodesics are only determined by the ratio $\alpha/\beta$.}. \rev{This allows fair comparison across metric choices without introducing significant sensitivity to hyperparameter tuning.} Further details about the metric calibration procedure are provided in Supp.~\ref{app:metric_normalization}. 
Importantly, all EBM-derived metrics are \textit{conformal}, they take the form $\lambda(\v{x})\v{I}$, where $\lambda$ is a scalar function. In other words, they scale the identity matrix uniformly in all directions, resulting in isotropic metrics:

\begin{itemize}[left=0em,nosep,topsep=0pt]
  \item[\textcolor{google_blue}{$\bullet$}] \logebm{G} defines a Riemannian metric by directly scaling the raw energy of a pretrained EBM. This is the simplest —yet surprisingly effective—formulation we consider: 
  \begin{equation}
  \textcolor{google_blue}{\mathbf{G}_{\text{E}_{\theta}}(\v{x})}=(\alpha*E_{\theta}(\v{x}) + \beta) \cdot \mathbf{I}.
  \label{eq:g_logebm}
  \end{equation}
 Intuitively, high-energy (low-density) regions receive a larger metric, penalizing movement away from the data. %Note that this is an affine rescaling of the negative log-likelihood $-\log p_{\mathcal{D}}$.
 Note that $E_\theta$ is an affine rescaling of the negative log-likelihood $-\log p_{\mathcal{D}}$.
 % For a judicious joice of $\alpha,\beta$ this metric is identical to $-\log{p_{\mathcal{M}}(x)}\v{I}$: the ``cost'' of movement is proportional to the negative log-likelihood of the data.

  \item[\textcolor{google_green}{$\bullet$}] \invebm{G} leverages the inverse of an unnormalized probability estimate:
  \begin{equation}
  \textcolor{google_green}{\mathbf{G}_{1/p_{\theta}}(\v{x})}= (\alpha*\exp(-E_{\theta}(\v{x})) + \beta)^{-1} \cdot \mathbf{I}.
  \label{eq:g_invebm}
   \end{equation}
   Compared to \logebm{G}, this metric applies an inverse to a decreasing exponential, forming a strong barrier against low-density regions. It stays small near the data manifold but rises sharply elsewhere, acting as a repulsive force. Its key advantages are: (i) a clear probabilistic interpretation via the unnormalized density, and (ii) direct comparability to \landm{G} and \rbf{G} as they share the same inverse form.
   %Compared to \logebm{G}, this formulation combines an inverse function with a decreasing exponential, which together create a strong barrier to paths entering low-density regions. The metric remains small and stable near the data manifold but grows rapidly in low-density areas, acting as a strong repulsive force. %Like \logebm{G}, it is a conformal Riemannian metric. 
   %Its main advantages are: (i) it has a clear probabilistic interpretation based on the unnormalized density, and (ii) it is directly comparable to \landm{G} and \rbf{G}, as they all share the same inverse form. 
\end{itemize}

In the next section, we introduce the baseline Riemannian metrics used for comparison. We also empirically evaluate their behavior across datasets of increasing complexity, focusing on how they capture the underlying manifold and shape geodesic paths.
%These four metrics span a range of approaches to modeling data geometry and serve as strong, interpretable baselines. In the next section, we evaluate their empirical behavior across datasets of increasing complexity, focusing on how they capture the underlying manifold and shape geodesic paths.

  %\textcolor{google_orange}{$$\displaystyle G_{LAND}=(\alpha*\textnormal{diag}(\v{h}(\v{x})) + \beta \mathbf{I})^{-1}\,\,\textnormal{with} \,\,h_{j}(\v{x})= \sum_{i=1}^{N}(x_k^{i} - x^{i})^{2}\exp{\bigg(-\frac{||\v{x} - \v{x_j} ||^{2})}{2\sigma^{2}}\bigg)}$$}

\section{Experiments}
\subsection{Baseline Riemannian Metrics}\label{sec:baseline}
\rbf{G}~\citep{arvanitidis2020geometrically,kapusniak2025metric} and \landm{G}~\citep{arvanitidis2016locally} are established metrics from the Riemannian geometry literature:

\begin{itemize}[left=0em,nosep]
\item[\textcolor{google_orange}{$\bullet$}] $\textcolor{google_orange}{\mathbf{G}_{\text{LAND}}}$, also known as the LAND metric~\citep{arvanitidis2016locally}, is a nonparametric Riemannian metric that adapts to the local geometry of the dataset. Around each point \( \v{x} \), it estimates a Gaussian distribution by weighting all data points \( \{\v{x}_i\}_{i=1}^{N} \) according to their distance to \( \v{x} \): %The metric is then defined as the inverse of a diagonal covariance matrix estimated via:
 \begin{equation}
  \displaystyle \textcolor{google_orange}{\mathbf{G}_{\text{LAND}}(\v{x})} = (\alpha\, \textnormal{diag}(\v{h}(\v{x})) + \beta \mathbf{I})^{-1}\,\,\textnormal{s.t}\,\,h^{(j)}(\v{x})= \sum_{i=1}^{N}(x_i^{(j)}\!-\!x^{(j)})^{2}\exp{\bigg(\!\!\!-\frac{||\v{x}\!-\!\v{x_i} ||^{2}}{2\sigma^{2}}\bigg)}
  \label{eq:LAND}
  \end{equation}Here, \( h^{(j)}(\v{x}) \) measures the local variance along dimension \( j \),  weighted by a Gaussian kernel with bandwidth \( \sigma \).
  %emphasizing contributions from nearby points through a Gaussian kernel with bandwidth \( \sigma \). 
  %Note that \landm{G} is the only non-conformal metric we consider---this is a diagonal metric---, therefore the metric measures local anisotropy. 
  \landm{G} is the only diagonal (i.e., non-conformal) metric we consider, allowing it to model local anisotropy.
  While flexible and model-free, LAND has practical drawbacks: it requires the full dataset at inference, is sensitive to the choice of \( \sigma \), and can behave non-smoothly near sharp neighborhood transitions (see Supp.~\ref{app:land} for examples).
  %While LAND is flexible and model-free, it comes with practical limitations: it requires access to the full dataset at inference time, is sensitive to the choice of \( \sigma \), and can yield non-smooth behavior in regions with sharp changes in neighborhood structure (see Supp.~\ref{app:land} for the effect of varying \( \sigma \)).
  %. We illustrate the effect of varying \( \sigma \) on geodesic paths in Appendix~\ref{app:land}. In practice, we select \( \sigma \) using hyperparameter search.

\item[\textcolor{google_red}{$\bullet$}]$\textcolor{google_red}{\mathbf{G}_{\text{RBF}}}$ is a conformal Riemannian metric %defined as a scalar field $h(\v{x})$ times the identity, where 
in which $h$ is a weighted sum of Radial Basis Functions (RBFs) centered on $K$ cluster centroids $\{\v{\hat{x}}_k\}_{k=1}^{K}$ computed via K-means~\citep{arvanitidis2020geometrically}:
\[
\textcolor{google_red}{\mathbf{G}_{\text{RBF}}(\v{x})} = (\alpha \cdot h(\v{x}) + \beta)^{-1} \cdot \mathbf{I}, \quad
h(\v{x}) = \sum_{k=1}^{K} w_k \exp\left(-\frac{1}{2} \cdot \lambda_k \|\v{x} - \v{\hat{x}}_k\|^2\right).
\]
The weights $w_k$ are trained so that $h(\v{x}) \approx 1$ on the data manifold, and $\lambda_k$ is set from inter-cluster distances (see Supp.~\ref{app:rbf}). This yields a smooth, efficient approximation of the data geometry and scales better than LAND~\citep{kapusniak2025metric}. However, it may miss fine-grained structure, especially in regions of complex or uneven density. Like other methods based on Euclidean distance (and K-means), it suffers from the curse of dimensionality. Its accuracy depends on $K$, $\lambda_k$, and centroid placement (illustrated in Supp.~\ref{app:rbf}). 
\end{itemize}

\rev{The scaling constants ($\alpha$, $\beta$) are introduced to ensure consistent dynamic range across metrics and have minimal impact on convergence or geodesic quality; the number of discretization steps ($T=100$) is chosen as a trade-off between efficiency and accuracy, consistent with prior work.} We evaluate \invebm{G}, \logebm{G}, \rbf{G}, and \landm{G} on three datasets of increasing complexity. Circular Mixture of Gaussians offers full control and ground-truth geodesics. The rotated characters dataset is higher-dimensional but still allows quantitative evaluation. Animal Faces is made of higher-dimensional images but with no ground truth. This progression tests metric performance as data complexity grows. The code to reproduce all our experiments is available at \rev{\url{https://github.com/VictorBoutin/RiemannEBM}}.

%(we set $T{=}100$ in all experiments)

\subsection{Circular Mixture of Gaussians}\label{sec:CircularGaussians}
\begin{figure}[h!]
\begin{tikzpicture}
\draw [anchor=north west] (0\linewidth, 0.98\linewidth) node {\includegraphics[width=0.6\linewidth]{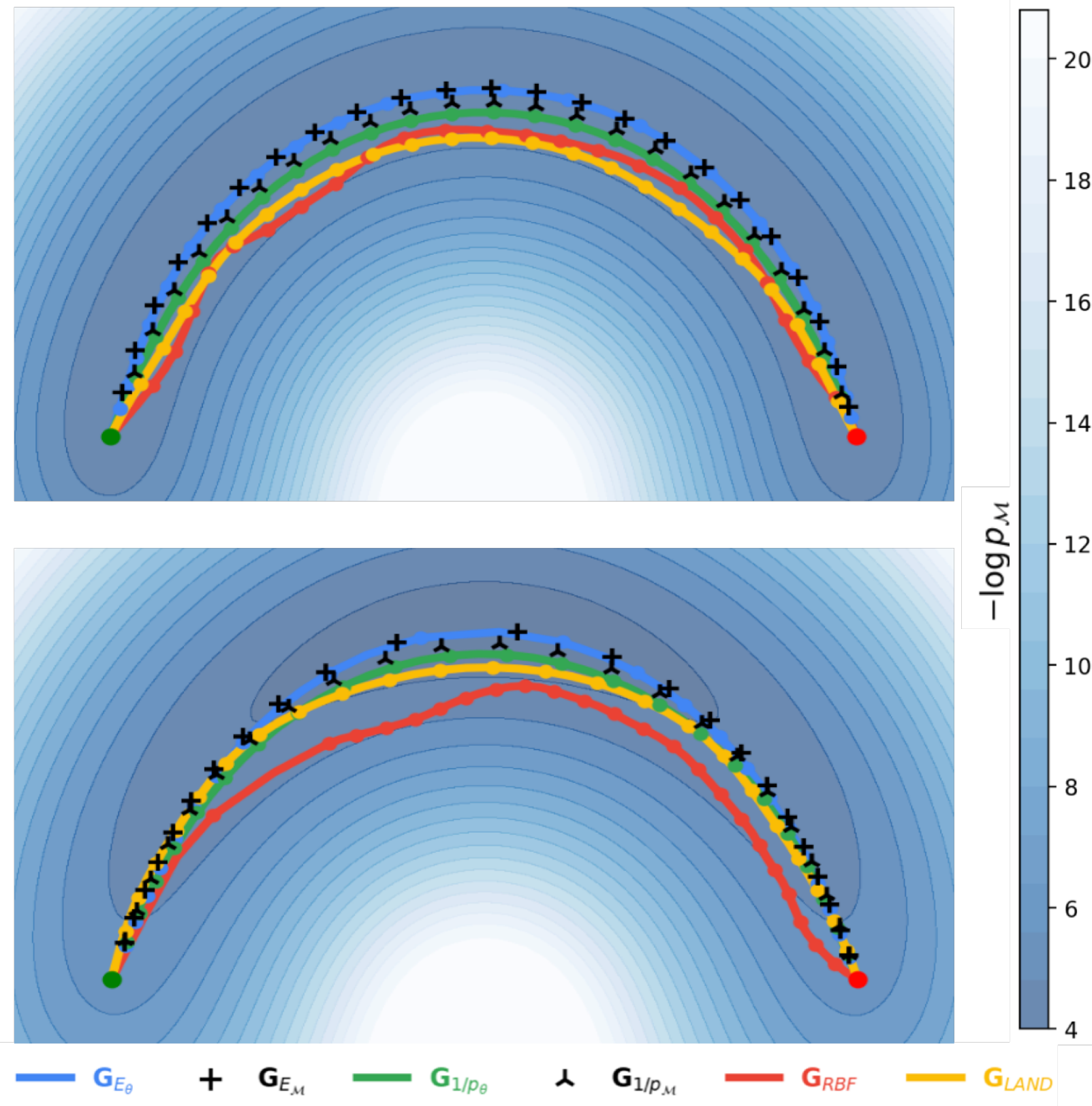}};

\draw [anchor=north west] (0.64\linewidth, 0.96\linewidth) node {
\begin{tabular}{l c c c }
\makecell[c]{Metric} & 
\makecell[c]{$p_{\mathcal{M}}(\v{\gamma}^{\star})$\\
$(\uparrow)$} & 
\makecell[c]{RMSE \\
$(\downarrow)$} \\
\hline
$\boldsymbol{G}_{E_{\mathcal{M}}}$ & $0.79$ & -\\
$\boldsymbol{G}_{1/p_{\mathcal{M}}}$ & $0.77$ & -\\
\hdashline
\logebm{G} & $\mathbf{0.78}$ & $0.12$\\
\invebm{G} & $0.73$ & {\bf 0.10}\\
\hdashline
\landm{G} & $0.60$ & $0.38$\\
\rbf{G} & $0.61$ & $0.39$\\
\end{tabular}
};

\draw [anchor=north west] (0.64\linewidth, 0.66\linewidth) node {
\begin{tabular}{l c c c }
\makecell[c]{Metric} & 
\makecell[c]{$p_{\mathcal{M}}(\v{\gamma}^{\star})$\\
$(\uparrow)$} & 
\makecell[c]{RMSE \\
$(\downarrow)$} \\
\hline
$\boldsymbol{G}_{E_{\mathcal{M}}}$ & 0.67 & -\\
$\boldsymbol{G}_{1/p_{\mathcal{M}}}$ & 0.73 & -\\
\hdashline
\logebm{G} & {\bf 0.67} & 0.18\\
\invebm{G} &  {\bf 0.67} & {\bf 0.14}\\
\hdashline
\landm{G} & 0.65 & 0.34\\
\rbf{G} &  0.47 & 2.2\\
\end{tabular}
};

\begin{scope}
    \draw [anchor=north west,fill=white, align=left] (0\linewidth, 1\linewidth) node {{\bf a)} Geodesics on UCG} ;
    
    \draw [anchor=north west,fill=white, align=left] (0.63\linewidth, 1\linewidth) node {{\bf b)} Geodesics evaluation on UCG};

    \draw [anchor=north west,fill=white, align=left] (0.0\linewidth, 0.7\linewidth) node {{\bf c)} Geodesics on WCG};
    
    \draw [anchor=north west,fill=white, align=left] (0.63\linewidth, 0.7\linewidth) node {{\bf d)} Geodesics evaluation on WCG};
%    \draw [anchor=north west,fill=white, align=left] (0.67\linewidth, 1\linewidth) node {\bf c)};
\end{scope}
\end{tikzpicture}
%\vspace{-7mm}
\caption{{\bf Geodesics on UCG and WCG datasets.} {\bf(a, c)}: Some geodesics obtained on UCG {\bf (a)} and WCG {\bf (c)}, for $6$ different Riemannian metrics.
%using 6 different Riemannian: \logebm{G}, \invebm{G}, \landm{G}, \rbf{G} along with two baseline metrics, $\boldsymbol{G}_{E_{\mathcal{M}}}$, $\boldsymbol{G}_{1/p_{\mathcal{M}}}$. Results are shown for the Uniform Circular Gaussian (UCG) dataset in (a) and the Weighted Circular Gaussian (WCG) dataset in (c). 
The contour plots represent the energy landscape given by $-\log p_{\mathcal{M}}$. {\bf(b, d)} Quantitative evaluation of geodesics on UCG {\bf (b)} and WCG {\bf (d)}. We report (i) the accumulated probability along the geodesic (the higher the better) and ii) RMSE between each geodesic and its corresponding baseline \rev{(i.e., $\boldsymbol{G}_{E_{\mathcal{M}}}$ for \logebm{G}, and  $\boldsymbol{G}_{1/p_{\mathcal{M}}}$ for \invebm{G}, \landm{G} and \rbf{G})}. See Supp.~\ref{app:toy_2sig} for the $2$-$\sigma$ error.
}
\label{fig:fig1}
%\vspace{-4mm}
\end{figure}
We consider two toy datasets built using a mixture of Gaussians arranged along a semicircle. In the first, called Uniform Circular Gaussians (UCG), the Gaussian components have equal weights (see Fig.~\ref{fig:fig1}a). In the second, Weighted Circular Gaussians (WCG), the weights are non-uniform, with higher density near the center of the arc, as reflected by the contour intensity shown in Fig.~\ref{fig:fig1}c. For both datasets, we have access to the closed-form probability distribution of the data, denoted $p_{\mathcal{M}}$ (see Supp. ~\ref{app:circular_dataset_det} for details of $p_{\mathcal{M}}$). We first train an Energy-Based Model (EBM) on each dataset to derive the metrics \logebm{G} and \invebm{G} (see Supp. ~\ref{app:ebm_gaussian} for training details). Then, we apply Algo.~\ref{algo:geodesic-interpolant} to both datasets using all Riemannian metrics described above. Additionally, we include two baseline Riemannian metrics derived directly from the true distribution $p_{\mathcal{M}}$:
\begin{equation}
\mathbf{G}_{\text{E}_{\mathcal{M}}}(\v{x})=-\alpha*\log p_{\mathcal{M}}(\v{x}) \mathbf{I} + \beta \,\,\,\,\,\textnormal{and}\,\,\,\,\, \mathbf{G}_{1/p_{\mathcal{M}}}(\v{x}) = (\alpha*p_{\mathcal{M}}(\v{x}) + \beta)^{-1} \cdot \mathbf{I}
\label{eq:baseline_metric}
\end{equation}

Eq.~\ref{eq:baseline_metric} uses calibration constants $\alpha$ and $\beta$, computed as in other metrics. Some geodesics obtained for the $6$ different metrics are shown in Fig.~\ref{fig:fig1}a and Fig.~\ref{fig:fig1}c for the UCG and WCG datasets, respectively. We refer the reader to Supp.~\ref{app:circular_dataset} for details on network architectures and hyperparameters.

To evaluate geodesic quality, we use two evaluation metrics. The first is the accumulated probability along the geodesic path, $p_{\mathcal{M}}(\boldsymbol{\gamma}^{\star})=\sum_{t=1}^{T}p_{\mathcal{M}}(\v{x}_{t,\eta^{\star}})$. It measures how closely the trajectory aligns with the data manifold — the higher the better. %To quantify the quality of the geodesics, we use 2 different metrics. The first is the accumulated probability along the geodesic, as defined by $p_{\mathcal{M}}(\boldsymbol{\gamma}^{\star})=\sum_{t=1}^{T}p_{\mathcal{M}}(\v{x}_{t,\eta^{\star}})$. It measures how closely the trajectory aligns with the data manifold — the higher the better.
The second is the RMSE to a baseline geodesic computed using the true distribution $p_{\mathcal{M}}$, matched by metric type (e.g., \logebm{G} vs. $\mathbf{G}_{\text{E}_{\mathcal{M}}}$). All quantitative results are averaged over $1,000$ geodesics with distinct endpoints (See Fig.~\ref{fig:fig1}b and~d). \logebm{G} achieves the highest accumulated probability, indicating closest alignment with the data manifold, while \invebm{G} yields the lowest RMSE to its baseline—best approximating the ground-truth geodesic. Both EBM-based metrics consistently outperform other methods across evaluation criteria.

%The second is the RMSE between each learned geodesic and a corresponding baseline geodesic constructed using the true distribution $p_{\mathcal{M}}$. Importantly, each RMSE is computed between geodesics derived from metrics of the same functional form. For example, the geodesic derived from \logebm{G} is compared to its baseline geodesic computed with $\mathbf{G}_{\text{E}_{\mathcal{M}}}(\v{x})$. All quantitative results are averaged over 1,000 geodesics with distinct endpoints (see Fig.\ref{fig:fig1}b and Fig.\ref{fig:fig1}d for the UCG and WCG datasets, respectively). %This ensures that each comparison isolates the effect of using a learned vs. known distribution, without conflating differences in metric structure. 
%Evaluation results are reported in Fig.\ref{fig:fig1}b and Fig.\ref{fig:fig1}c for the UCG and WCG datasets, respectively.

%We observe that \logebm{G} yields the geodesics with the highest accumulated probability $p_{\mathcal{M}}(\boldsymbol{\gamma}^{\star})$, showing it produces trajectories most closely aligned with the data manifold. In contrast, \invebm{G} achieves the lowest RMSE relative to its baseline, suggesting that it best approximates the geodesic that would be obtained if the true distribution were known. Overall, both EBM-based metrics consistently outperform the other approaches across evaluation criteria.

To test how different metrics behave when the density varies along the data manifold, we switch \begin{wrapfigure}{r}{0.4\textwidth}
  \centering
  \includegraphics[width=0.4\textwidth]{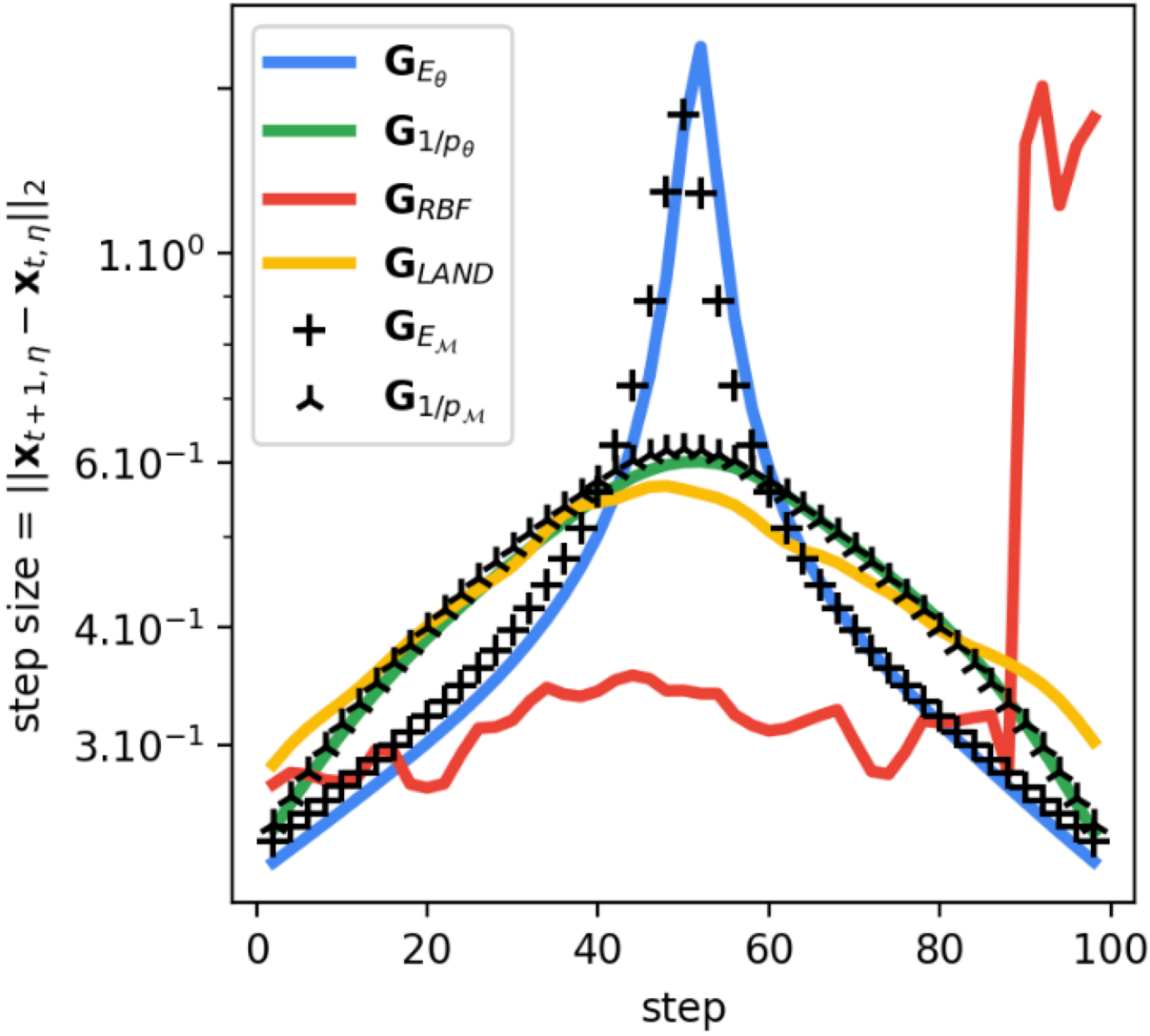}
  \caption{{\bf Step size along geodesics in the WCG dataset}. Log-based metrics (\logebm{G} and $\boldsymbol{G}_{E_{\mathcal{M}}}$) produce sharper variations, reflecting stronger sensitivity to density curvature.}
  \label{fig:curvature}
  \vspace{-10mm}
\end{wrapfigure} from the uniformly populated UCG semicircle to the Weighted Circular Gaussian (WCG), whose samples cluster near the arc’s centre.
%We leverage the WCG dataset to highlight how well geodesics capture manifold curvature---unlike UCG’s which is uniform along the semicircle, WCG concentrates density near the arc’s center. 
As shown in Fig.~\ref{fig:curvature}, log-based metrics (\logebm{G}, $\boldsymbol{G}_{E_{\mathcal{M}}}$) accentuate the manifold curvature more than $1/p$-based ones (\invebm{G}, \rbf{G}, \landm{G}, $\boldsymbol{G}_{1/p_{\mathcal{M}}}$), producing larger steps in high-density regions. This is because $-\log p$ diverges as $p \to 0$, amplifying distortions and speed variations.
%The WCG dataset allows us to evaluate how closely each geodesic follows the manifold’s curvature. Unlike the UCG setting, where density is nearly uniform along the semicircle, the WCG has a peak density at the center of curvature. To investigate this, Fig.~\ref{fig:curvature} shows the Euclidean step sizes along the geodesics from Fig.~\ref{fig:fig1}. We observe that log-based metrics (\logebm{G} and $\boldsymbol{G}_{E_{\mathcal{M}}}$) accentuate the curvature more strongly: they cover greater distances in high-density regions compared to metrics based on $1/p$ (\invebm{G}, \rbf{G}, \landm{G}, and $\boldsymbol{G}_{1/p_{\mathcal{M}}}$). This occurs because $-\log p$ diverges as $p \to 0$, leading to sharper geometric distortions and more pronounced variations in trajectory speed.

\subsection{Rotated Characters}\label{sec:RotatedCharacters}
\begin{figure}[h!]
\begin{tikzpicture}
\draw [anchor=north west] (0.02\linewidth, 0.98\linewidth) node {\includegraphics[width=0.58\linewidth]{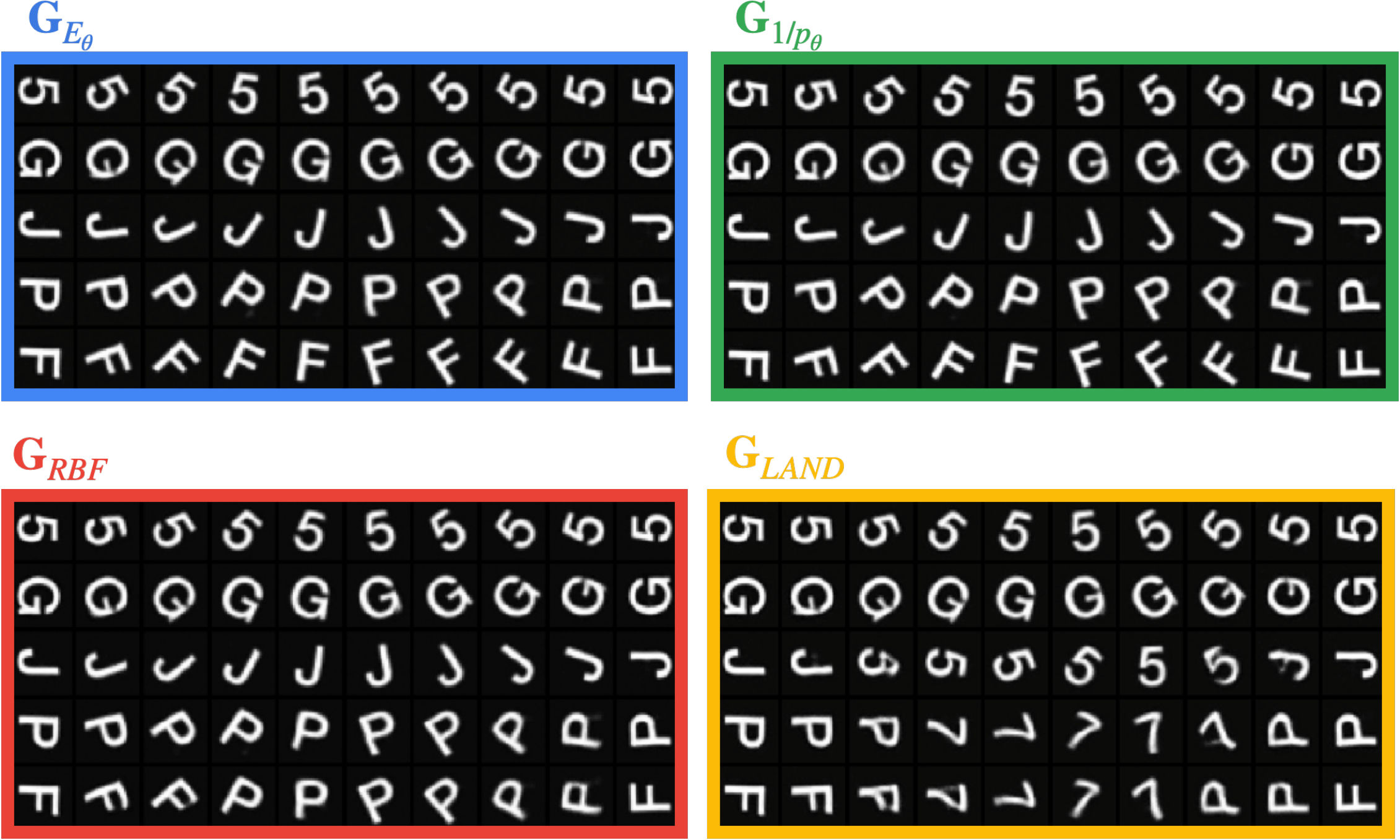}};

\draw [anchor=north west] (0.62\linewidth, 0.90\linewidth) node {
\begin{tabular}{c c c c }
\makecell[c]{Metric} & 
\makecell[c]{$\mathcal{D}$-RMSE\\
$(\downarrow)$} & 
\makecell[c]{$\gamma^{\star}$-RMSE \\
$(\downarrow)$} \\
\hline
\makecell[c]{linear\\interp.}
 & 2.96 & 3.52\\
\hdashline
\logebm{G} & {\bf 0.11} & {\bf 0.40}\\
\invebm{G} & 0.14 & 0.44\\
\hdashline
\landm{G} & 0.66 & 2.39\\
\rbf{G} & 0.36 & 0.86\\
\end{tabular}
};

\begin{scope}
    \draw [anchor=north west,fill=white, align=left] (0.0\linewidth, 1\linewidth) node {{\bf a)}} ;
    
    \draw [anchor=north west,fill=white, align=left] (0.63\linewidth, 1\linewidth) node {{\bf b)}};

\end{scope}
\end{tikzpicture}
%\vspace{-7mm}
\caption{{\bf Geodesics on the URC dataset.} {\bf (a)} Geodesics computed with different Riemannian metrics, projected into pixel space for visualization. \rbf{G} and \landm{G} often deviate from the intended path, sometimes drifting toward other characters (e.g., the letter {\it F}). {\bf (b)} Quantitative evaluation using two metrics: (i) $\mathcal{D}$-RMSE, which measures proximity to the dataset manifold, and (ii) $\gamma$-RMSE, which measures the deviation from an ideal smooth rotation. See Supp.~\ref{app:rot_2sig} for the $2$-$\sigma$ error.}

\label{fig:fig3}
\vspace{-6mm}
\end{figure}
We use an image dataset of seven rotated, non-symmetric characters in two variants: Uniform Rotated Characters (URC), with evenly distributed angles, and Biased Rotated Characters (BRC), concentrated near $0^\circ$. In this subsection, all computations are done in the $64$-dimensional latent space of a regularized autoencoder trained with a triplet loss, ensuring that small angular differences yield short latent distances. 
\rev{This setup provides a unique middle ground: although the underlying Riemannian metric is unknown, we can treat the smooth in-plane rotation between two instances of the same character as a proxy for the ground-truth geodesic. Thanks to the triplet loss, the latent space is structured so that nearby points correspond to slight rotations of the same character, making the shortest path between two orientations a meaningful approximation of the true geodesic in the task-relevant transformation space.}
%This setup lets us test Riemannian metrics in high-dimensional space by treating the smooth rotation that maps one character pose to another as the ground-truth geodesic.
%This setup allows us to evaluate Riemannian metrics in high-dimensional space, using character rotations as a proxy for ground-truth geodesics. 
Separate EBMs and interpolant networks are trained for each dataset variant. Full experimental details (datasets, architectures, and hyperparameters) are provided in Supp.~\ref{app:rotated_char}.
%We use a dataset of seven rotated, non-symmetric typed characters, available in two variants: one with uniformly distributed orientations—Uniform Rotated Characters (URC)—and another with orientations concentrated around 0°—Biased Rotated Characters (BRC) (see Supp.~\ref{app:rotated_char_details} for details). All computations are performed in the 64-dimensional latent space of a regularized autoencoder, rather than in pixel space. The autoencoder is trained with a triplet loss to enforce that small angular differences correspond to shorter distances in latent space, promoting a geometry where slight rotations translate into short paths. This setup enables testing Riemannian metrics in high-dimensional space, using known character rotations as an approximate reference for ground-truth geodesics. Separate EBMs are trained on each dataset variant. See Supp.\ref{app:rotated_char} for all experimental details (datasets, architectures, and hyperparameters).

In Fig.~\ref{fig:fig3}a, we visualize geodesics computed on the URC dataset, projected back into \begin{wrapfigure}{r}{0.4\textwidth}
\vspace{-5mm}
  \centering
  \includegraphics[width=0.4\textwidth]{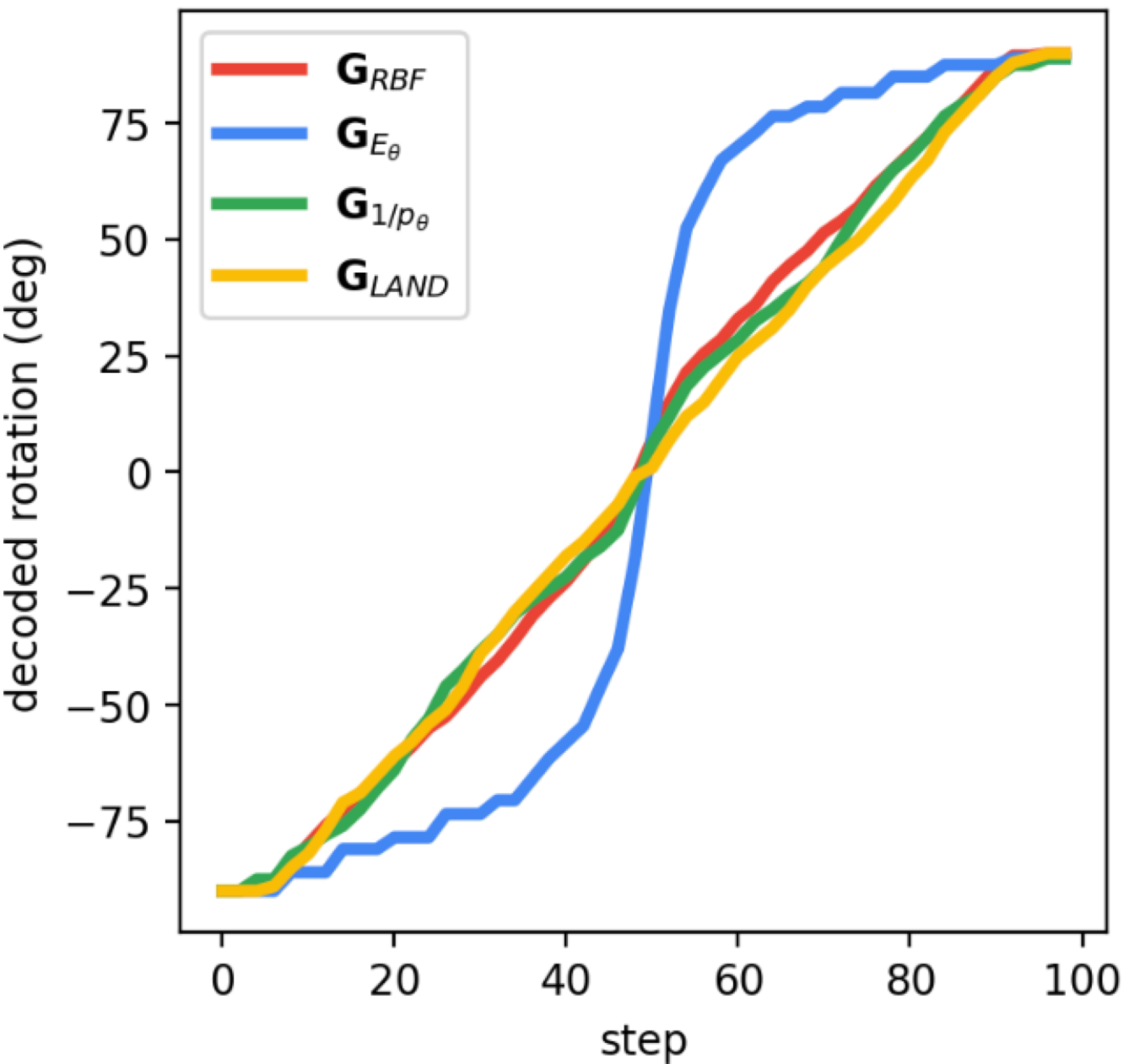}
  \vspace{-7mm}
  \caption{{\bf Step size along geodesics in the WCG dataset}. Log-based metric (\logebm{G}) produces sharper variations, reflecting stronger sensitivity to density curvature.}
  \label{fig:rotation}
  \vspace{-3mm}
\end{wrapfigure}pixel space (see Supp.~\ref{app:More_geodesics_rotated_char} for additional results on both URC and BRC). EBM-based metrics (\logebm{G} and \invebm{G}) yield smooth rotations that preserve character identity, while \rbf{G} and especially \landm{G} often deviate from the intended trajectory. To illustrate these failures, Fig.~\ref{fig:fig0} shows all geodesics projected into PCA space for a case involving the letter F. While \logebm{G} and \invebm{G} remain on the manifold of rotated F instances, linear interpolation cuts through low-density regions, and \rbf{G} and \landm{G} drift toward other character classes. To quantify this, we use two metrics: $\mathcal{D}$-RMSE, which measures the average distance from each geodesic point to its nearest neighbor in the dataset—lower values indicate better adherence to the data manifold; and $\gamma$-RMSE, which evaluates how closely the geodesic follows an ideal smooth rotation between endpoints. All results are averaged over $1,000$ geodesics with random endpoint orientations. As shown in Fig.~\ref{fig:fig3}b, EBM-based metrics consistently outperform others; \rbf{G} performs reasonably well, while \landm{G} shows large deviations on both metrics. Overall, these results suggest that EBM-based metrics scale more effectively to high-dimensional data than alternative approaches.

As in the previous section, we examine how different metrics influence a geodesic’s ability to follow the manifold’s curvature. We focus on the BRC dataset, where orientations are biased toward 0°, creating sharper curvature near that region. To assess this, we decode the orientation at each time step along geodesics connecting fixed endpoints. As shown in Fig.~\ref{fig:rotation}, geodesics under \logebm{G} rotate significantly faster near 0° than those under \invebm{G} and \rbf{G}, reflecting stronger sensitivity to density variations. 

At first glance, it may seem counterintuitive that trajectories following the geodesics move faster in high-density regions. However, this is consistent with minimizing the kinetic energy $\mathcal{E}$ in Eq.\ref{Eq:eq_energy}, which enforces constant Riemannian speed (i.e., the quantity $||\dot{\v{\gamma}}(t)||_{\v{\gamma}(t)}$ is preserved along the trajectory) but not a constant Euclidean speed (i.e., $||\dot{\v{\gamma}}(t)||$ is not constant). Since EBM-derived metrics assign lower Riemannian cost in high-density regions, maintaining constant Riemannian speed requires moving faster in Euclidean terms through these regions. The faster rotation near 0°, observed in Fig.\ref{fig:rotation} and Fig.~\ref{fig:curvature}, thus reflects the lower Riemannian cost of traveling through high-density regions. These results confirm and extend our previous findings: metrics based on energy (i.e., proportional to $-\log p$) more effectively capture the curvature of the data manifold.

\subsection{Animal Faces}\label{sec:AFHQ}

We now evaluate our method on the Animal Faces High Quality (AFHQ) dataset~\citep{choi2020starganv2}, using the latent space of the pretrained Stable Diffusion v1 VAE~\citep{rombach2022high} (latent dimension: $4 \times 16 \times 16$). An EBM is trained to model the distribution of latent codes, and Algo.~\ref{algo:geodesic-interpolant} is used to compute geodesics between a cat and a dog representation. We compare the resulting paths to two baselines: (i) linear interpolation and (ii) spherical interpolation (slerp)~\citep{slerpkarpathy2022}, which is known to better preserve the structure of VAE latent spaces under Gaussian priors (see Supp.~\ref{app:spherical_interpolation}). Full experimental details are in Supp.~\ref{app:afhq_experiment}.
%We extend our Riemannian metric framework to a natural image domain using the Animal Faces High Quality (AFHQ) dataset~\citep{choi2020starganv2}. As in previous sections, all dataset images are projected into the latent space of the pretrained VAE from Stable Diffusion v1~\citep{rombach2022high}, which has a dimensionality of $4 \times 16 \times 16$. The Energy-Based Model (EBM) used here is trained to capture the distribution of VAE latent codes. We then apply Algorithm~\ref{algo:geodesic-interpolant} to learn approximate geodesics in latent space, interpolating between representations of a cat (start point) and a dog (end point). To evaluate the quality of the learned paths, we compare against two baselines: (i) linear interpolation in latent space, and (ii) spherical linear interpolation (slerp)\citep{slerpkarpathy2022}, which is known to better preserve on-manifold structure, particularly in latent spaces where the data distribution approximately lies on a hypersphere due to normalization or the Gaussian prior (see Supp.~\ref{app:spherical_interpolation} for further discussion). Full experimental details (datasets, architectures, and hyperparameters) are provided in Supp.~\ref{app:afhq_experiment}.

\begin{figure}[h!]
\begin{tikzpicture}
\draw [anchor=north west] (0.0\linewidth, 0.97\linewidth) node {\includegraphics[width=1\linewidth]{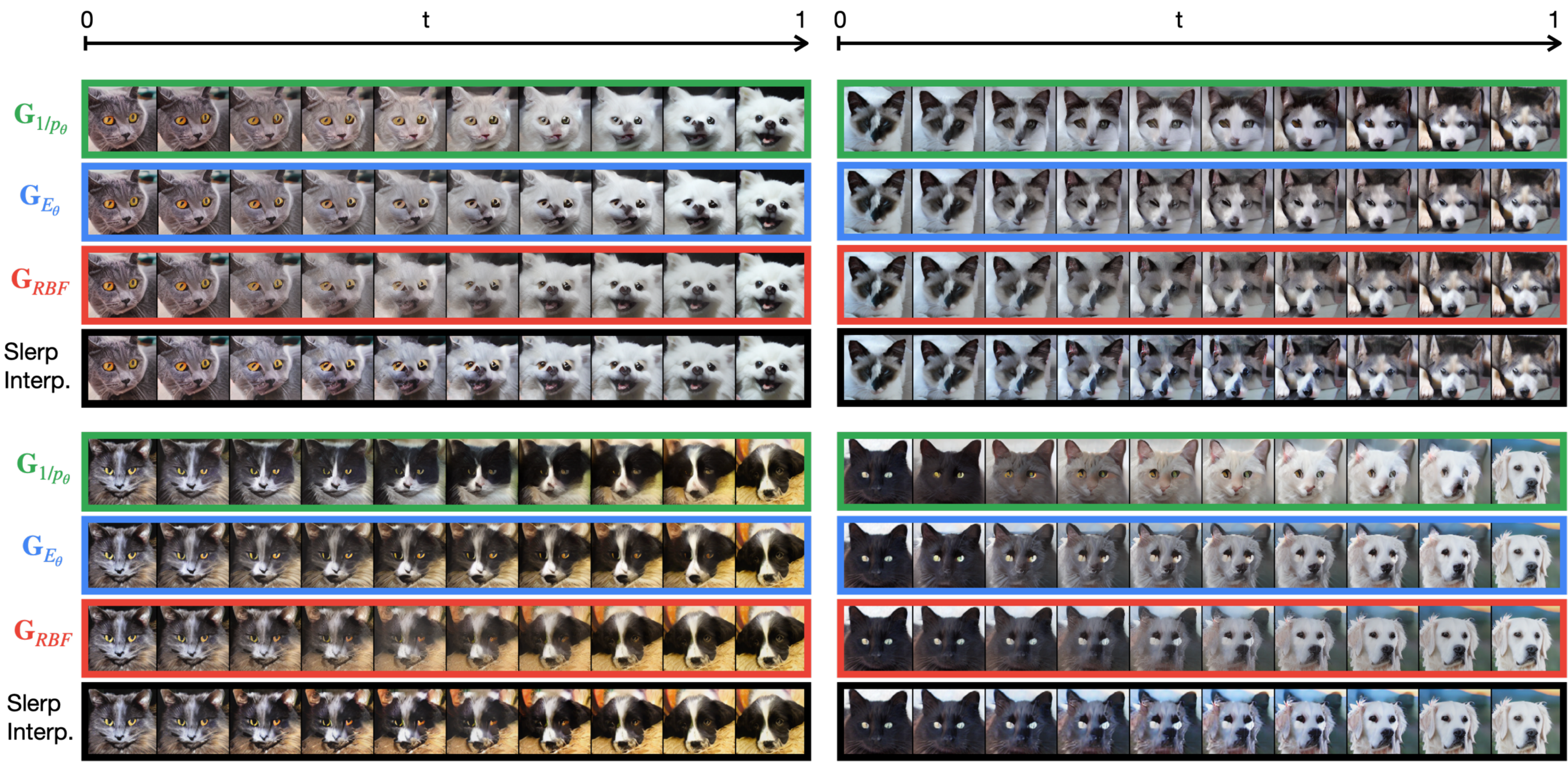}};
\end{tikzpicture}
\caption{{\bf Geodesics on the AFHQ dataset.} Each block shows an interpolated trajectory between two animal images (cats and dogs), projected back into image space for visualization. We compare geodesics computed with two EBM-based metrics (\invebm{G}, \logebm{G}), a parametric RBF-based metric (\rbf{G}), and spherical interpolation (slerp). Results using \landm{G}, linear interpolation, and additional examples are provided in Supp.~\ref{app:AFHQ_samples}.}
\label{fig:fig5}
\end{figure}

Fig.~\ref{fig:fig5} illustrates geodesics computed in the latent space of a pretrained VAE and projected back into image space (see~\ref{app:AFHQ_samples} for additional samples as well as samples for \landm{G} and linear interpolation). Qualitatively, we observe that geodesics computed with the \invebm{G} metric best adhere to the data manifold. The \logebm{G} metric also shows noticeable improvements over the other metrics. Despite extensive tuning, \rbf{G} and \landm{G} produce trajectories only slightly better than linear interpolation—suggesting these parametric metrics struggle to scale in high dimensions, consistent with prior findings~\citep{arvanitidis2016locally, arvanitidis2020geometrically}.

\begin{wraptable}{r}{0.4\textwidth}
\vspace{-4mm}
  \centering
  \begin{tabular}{cc}
    Metric & FID $(\downarrow)$\\
    \hline
    Linear interp. & 42.47 \\
    Slerp interp. & 32.67 \\
    \hdashline
    \logebm{G} & 20.79 \\
    \invebm{G} & {\bf 16.47} \\
    \hdashline
    \landm{G} & 39.17 \\
    \rbf{G} & 37.98 \\
  \end{tabular}
  \caption{{\bf FID along geodesics for different Riemannian metrics}. FID is computed at each trajectory point to assess on-manifold alignment. See Supp.~\ref{app:FID_error} for the $2$-$\sigma$ error.}
  \label{table:FID_AFHQ}
  \vspace{-5mm}
\end{wraptable}
To quantitatively assess geodesic quality, we report FID scores~\citep{heusel2017gans} in Table~\ref{table:FID_AFHQ}, computed over $50,000$ trajectories that interpolate from randomly chosen cat images to randomly chosen dog images. The results are consistent with qualitative observations: \invebm{G} and \logebm{G} yield the lowest FIDs, followed by the model-free slerp baseline, then \rbf{G}, \landm{G}, and linear interpolation. Note that the FID measures how aligned individual samples are with the training distribution—on-manifold alignment—but does not assess whether the full trajectory respects the true manifold curvature. Unfortunately, AFHQ lacks ground-truth geometry for such evaluation.

%To quantitatively assess geodesic quality, we report FID~\citep{heusel2017gans} scores in Table~\ref{table:FID_AFHQ}, computed over 50,000 interpolated trajectories. These results align with our qualitative observations: \invebm{G} and \logebm{G} outperform other metrics, followed by the model-free slerp baseline, then \rbf{G}, \landm{G}, and linear interpolation. We emphasize that the FID score measures the similarity of generated samples to the training distribution—capturing on-manifold realism—but does not evaluate whether trajectories faithfully follow the intrinsic geometry of the manifold. Unfortunately, the AFHQ dataset does not provide a ground truth for such geometric alignment.

\section{Conclusion}
In this work, we use pretrained Energy-Based Models (EBMs) to derive conformal Riemannian metrics, \logebm{G} and \invebm{G}, and we compare them to established alternatives (\landm{G}~\citep{arvanitidis2016locally} and \rbf{G}~\citep{arvanitidis2020geometrically}). On both synthetic and high-dimensional data, EBM-derived metrics yield geodesics that stay closer to the data manifold and better capture its curvature---especially with \logebm{G}.

We focus on conformal metrics, which scale the identity by a scalar field to encode density. \rev{While more complex, non-conformal and anisotropic metrics (e.g., the Stein metric~\citep{perone2024geometry}) are accessible from the EBM score, we found that conformal metrics offer comparable performance with simpler interpretation and reduced computational cost, justifying our focus in this work.} Future work may explore these extensions with regularization or structural priors to ensure smoothness and scalability (See Supp.\ref{app:limitations} for a discussion of limitations and Supp.\ref{app:broader_impact} for broader impact). \rev{To keep computational cost manageable, we train the EBM in the latent space of a pretrained autoencoder and compute geodesics using finite-difference optimization, two design choices that substantially reduce complexity and memory use without compromising performance.}

Although this article is primarily methodological, it points to promising applications. One example is the mental rotation task, in which humans mentally rotate objects to match a target~\cite{shepard1971mental}. In such experiments, reaction times tend to decrease with training~\cite{cooper1973chronometric}, suggesting that repeated exposure sharpens internal representations around training examples. These refined representations may concentrate in high-density regions, where mental transformations occur more quickly. As shown in Fig.~\ref{fig:curvature} and \ref{fig:rotation}, our geodesics naturally accelerate in such high-density regions, echoing these psychophysical findings. Modeling mental simulation as geodesics on Riemannian manifolds shaped by a generative model offers a principled computational framework to understand human cognition. It provides a way to formalize and test the hypothesis that the human cognition relies on generative models to support flexible inference~\citep{friston2006free,rao1999predictive,boutin2022pooling,boutin2022diversity,boutin2024latent}. \rev{Our approach is also particularly relevant for neuroscience, where datasets are high-dimensional, often sparsely sampled, and where understanding the geometry of neural population activity is central to scientific insight. In such settings, high-fidelity geodesics are essential for capturing the true structure of neural trajectories---approximations may distort the manifold and lead to misinterpretation of brain dynamics. While training EBMs is costly, the benefits in terms of interpretability and geometric accuracy make this approach compelling for applications where precision is critical.}

As machine learning models are increasingly used to capture complex data distributions, understanding the geometry of their latent spaces becomes essential. Our work contributes to this effort by showing that geometry can serve as a useful tool for building models that better reflect data structure, align with human perception, and shed light on cognitive processes.

\rev{
\section*{Acknowledgments}
This work was supported by ANR-3IA Artificial and Natural Intelligence Toulouse Institute (ANR-19-PI3A-0004). Part of this work was carried out within the DEEL project, which is part of IRT Saint Exupéry and the ANITI AI cluster. The authors acknowledge the financial support from DEEL’s industrial and academic members and the ``France 2030'' program (NR-10-AIRT-01 and ANR-23-IACL-0002). Additional support for TS provided by ONR (N00014-24-1-2026 and REPRISM MURI N00014-24-1-2603) and NSF (IIS-2402875).
}

\newpage
\newpage
\newpage
\bibliographystyle{unsrtnat}
\bibliography{mybib}

%%%%%%%%%%%%%%%%%%%%%%%%%%%%%%%%%%%%%%%%%%%%%%%%%%%%%%%%%%%%
\newpage
\appendix

\section{Extended related work}~\label{app:extended_related}

%%%%%%%%%%%%%%%%%%%%%%%%%%%%
%%%%% DO NOT MODIFY %%%%%%%%
%%%%%%%%%%%%%%%%%%%%%%%%%%%%

Several tools have been developed to study the geometrical properties of distributions. We survey some prominent approaches below.

\textbf{Information geometry}, initiated by the seminal works of~\citep{rao1992information,amari1983foundation}, was the first to apply ideas from differential geometry to the field of statistics. Unlike our present work, the goal was not to understand the geometry of the data $\bm{x}$, but rather to understand the geometry of a smooth manifold $\theta\in\Theta$ of parameters of an estimator $p_{\theta}$. In particular, starting from the Taylor expansion of \textit{reverse} Kullback-Leibler~\citep{kullback1951information} divergence to $\bm{p_{\theta}}$, in the neighborhood of $p_{\theta}$ itself, with $\theta'=\theta+\epsilon$, we get
\begin{equation}
    D_{KL}(p_{\theta'}\|\bm{p_{\theta}})\approx \underbrace{D_{KL}(p_{\theta}\|\bm{p_{\theta}})}_{=0}+\underbrace{\nabla_1 D_{KL}(p_{\theta}\|\bm{p_{\theta}})}_{=\bm{0}}\epsilon+\epsilon^T\nabla^2_1 D_{KL}(p_{\theta}\|\bm{p_{\theta}})\epsilon.
\end{equation}
One can show that, since $\theta'=\theta$ is the global minimum of this function, the first-order term vanishes and the second-order term $\nabla^2_1D_{KL}(p_{\theta}\|\bm{p_{\theta}})$ must be a positive definite form - i.e an inner product. This quantity, called \textit{Fisher information}~\citep{fisher1922mathematical}, gives $\Theta$ the structure of a Riemannian manifold.  
The Riemannian gradient associated with this manifold yields a second-order optimization method coined \textit{natural gradient descent}~\citep{amari1998natural}, that has been proved helpful in deep learning~\citep{pascanu2013revisiting}. Our method inherits some spirit of this approach, since we define a local inner product as a function of the density, to give a Riemannian structure to the data manifold. However, we focus on the geometry of the \textit{data} $\bm{x}$, not the geometry of the model's parameters $\bm{\theta}$.

\textbf{Riemannian structure of data manifolds} has already been proposed in the past. For example, the seminal LAND metric~\citep{arvanitidis2016locally} is a non-conformal metric built from the samples, with the intent of generalizing multivariate normal distributions to manifolds. The RBF metric~\citep{arvanitidis2020geometrically} is a conformal metric, derived from a kernel density estimator, with some learnable coefficients. More recently, ~\citet{kapusniak2025metric} proposed to use those metrics and learn a flow matching algorithm to fit geodesics in the data manifold. The Jacobian of a generative model also defines a metric~\citep{arvanitidis2018latent}.  The (unpublished) work of~\citet{perone2024geometry} has been inspirational for our contribution. They use the Stein score function to build the metric, an approach also chosen by~\cite{diepeveen2024score} - although restricted to unimodal densities.  

\textbf{Pullback geometry of latent manifolds} is an active research area. \cite{lange2023deep} studies the manifold of representations of a given network, while~\cite{sungeometry} builds a generative autoencoder to represent the manifold. Shortest paths are computed with fixed-point methods~\cite{arvanitidis2019fast}, or using a discrete graph~\citep{chen2019fast}. While we may rely on the latent space of a VAE for some challenging tasks, studying latent representations of a neural network is beyond the scope of our work.  

\textbf{On-manifold generative models} can be found in the literature. For example, we can mention flow and bridge matching approaches~\citep{lipman2022flow,de2023augmented}, which learn a flow between a source and a target distribution, including on Riemannian manifolds~\citep{chenflow2024,de2024pullback}. In particular, the Schr{\"o}dinger bridge~\citep{wang2021deep,shi2023diffusion} focuses on an optimization problem involving paths in the space of probability distributions, and was also generalized to non-Euclidean geometries~\citep{de2022riemannian,thornton2022riemannian}. These works differ significantly from ours: they assume the Riemannian manifold to be given, not chosen, and they build a generative model on top of it. To the contrary, given a special class of generative models to represent the data, we \textit{choose} the metric to build the manifold.  

\textbf{Topological data analysis}~\citep{carlsson2009topology,zomorodian2012topological} studies the topological properties of the data manifold. This field aims to estimate some topological invariants such as the Euler characteristic~\citep{hacquard2024euler} and persistent Betti numbers~\citep{bubenik2015statistical} (which are the number of connected components, number of closed loops, etc.) from a finite sample. It relies on tools such as persistent homology~\citep{zomorodian2004computing,edelsbrunner2013persistent,otter2017roadmap} to design algorithms. This approach typically focuses on the \textit{global} properties: it assumes that the data accumulate on a well-defined manifold, from which these high-level features must be computed. To the contrary, our approach focuses on the \textit{local} structure defined by the metric, while the global structure is inherited from the induced geodesics. Furthermore, we consider the whole ambient space for our manifold, tweaking only the metric to account for low-density regions.  

\textbf{Symmetries and geometry in representations} have gathered considerable attention from the deep learning community, warranting no fewer than 3 workshops at Neurips alone~\footnote{\url{https://www.neurreps.org}}. Symmetries are operations under which a structure is left invariant, or equivariant. In particular, some neural architectures are leveraged to reflect priors about the underlying symmetries of the data~\citep{cohen2016group,cohen2017steerable,cohen2019gauge,finzi2020generalizing,satorras2021n}.  
In other cases, symmetries are discovered and learned from observations~\citep{dangovskiequivariant,rommel2022deep,yang2023generative}.  
Unlike these approaches, we do not seek symmetries in data, and we make minimal assumptions about the model; we are mainly interested in the density to build the structure.  

\textbf{Non-Euclidean 2D and 3D manifolds} are first-class citizens in computer graphics. The works of~\citep{lichtenstein2019deep,zhang2023neurogf} define a way to find shortest paths over such manifolds. However, this requires solving the Eikonal equation, which is prohibitively expensive in high dimensions or restricted to Euclidean geometries~\citep{bethune2023robust}. Geodesics can be learned, but this is restricted to low dimensions~\citep{zhang2023neurogf}. These setups are beyond the scope of our work, as we focus on higher-dimensional and sparsely populated spaces, and no discrete meshes can be built from samples.    

\textbf{Metric learning} (or \textit{distance learning}) is another field whose purpose is to learn a distance function between samples, typically in a weakly-supervised manner with contrastive losses~\citep{chopra2005learning,schroff2015facenet,sohn2016improved}. Often, these distances cannot be realized as a geodesic distance and are intended for a specific task, like classification or retrieval.

\section{Energy-Based Model}~\label{app:EBM}

\subsection{Derivation of the Gradient of the EBM Log-Likelihood}\label{app:EBM_ML_derivation}
The demonstration below is adapted from~\cite{woodford2006notes} to fit our notation. Even though this mathematical derivation is not crucial for a good understanding of our work, we include
it to make sure our article is self-contained and complete.

We consider an Energy-Based Model (EBM) defining a probability distribution via the Boltzmann form:
\begin{equation}
p_{\theta}(\v{x}) = \frac{\exp(-E_{\theta}(\v{x}))}{Z(\theta)} \quad \text{with} \quad Z(\theta) = \int \exp(-E_{\theta}(\v{x})) \, d\v{x}. \nonumber
\end{equation}

Our goal is to minimize the negative log-likelihood with respect to the empirical data distribution $p_{\mathcal{D}}$:
\begin{equation}
\mathcal{L}_{\text{ML}}(\theta) = \mathbb{E}_{\v{x} \sim p_{\mathcal{D}}}[-\log p_{\theta}(\v{x})]. \nonumber
\end{equation}

We first expand the log-probability:
\begin{equation}
-\log p_{\theta}(\v{x}) = E_{\theta}(\v{x}) + \log Z(\theta). \nonumber
\end{equation}

Taking the gradient with respect to $\theta$:
\begin{equation}
\nabla_{\theta} \mathcal{L}_{\text{ML}} = \mathbb{E}_{\v{x} \sim p_{\mathcal{D}}} \left[ \nabla_{\theta} E_{\theta}(\v{x}) + \nabla_{\theta} \log Z(\theta) \right]. \nonumber
\end{equation}

The derivative of the log-partition function could be simplified:
\begin{align}
\nabla_{\theta} \log Z(\theta) 
&= \frac{1}{Z(\theta)} \nabla_{\theta} Z(\theta) \nonumber \\
&= \frac{1}{Z(\theta)} \nabla_{\theta} \int \exp(-E_{\theta}(\v{x})) \, d\v{x} \nonumber \\
&= - \frac{1}{Z(\theta)} \int \exp(-E_{\theta}(\v{x})) \nabla_{\theta} E_{\theta}(\v{x}) \, d\v{x} \nonumber \\
&= - \int p_{\theta}(\v{x}) \nabla_{\theta} E_{\theta}(\v{x}) \, d\v{x} \nonumber \\
&= - \mathbb{E}_{\v{x} \sim p_{\theta}} \left[ \nabla_{\theta} E_{\theta}(\v{x}) \right]. \nonumber
\end{align}

Substituting this back into the gradient of the loss:
\begin{equation}
\nabla_{\theta} \mathcal{L}_{\text{ML}} = \mathbb{E}_{\v{x} \sim p_{\mathcal{D}}} \left[ \nabla_{\theta} E_{\theta}(\v{x}) \right] - \mathbb{E}_{\v{x} \sim p_{\theta}} \left[ \nabla_{\theta} E_{\theta}(\v{x}) \right]. \nonumber
\end{equation}

In practice, we denote the $\v{x}^{+}$ the "positive" samples from the empirical data distribution $p_{D}$, and $\v{x}^{-}$ the "negative" samples from the model:
\begin{equation}
\nabla_{\theta} \mathcal{L}_{\text{ML}} \approx \mathbb{E}_{\v{x}^+ \sim p_{\mathcal{D}}} \left[ \nabla_{\theta} E_{\theta}(\v{x}^+) \right] - \mathbb{E}_{\v{x}^- \sim p_{\theta}} \left[ \nabla_{\theta} E_{\theta}(\v{x}^-) \right]. \nonumber
\label{eq:ml_ebm_derivation}
\end{equation}

\subsection{EBM training algorithm}\label{app:EBM_algorithm}

To train our Energy-Based Models (EBMs), we follow the approach of~\cite{du2019implicit}. Algo.~\ref{algo:train-ebm} details the general training procedure: 

\begin{algorithm}[H]
\SetAlgoLined
\KwIn{Training dataset :$\mathcal{D}$, learning rate $\eta$, Replay Buffer $\mathcal{B}$, Langevin step size $\alpha$, noise scale $\sigma$, number of Langevin steps $L$}
\While{\textnormal{Training}}{
    $\v{x}^+ \sim \mathcal{D}$ \,\,\,\textcolor{gray}{sample from the dataset}\\
    $\v{x}^{0} \sim \mathcal{B}$ \,\,\,\textcolor{gray}{\# sample from a replay buffer with probability $95\%$}\\
    \textcolor{gray}{ \#\# Refine negative samples using Langevin dynamics} \\
    \For{$t \gets 1$ \KwTo $L$} {
    $\v{x}^{t+1} \leftarrow \v{x}^{t} - \alpha \nabla_{\v{x}^{t}}E_{\theta}(\v{x}^{t}) + \omega$ \,\,\, with $\omega \sim \mathcal{N}(0,\sigma)$ \\
    }
    $\v{x}^{-} = \v{x}^{L}\textnormal{.detach()}$ \\
    
    $\displaystyle \nabla_\theta \mathcal{L}_{\text{ML}} \approx \mathbb{E}_{\textnormal{Batch}} \left[\nabla_\theta E_\theta(\v{x}^+_i) - \nabla_\theta E_\theta(\v{x}^-_i)\right]$ \textcolor{gray}{ \#\# Compute the ML loss} \\
    $ \displaystyle \mathcal{L}_{REG}(\theta) = \mathbb{E}_{\textnormal{Batch}} \left[\nabla_\theta E_\theta(\v{x}^+_i)^{2} + \nabla_\theta E_\theta(\v{x}^-_i)^{2}\right]$ \textcolor{gray}{ \#\# Compute Regularization loss}\\
    $\theta \leftarrow \theta - \eta \nabla_\theta \mathcal{L}_{\text{ML}} - \eta \nabla_\theta \mathcal{L}_{REG}$\textcolor{gray}{ \#\# update parameters with gradient descent}\\
    $\mathcal{B} \leftarrow \mathcal{B} \cup \v{x}^{+}$
}
\caption{Training Energy-Based Model using Langevin Dynamics}
\label{algo:train-ebm}
\end{algorithm}

In all experiments, we use $L = 100$ Langevin steps with step size $\alpha = 1$ and noise scale $\sigma = 10^{-2}$. The energy function is optimized using the Adam optimizer~\citep{kingma2014adam} with a learning rate of $\eta = 10^{-4}$. In addition to the maximum likelihood (ML) loss, we include a regularization term that encourages the energy values to remain close to zero, a technique shown to be effective in prior work~\citep{du2019implicit}.

We observed that training can be unstable, particularly for high-dimensional datasets. We attribute this instability to the lack of gradient supervision: the loss is not backpropagated through the Langevin dynamics to reduce memory usage. To mitigate this, we introduce a small Denoising Score Matching (DSM) loss—only for the AFHQ dataset—which provides weak supervision of the energy gradient. This additional regularization loss is similar to the DSM loss in~\citep{li2023learning}. We found this trick to strongly improve stability without degrading the performance.

The energy network architecture is adapted to the complexity of each dataset. Full details are provided in Appendix~\ref{app:ebm_gaussian}, \ref{app:ebm_alphanum}, and \ref{app:ebm_afhq}. Following~\citet{li2023learning}, we design the output layer of the energy function to take a quadratic form.

\subsection{Other training procedure in literature}

EBM can also be trained by minimizing the so-called \textit{Stein discrepancy}~\citep{grathwohl2020learning}, Denoising Score Matching~\citep{vincent2011connection}, Sliced Score Matching~\citep{song2020sliced}, Noise Contrastive Estimation~\citep{gutmann2010noise}. A related objective to contrastive divergence is \textit{energy discrepancy}~\citep{schroder2023energy}.  We refer the reader~\cite{song2021train}, for a complete review of the different methods to train EBMs.  

\newpage
\section{Riemannian Metrics}
\subsection{Calibration}\label{app:metric_normalization}
We normalize each metric using calibration coefficients $\alpha$ and $\beta$, with two goals: (i) ensuring that the Riemannian metric averages to the identity matrix $\v{I}$ on the manifold, and (ii) aligning the overall scale of all metrics to allow fair comparisons. Here are more details on the calibration procedure:

First, we randomly sample data pairs $(\v{x}_0, \v{x}_1)$ from the dataset $\mathcal{D}$ (it corresponds to the geodesics endpoints) and generate linear interpolations between them using:
\begin{equation}
\v{x}_t = (1 - t) \v{x}_0 + t \v{x}_1
\label{app:eq_linear_interpolation}
\end{equation}

Second, we define two sets of samples: $\mathcal{S}_{\mathcal{M}}$, which contains the endpoints $\v{x}_0$ and $\v{x}1$ \textit{lying on the data manifold}, and $\mathcal{S}_{\bar{\mathcal{M}}}$, which contains the midpoints at $t = \frac{1}{2}$. These sets are then used to estimate the calibration coefficients $\alpha$ and $\beta$:
\[
\begin{aligned}
\displaystyle
\v{G}(\v{x})=\alpha \, \mathbf{h}(\boldsymbol{x}) + \beta \quad \textnormal{s.t}\,\,\,
&\begin{cases}
\displaystyle \alpha = \frac{g_{\max} - g_{\min}}{\frac{1}{|\mathcal{S}_{\bar{\mathcal{M}}}|}\sum_{\v{x}\in\mathcal{S}_{\bar{\mathcal{M}}}}h(\v{x}) - \frac{1}{|\mathcal{S}_{\mathcal{M}}|}\sum_{\v{x}\in\mathcal{S}_{\mathcal{M}}}h(\v{x})} \\
\beta = g_{\min} - \alpha \cdot \frac{1}{|\mathcal{S}_{\mathcal{M}}|}\sum_{\v{x}\in\mathcal{S}_{\mathcal{M}}}h(\v{x})
\end{cases} \\[1em]
\v{G}(\v{x})=(\alpha \, \mathbf{h}(\boldsymbol{x}) + \beta)^{-1} \quad \textnormal{s.t}\,\,\,
&\begin{cases}
\alpha = \dfrac{1/g_{\max} - 1/g_{\min}}{\frac{1}{|\mathcal{S}_{\bar{\mathcal{M}}}|}\sum_{\v{x}\in\mathcal{S}_{\bar{\mathcal{M}}}}h(\v{x}) - \frac{1}{|\mathcal{S}_{\mathcal{M}}|}\sum_{\v{x}\in\mathcal{S}_{\mathcal{M}}}h(\v{x})} \\
\beta = \dfrac{1}{g_{\min}} - \alpha \cdot \frac{1}{|\mathcal{S}_{\mathcal{M}}|}\sum_{\v{x}\in\mathcal{S}_{\mathcal{M}}}h(\v{x})
\end{cases}
\end{aligned}
\]
This calibration strategy adjusts the metric based on both on-manifold and off-manifold regions. It ensures that all metrics operate within a comparable dynamic range and promotes a useful geometric prior: lower metric values near the data manifold and higher values farther away. As a result, geodesics are encouraged to stay close to high-density areas, aligning the geometry with the data distribution.

\subsection{LAND metric}\label{app:land}
\begin{figure}[h!]%{0.6\textwidth}
  \centering
  \includegraphics[width=0.6\textwidth]{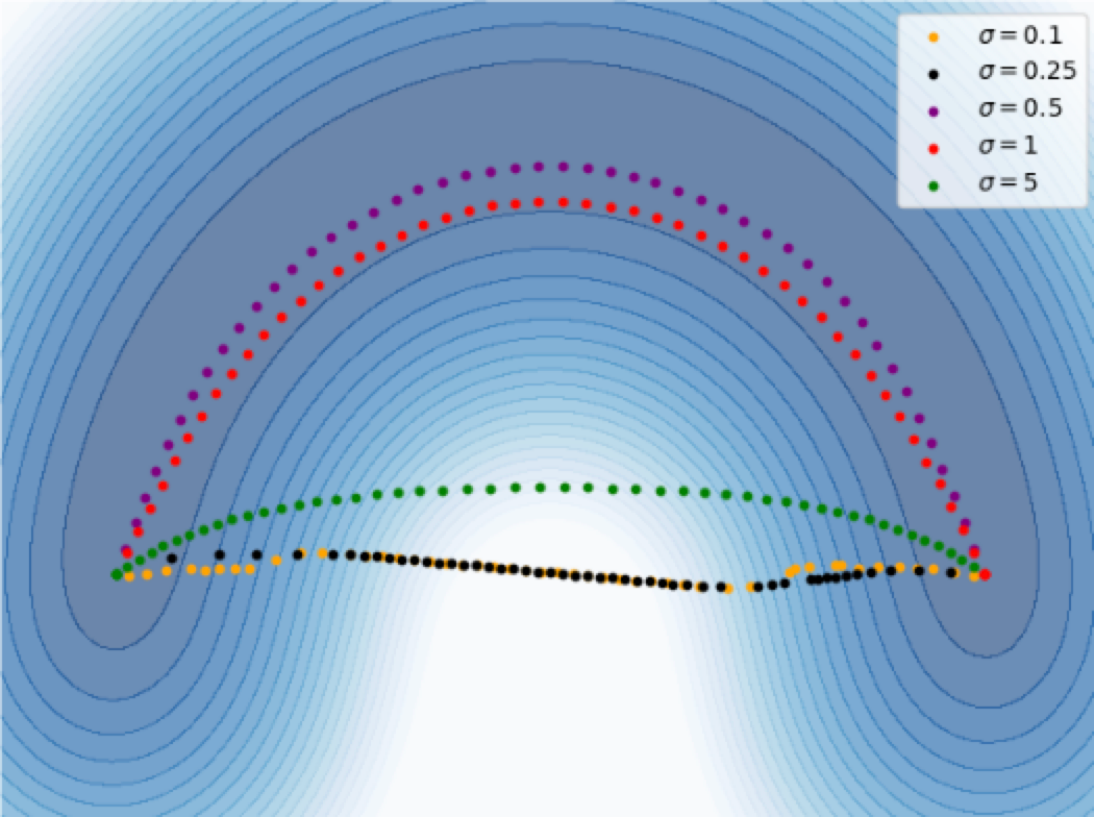}
  \caption{{\bf Effect of the bandwidth $\sigma$} on the geodesics obtained with the LAND metric. Here we have explored five $\sigma$ values ($\sigma \in \{0.1, 0.25, 0.5, 1, 5\}$). We observed that $\sigma$ has a major impact on the shape of the geodesics.}
  \label{app:fig_sigma_land}
\end{figure}

We remind the land metric formula (see Eq.~\ref{eq:LAND}):
\begin{equation}
  \displaystyle \textcolor{google_orange}{\mathbf{G}_{\text{LAND}}(\v{x})} = (\alpha\, \textnormal{diag}(\v{h}(\v{x})) + \beta \mathbf{I})^{-1}\,\,\textnormal{s.t}\,\,h^{(j)}(\v{x})= \sum_{i=1}^{N}(x_i^{(j)}\!-\!x^{(j)})^{2}\exp{\bigg(\!\!\!-\frac{||\v{x}\!-\!\v{x_i} ||^{2}}{2\sigma^{2}}\bigg)}
\end{equation}

This metric is highly sensitive to the choice of the $\sigma$ parameter, which controls the "locality" of the metric. A small $\sigma$ results in a very local metric that is strongly influenced by nearby points, while a large $\sigma$ smooths the metric by averaging over a wider region. This directly affects the trade-off between how closely geodesics follow the data manifold and how smooth or stable they are. In practice, we observe that $\sigma$ has a major impact on the shape of the geodesics, as shown in Fig.\ref{app:fig_sigma_land}, confirming earlier findings by\cite{arvanitidis2016locally}. To illustrate this, we plot geodesics for five different values of $\sigma$ ($\sigma \in \{0.1, 0.25, 0.5, 1, 5\}$) and find that they closely follow the data manifold only within a narrow range, particularly around $\sigma=0.5$.

\subsection{RBF metric}\label{app:rbf}
We first remind the RBF formula : 
\[
\textcolor{google_red}{\mathbf{G}_{\text{RBF}}(\v{x})} = (\alpha \cdot h(\v{x}) + \beta)^{-1} \cdot \mathbf{I}, \quad
h(\v{x}) = \sum_{k=1}^{K} w_k \exp\left(-0.5 \cdot \lambda_k \|\v{x} - \v{\hat{x}}_k\|^2\right).
\]
In the equation, the $\{\hat{\v{x}}\}_{i=1}^{K}$ are centroids evaluated using a K-Means algorithm. Following~\cite{arvanitidis2020geometrically}, the bandwidth ($\lambda_k$) using the inter-distance to prototype (see Eq.~\ref{app:eq_bandwidth}):
\begin{equation}
\displaystyle \lambda_k= \frac{1}{2} \bigg(\frac{\kappa}{2K}\sum_{k=1}^{K}|| \v{x} - \v{\hat{x}_k}||\bigg)^{-2}
\label{app:eq_bandwidth}
\end{equation}
The bandwidth, $\lambda_k$, controls the spatial extent of each radial basis function. In Eq.~\ref{app:eq_bandwidth}, $\kappa$ is a tunable hyperparameter controlling how concentrated or spread out the RBFs are. Intuitively, a larger $\kappa$ results in narrower kernels (stronger locality) while a smaller one yields wider coverage. This trade-off is explored via hyperparameter search. The weights $w_k$ modulate the relative contribution of each RBF to the resulting scalar field. These weights are optimized to ensure that $h(x)$ remains close to 1 on the training data, using the following loss:
\begin{equation}
\mathcal{L}(\v{w})=\sum_{n=1}^{N} ||1 - h(\v{x_i})||^{2}
\end{equation}
This encourages the RBF combination to approximate a constant value (here, 1) across the data distribution, ensuring consistency and stability of the field on the manifold.

\begin{figure}[h!]%{0.6\textwidth}
  \centering
  \includegraphics[width=0.6\textwidth]{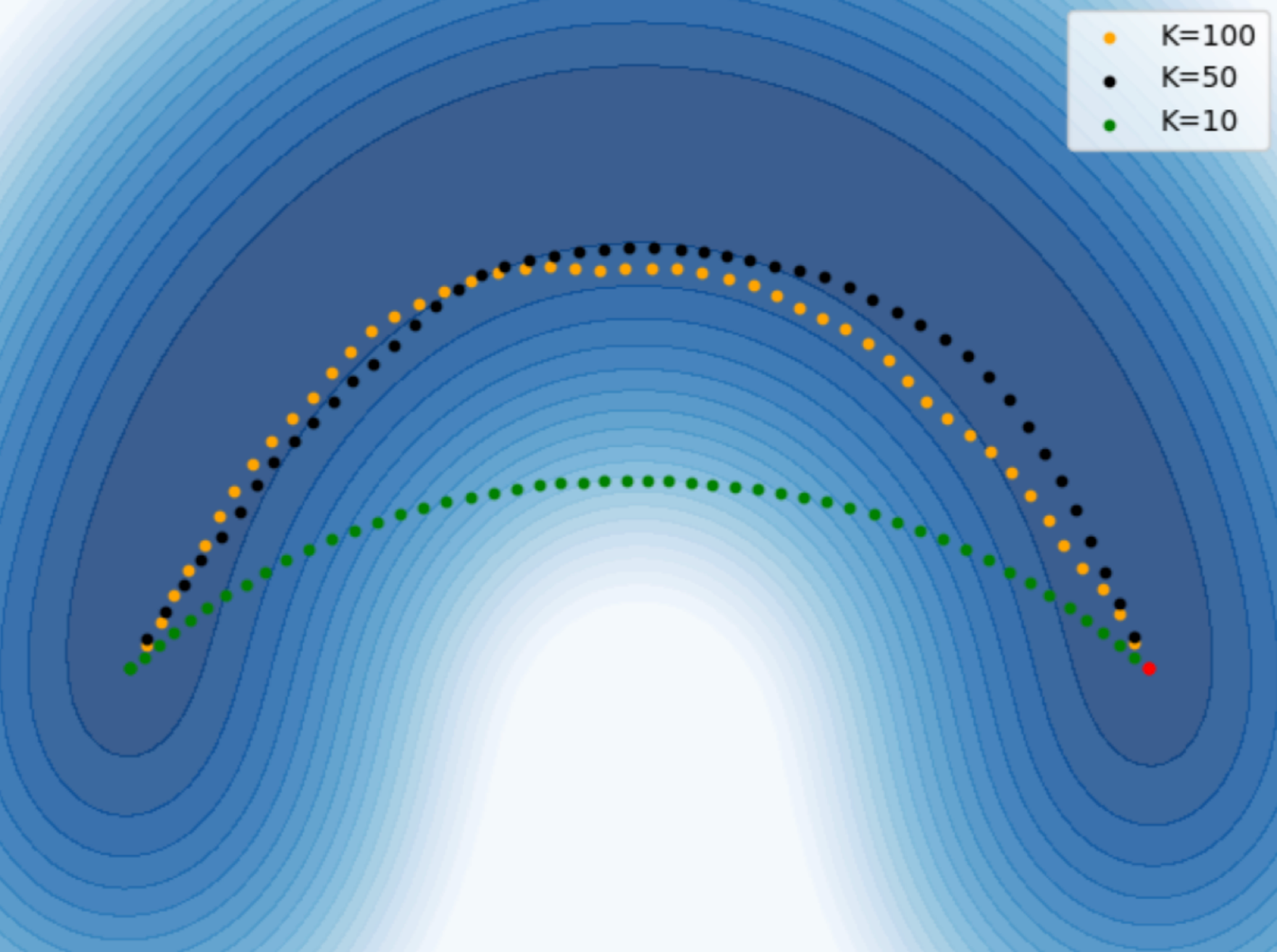}
  \caption{{\bf Effect of the number of centroids $K$} on the geodesics obtained with the RBF metric ($K \in \{10, 50, 100\}$). We observed that $K$ has a major impact on the shape of the geodesics.}
  \label{app:fig_K_RBF}
\end{figure}

In Fig.~\ref{app:fig_K_RBF}, we evaluate how the number of centroids $K$ affects the shape of the geodesics. The results show that geodesics are highly sensitive to this parameter. When $K$ is too small, the geodesics fail to follow the data manifold accurately. Conversely, when $K$ is too large, the trajectories become overly sinuous—passing through many centroids that are not necessarily aligned with the true manifold.

\newpage
\section{Experimental details on the Circular Mixture of Gaussian datasets}\label{app:circular_dataset}
\subsection{Datasets}\label{app:circular_dataset_det}

To design our toy datasets, we have used a mixture of K (2D) Gaussians. Specifically, $K=200$ in all our datasets. The resulting probability distribution is therefore:
\begin{equation}
p(\boldsymbol{x}) = \sum_{k=1}^{K} \pi_k \, \mathcal{N}(\boldsymbol{x} \mid \boldsymbol{\mu}_k, \mathbf{I}),
\end{equation}
where \( \mathcal{N}(\boldsymbol{x} \mid \boldsymbol{\mu}_k, \mathbf{I}) \) denotes a 2D isotropic Gaussian centered at \( \boldsymbol{\mu}_k \). Here, $\mathbf{I}$ is the identity matrix of size $2\times2$. In both datasets, the centers of the Gaussians are uniformly positioned along a semi-circle or Radius $R$ (here $R=8$). Specifically, the centers are given by:
\begin{equation}
\boldsymbol{\mu}_k = R \cdot 
\begin{bmatrix}
\cos\left(\theta_k\right) \\
\sin\left(\theta_k\right)
\end{bmatrix}
\quad \text{with} \quad
\theta_k = \frac{k}{K} \cdot \pi, \quad k = 0, \dots, K-1.
\end{equation}
The only difference between the Uniform Circular Gaussian (UCG) dataset and the Weighted Circular Gaussian dataset (WCG) is the weighting coefficient $\{\pi_k\}_{k=1}^{K}$

\begin{wrapfigure}{r}{0.5\textwidth}
\vspace{-3mm}
  \centering
  \includegraphics[width=0.5\textwidth]{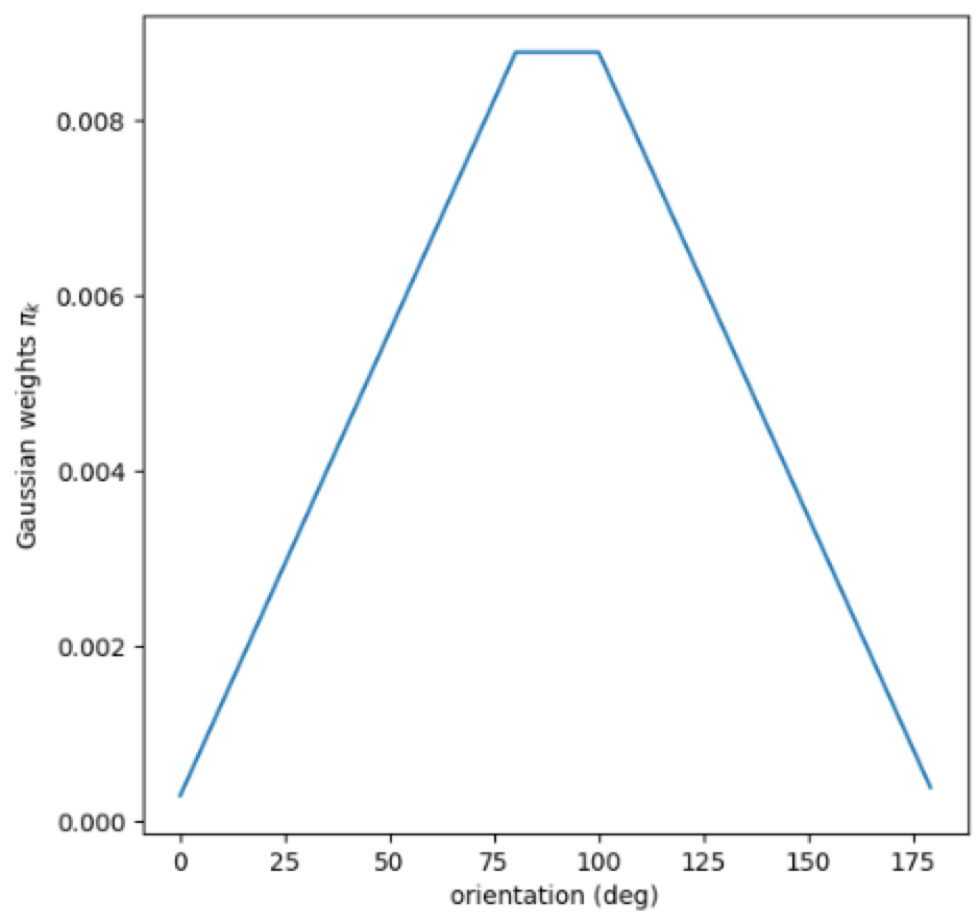}
  \caption{{\bf Profile of the Gaussians weights $\pi_k$}}
  \label{app:fig_weight_profile_WCG}
\vspace{-3mm}
\end{wrapfigure}
\paragraph{Uniform Circular Gaussian dataset.} Here all the weights are similar and equal to $1/K$. As a result, the energy landscape forms a semi-circular basin with constant depth (see contour plot of Fig.~\ref{fig:fig1}a for an illustration of the energy landscape).

\paragraph{Weighted Circular Gaussian dataset.}
In this setting, the mixture weights vary, concentrating the distribution toward the center of the arc. The weights are symmetric with respect to the horizontal axis, producing an energy landscape with a semi-circular shape and slopes symmetric around the arc’s midpoint (see the contour plot in Fig.\ref{fig:fig1}c). Fig.\ref{app:fig_weight_profile_WCG} shows the weights $\pi_k$ as a function of orientation, with all weights summing to 1. This setup generates a curved, non-uniform density with higher mass near the center of curvature (i.e., at 90 degrees), allowing us to introduce a controlled curvature in the data manifold and assess how well different metrics capture it.

\subsection{Neural networks architectures and Hyperparameters on the Circular Mixture of Gaussian Dataset}\label{app:ebm_gaussian}

Here, we describe the architecture of the energy function (see Table~\ref{tab:mlp_ucg_energy_architecture}), the interpolant network (see Table~\ref{tab:mlp_ucg_interpolant_architecture}), and the hyperparameters used for the \landm{G} and \rbf{G} metrics. Note that the architectures and settings are the same for both the UCG and WCG datasets.

\begin{figure}[h!]
\begin{tikzpicture}
\draw [anchor=north west] (0\linewidth, 0.98\linewidth) node {\includegraphics[width=1\linewidth]{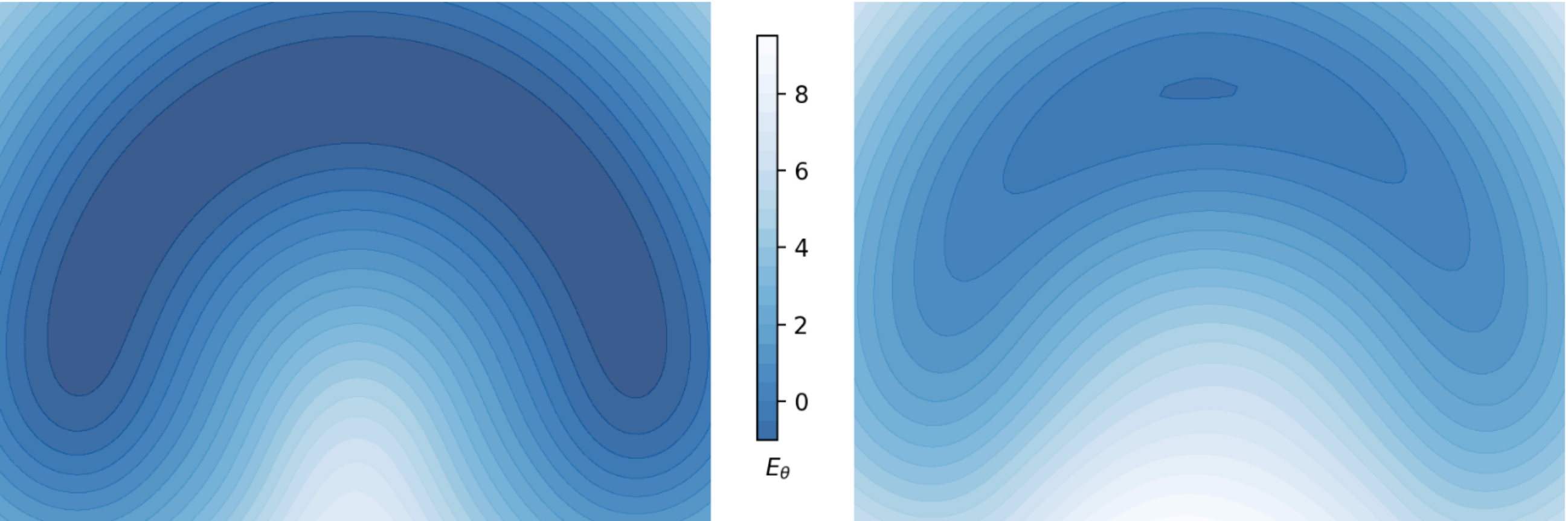}};
\draw [anchor=north west] (0\linewidth, 0.60\linewidth) node {\includegraphics[width=1\linewidth]{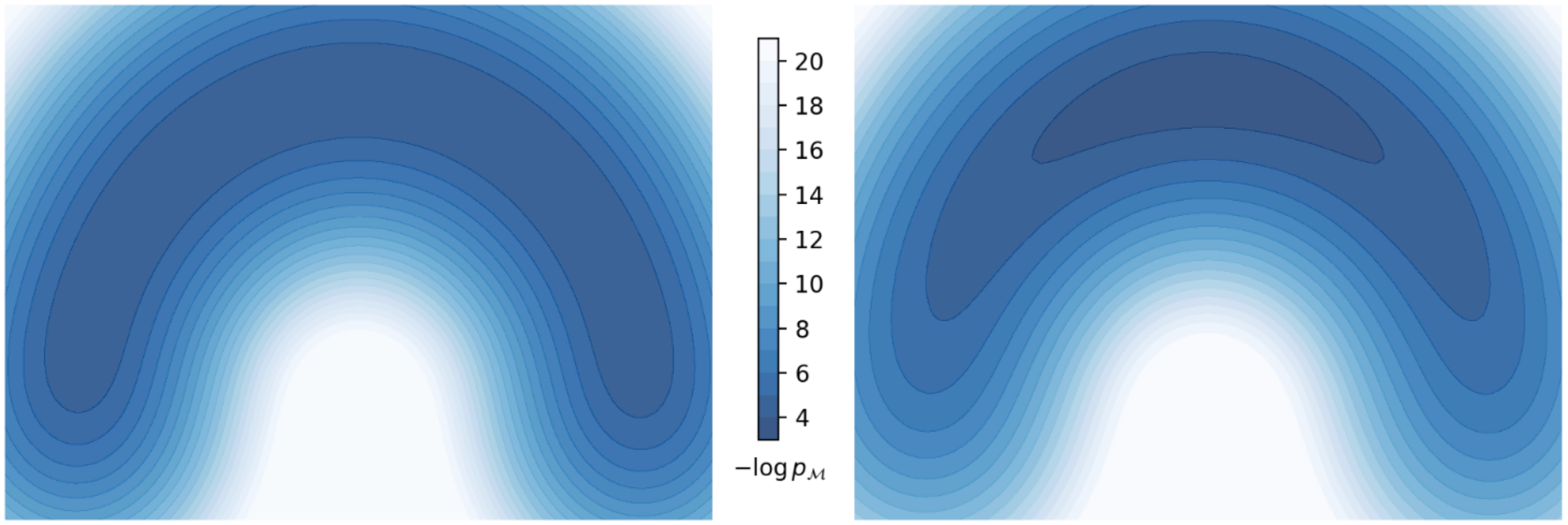}};

\begin{scope}
    \draw [anchor=north west,fill=white, align=left] (0.0\linewidth, 1\linewidth) node {{\bf a)} $E_{\theta}$ on the UCG dataset} ;
    
    \draw [anchor=north west,fill=white, align=left] (0.5\linewidth, 1\linewidth) node {{\bf b)} $E_{\theta}$ on the WCG dataset};

    \draw [anchor=north west,fill=white, align=left] (0.0\linewidth, 0.62\linewidth) node {{\bf c)} $-\log p_{\mathcal{M}}$ on the UCG dataset} ;
    
    \draw [anchor=north west,fill=white, align=left] (0.5\linewidth, 0.62\linewidth) node {{\bf d)} $-\log p_{\mathcal{M}}$ on the WCG dataset};
\end{scope}
\end{tikzpicture}
%\vspace{-7mm}
\caption{\textbf{Energy Landscape on the UCG and WCG datasets.} {\bf (a, b)} shows the energy landscape learned by the EBMs on the UCG and WCG datasets, respectively. {\bf (c, d)} shows the true energy landscape (i.e., $-\log p_{\mathcal{M}}$) on the UCG and WCG datasets, respectively.}
\label{app:fig_ebm_landcape_toy}
%\vspace{-4mm}
\end{figure}

\paragraph{Energy-Based Model}
Table~\ref{tab:mlp_ucg_energy_architecture} summarizes the architecture used for the energy function of the EBM. The output is designed to follow a quadratic form, similar to the approach in~\cite{li2023learning}, which we found improves performance across all datasets. To assess whether the EBM successfully learns the target distribution, we visualize the learned energy landscapes for both the UCG and WCG datasets (see Fig.~\ref{app:fig_ebm_landcape_toy}a and Fig.~\ref{app:fig_ebm_landcape_toy}b, respectively). For reference, we also include the ground-truth energy landscapes of the target distributions (see Fig.~\ref{app:fig_ebm_landcape_toy}c and Fig.~\ref{app:fig_ebm_landcape_toy}d for UCG and WCG, respectively). We observe that the EBM accurately captures the overall shape of the energy landscape for both distributions. However, in the WCG dataset, the true energy spans a broader range than the EBM's learned energy. This discrepancy is partially corrected by the normalization procedure described in Appendix~\ref{app:metric_normalization}.

\paragraph{Interpolant Network} Table~\ref{tab:mlp_ucg_interpolant_architecture} summarizes the architecture used for the interpolant network (i.e., $\boldsymbol{\varphi}_{t, \eta}$ in Algo.~\ref{algo:geodesic-interpolant} and Eq.~\ref{Eq:geodesic_parametrization}). For all datasets, we use an autoencoder-like architecture for the interpolant, following a similar approach to~\cite{kapusniak2025metric}.

\begin{table}[h]
\centering
\begin{minipage}{0.48\textwidth}
\centering
\begin{tabular}{|c|c|}
\hline
\textbf{Nb. Layers}
 & \textbf{Layer type} \\
\hline
\multirow{2}{*}{1} & Linear (2, 32) \\
 & SiLU \\
\hline
\multirow{2}{*}{4} & Linear (32, 32) \\
 & SiLU \\
\hline
1 & Linear (32, 32) \\
\hline
 \multirow{4}{*}{1} & Three output heads: \\
  & \quad Linear (32, 1) for $f_1$ \\
  & \quad Linear (32, 1) for $f_2$ \\
  & \quad Linear (32, 1) for $f_3$ \\
\hline
 output & $f_1(x) \cdot f_2(x) + f_3(x^2)$ \\
\hline
\end{tabular}
\caption{MLP architecture of the energy function on both UCG and WCG datasets.}
\label{tab:mlp_ucg_energy_architecture}
\end{minipage}%
\hfill
\begin{minipage}[c]{0.48\textwidth}
\centering
\begin{tabular}{|c|c|}
\hline
\textbf{NB. Layers} & \textbf{Layer type} \\
\hline
\multirow{2}{*}{1} & Linear (3, 32) \\
 & SiLU \\
\hline
\multirow{2}{*}{1} & Linear (32, 64) \\
 & SiLU \\
 \hline
\multirow{2}{*}{1} & Linear (64, 64) \\
 & SiLU \\
 \hline
\multirow{2}{*}{1} & Linear (64, 32) \\
 & SiLU \\
 \hline
1 & Linear (32, 3) \\
 \hline
\end{tabular}
\caption{MLP architecture of the interpolant network $\v{\varphi}_{t,\eta}$for WCG dataset.}
\label{tab:mlp_ucg_interpolant_architecture}
\end{minipage}
\end{table}

\paragraph{LAND metric} We performed a hyperparameter search to tune the $\sigma$ parameter. We found that $\sigma = 1$ yielded the best performance. Parameters are similar for both UCG and WCG.

\paragraph{RBF metric} We conducted a hyperparameter search to tune both the number of centroids $K$ and the scaling factor $\kappa$. The best results were obtained with $K = 30$ and $\kappa = 1$. Parameters are similar for both UCG and WCG.

\subsection{Quantitative evaluation with error bars}\label{app:toy_2sig}

In Fig.~\ref{fig:fig1_2sig}, we report the same quantitative results as in Fig.~\ref{fig:fig1}, now including $2\text{-}\sigma$ error bars. The standard deviation $\sigma$ is computed over evaluation metrics, each averaged on a different set of randomly sampled trajectories (five sets in total).

\begin{figure}[h!]
\begin{tikzpicture}
\draw [anchor=north west] (0.\linewidth, 0.96\linewidth) node {
\begin{tabular}{l c c c }
\makecell[c]{Metric} & 
\makecell[c]{$p_{\mathcal{M}}(\v{\gamma}^{\star})$\\
$(\uparrow)$} & 
\makecell[c]{RMSE \\
$(\downarrow)$} \\
\hline
$\boldsymbol{G}_{E_{\mathcal{M}}}$ & $0.79 \pm 0.02$ & -\\
$\boldsymbol{G}_{1/p_{\mathcal{M}}}$ & $0.77 \pm 0.04$ & -\\
\hdashline
\logebm{G} & $\mathbf{0.78} \pm 0.03$ & $0.12 \pm 0.02$\\
\invebm{G} & $0.73 \pm 0.01$ & $\mathbf{0.10} \pm 0.03$\\
\hdashline
\landm{G} & $0.60 \pm 0.07$ & $0.38 \pm 0.05$\\
\rbf{G} & $0.61 \pm 0.06$ & $0.39 \pm 0.1$\\
\end{tabular}
};

\draw [anchor=north west] (0.5\linewidth, 0.96\linewidth) node {
\begin{tabular}{l c c c }
\makecell[c]{Metric} & 
\makecell[c]{$p_{\mathcal{M}}(\v{\gamma}^{\star})$\\
$(\uparrow)$} & 
\makecell[c]{RMSE \\
$(\downarrow)$} \\
\hline
$\boldsymbol{G}_{E_{\mathcal{M}}}$ & $0.67\pm 0.05$ & -\\
$\boldsymbol{G}_{1/p_{\mathcal{M}}}$ & $0.73\pm 0.07$ & -\\
\hdashline
\logebm{G} & $\mathbf{0.67}\pm 0.06$ & $0.18\pm 0.07$\\
\invebm{G} &  $\mathbf{0.67}\pm 0.09$ & $\mathbf{0.14}\pm 0.06$\\
\hdashline
\landm{G} & $0.65\pm 0.11$ & $0.34\pm 0.05$\\
\rbf{G} &  $0.47\pm 0.14$ & $2.2\pm 0.1$\\
\end{tabular}
};

\begin{scope}
    \draw [anchor=north west,fill=white, align=left] (0\linewidth, 1\linewidth) node {{\bf a)} Geodesics evaluation on UCG} ;
    
    \draw [anchor=north west,fill=white, align=left] (0.5\linewidth, 1\linewidth) node {{\bf b)} Geodesics evaluation on WCG};

%    \draw [anchor=north west,fill=white, align=left] (0.67\linewidth, 1\linewidth) node {\bf c)};
\end{scope}
\end{tikzpicture}
%\vspace{-7mm}
\caption{{\bf Quantitative evaluation of the geodesics on the UCG and WCG datasets.}  We report (i) the accumulated probability along the geodesic (the higher the better) and ii) RMSE between each geodesic and its corresponding baseline (the lower the better). Values after the $\pm$ sign indicate the $2\text{-}\sigma$ error.
}
\label{fig:fig1_2sig}
%\vspace{-4mm}
\end{figure}

\newpage

\section{Experimental details on the Rotated Character Dataset}\label{app:rotated_char}

\subsection{Datasets}\label{app:rotated_char_details}
\begin{figure}[h!]
  \centering
  \includegraphics[width=0.5\textwidth]{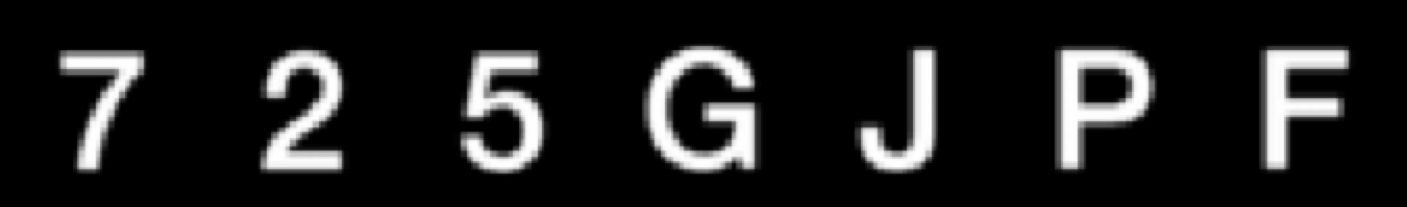}
  \caption{{\bf Original (non-rotated) samples from the Rotated Character Dataset}}
  \label{app:sample_rotated}
\end{figure}

The Rotated Character Datasets consist of 7 printed characters (5, G, F, P, J, 7, 2), represented as black-and-white images of size $32 \times 32$. These characters were selected for two main reasons: (i) they are commonly used in psychophysics experiments~\citep{cooper1973chronometric}, and (ii) they are asymmetric and visually distinct, which helps avoid ambiguities in the resulting geodesic trajectories. Fig.~\ref{app:sample_rotated} shows all characters in their unrotated form.

\begin{wrapfigure}{r}{0.5\textwidth}
\vspace{-3mm}
  \centering
  \includegraphics[width=0.5\textwidth]{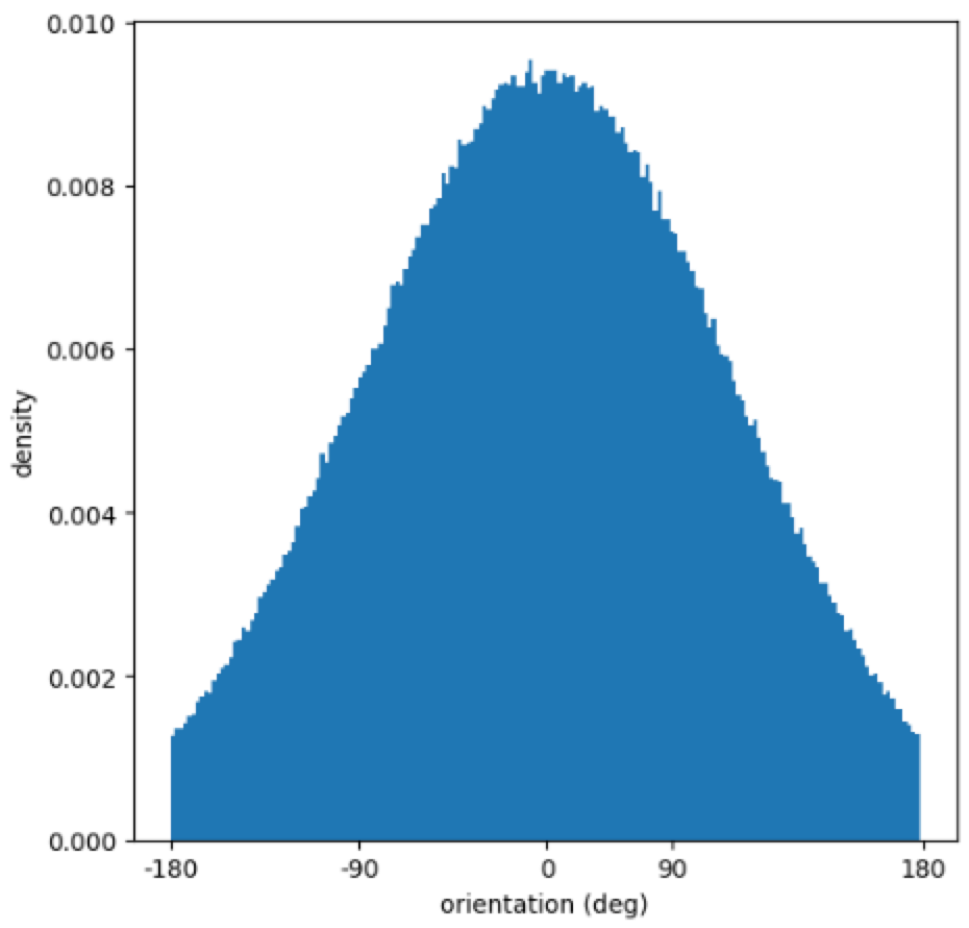}
  \caption{{\bf Distribution of orientation for the BRC dataset}}
  \label{app:distrib_orientation_brc}
\vspace{-12mm}
\end{wrapfigure}
The only difference between the Uniform Rotated Character (URC) and Biased Rotated Character (BRC) datasets lies in the distribution of character orientations. 

\paragraph{Uniform Rotated Character (URC)} In this setting, character orientations are sampled uniformly across the full range of $[-179^\circ, 180^\circ]$, using a one-degree step. This ensures that each possible orientation within this interval is equally likely. Importantly, the distribution is consistent across all characters, meaning that each character appears with the same uniform spread of rotations.

\paragraph{Biased Rotated Character (BRC)} Here, orientations follow a truncated Gaussian distribution centered at $0^\circ$, designed to mimic natural rotation statistics (see Fig.~\ref{app:distrib_orientation_brc}). Unlike the Mixture of Gaussian datasets, we do not have access to a closed-form expression for the underlying distribution $p_{\mathcal{M}}$, but we do control its empirical form. This setup introduces a controlled curvature in the data manifold, allowing us to assess how well different metrics adhere to it.

\subsection{Architecture and algorithm of the Triplet Loss autoencoder}
\begin{wrapfigure}{r}{0.7\textwidth}
  \vspace{-10pt} % Adjust vertical spacing as needed
  \begin{minipage}{0.65\textwidth}
    \begin{algorithm}[H]
    \SetAlgoLined
    \KwIn{Dataset $\mathcal{D} = \{(\v{x}_i, \theta_i)\}$, Encoder $E_\phi$, Decoder $D_\psi$}
    \While{\textnormal{training}}{
      Sample bath of triplet B=$(\v{x}_a, \v{x}_p, \v{x}_n)$ from $\mathcal{D}$ \\
      \quad \textcolor{gray}{\# Same character; $\theta_p$ close to $\theta_a$, $\theta_n$ farther} \\
      $\v{z}_a = E_\phi(\v{x}_a)$,\quad $\v{z}_p = E_\phi(\v{x}_p)$,\quad $\v{z}_n = E_\phi(\v{x}_n)$ \\
      $\mathcal{L}_{\text{rec}} = \| D_\psi(\v{z}_a) - \v{x}_a \|^2$ \\
      $\Delta\theta_p = |\theta_a - \theta_p|$,\quad $\Delta\theta_n = |\theta_a - \theta_n|$ \\
      $\mathcal{L}_{\text{T}} = \mathbb{E}_{\textnormal{B}}\bigg(\!\!\left( \| \v{z}_a - \v{z}_p \|\!-\! \alpha\Delta\theta_p \right)^2\!+\!\left( \| \v{z}_a\!-\!\v{z}_n \|\!-\!\alpha\Delta\theta_n \right)^2\!\!\bigg)$ \\
      $\mathcal{L}_{\text{total}} = \mathcal{L}_{\text{rec}} + \lambda \cdot \mathcal{L}_{\text{T}}$ \\
      Update $(\phi, \psi)$ using gradient $\nabla \mathcal{L}_{\text{total}}$
    }
    \caption{Autoencoder with Triplet regularization}
    \label{algo:autoencoder-triplet-angular}
    \end{algorithm}
  \end{minipage}
  %\vspace{-10pt}
\end{wrapfigure}
We computed geodesics in the latent space of an autoencoder trained with a Triplet Loss~\citep{hoffer2015deep}. This approach is motivated by the fact that image space is inherently non-Euclidean, making it poorly suited for defining meaningful distances. In contrast, the latent space of our autoencoder is explicitly regularized so that Euclidean distances correspond to differences in orientation. By treating the latent space as the ambient space for geodesic computation, we align with the assumption that the data manifold is embedded in an Euclidian Manifold. The training procedure is described in Algo.~\ref{algo:autoencoder-triplet-angular}, and the encoder and decoder architectures—based on the Regularized Autoencoder (RAE) framework~\citep{ghosh2020from}—are detailed in Table~\ref{tab:autoencoder_encoder_architecture} and Table~\ref{tab:autoencoder_decoder_architecture}, respectively.

We trained the model using the Adam optimizer~\citep{kingma2014adam} with a learning rate of $1 \times 10^{-4}$ and a batch size of 128. In Algorithm~\ref{algo:autoencoder-triplet-angular}, we set $\alpha = 1$ and $\lambda = 0.1$. For the architecture, the number of input features (i.e., the number of channels in the first convolutional layer) was set to $F = 128$. In Table~\ref{tab:autoencoder_encoder_architecture} and Table~\ref{tab:autoencoder_decoder_architecture}, the notation "Conv2D($n_c$, $n_f$, 3, 1)" refers to a convolutional layer with $n_c$ input channels, $n_f$ output channels, a kernel size of 3, and padding of 1. Similarly, "ConvTr2D" denotes a transposed convolution. The RAE blocks are modules introduced in~\citep{ghosh2020from}, referred to here as RaeBlockDown and RaeBlockUp, and are used for efficient downsampling and upsampling, respectively.

\begin{table}[h]
\centering
\begin{minipage}{0.48\textwidth}
\centering
\begin{tabular}{|c|c|}
\hline
\textbf{Nb. Layers} & \textbf{Layer Type} \\
\hline
1 & Conv2d (1, $F$, 3, 1) \\
\hline
\multirow{2}{*}{1} & RaeBlockDown ($F$, $2F$) \\
 & ReLU \\
\hline
1 & Conv2d ($2F$, $2F$, 3, 1) \\
\hline
 \multirow{2}{*}{1} & RaeBlockDown ($2F$, $4F$) \\
 & ReLU \\
\hline
\multirow{2}{*}{1} & Conv2d ($4F$, $4F$, 3, 1) \\
& ReLU \\
\hline
1 & Linear ($4F*8*8$,z) \\
\hline
\end{tabular}
\caption{Encoder architecture of the autoencoder. $F$ is the number of features ($F=128$), and z is the size of the latent space ($z=64$).}
\label{tab:autoencoder_encoder_architecture}
\end{minipage}%
\hfill
\begin{minipage}{0.48\textwidth}
\centering
\begin{tabular}{|c|c|}
\hline
\textbf{Nb. Layers} & \textbf{Layer Type} \\
\hline
\multirow{2}{*}{1} & ConvTr2d ($z$, $4F$, 8, 0) \\
 & ReLU \\
\hline
1 & Conv2d ($4F$, $4F$, 3, 1) \\
\hline
\multirow{2}{*}{1} & RaeBlockUp ($4F$, $2F$) \\
 & ReLU \\
\hline
1 & Conv2d ($2F$, $2F$, 3, 1) \\
\hline
\multirow{2}{*}{1} & RaeBlockUp ($2F$, $F$) \\
 & ReLU \\
\hline
1 & Conv2d ($F$, $F$, 3, 1) \\
\hline
\multirow{2}{*}{1} & Conv2d ($F$, 1, 4, 1) \\
 & Tanh \\
\hline
\end{tabular}
\caption{Decoder architecture of the autoencoder. $F$ is the number of features ($F=128$), and z is the size of the latent space ($z=64$).}
\label{tab:autoencoder_decoder_architecture}
\end{minipage}
\end{table}

\subsection{Architecture of the energy function and the interpolant network on the Rotated Character Dataset}\label{app:ebm_alphanum}

\begin{figure}[h!]
  \centering
  \includegraphics[width=0.5\textwidth]{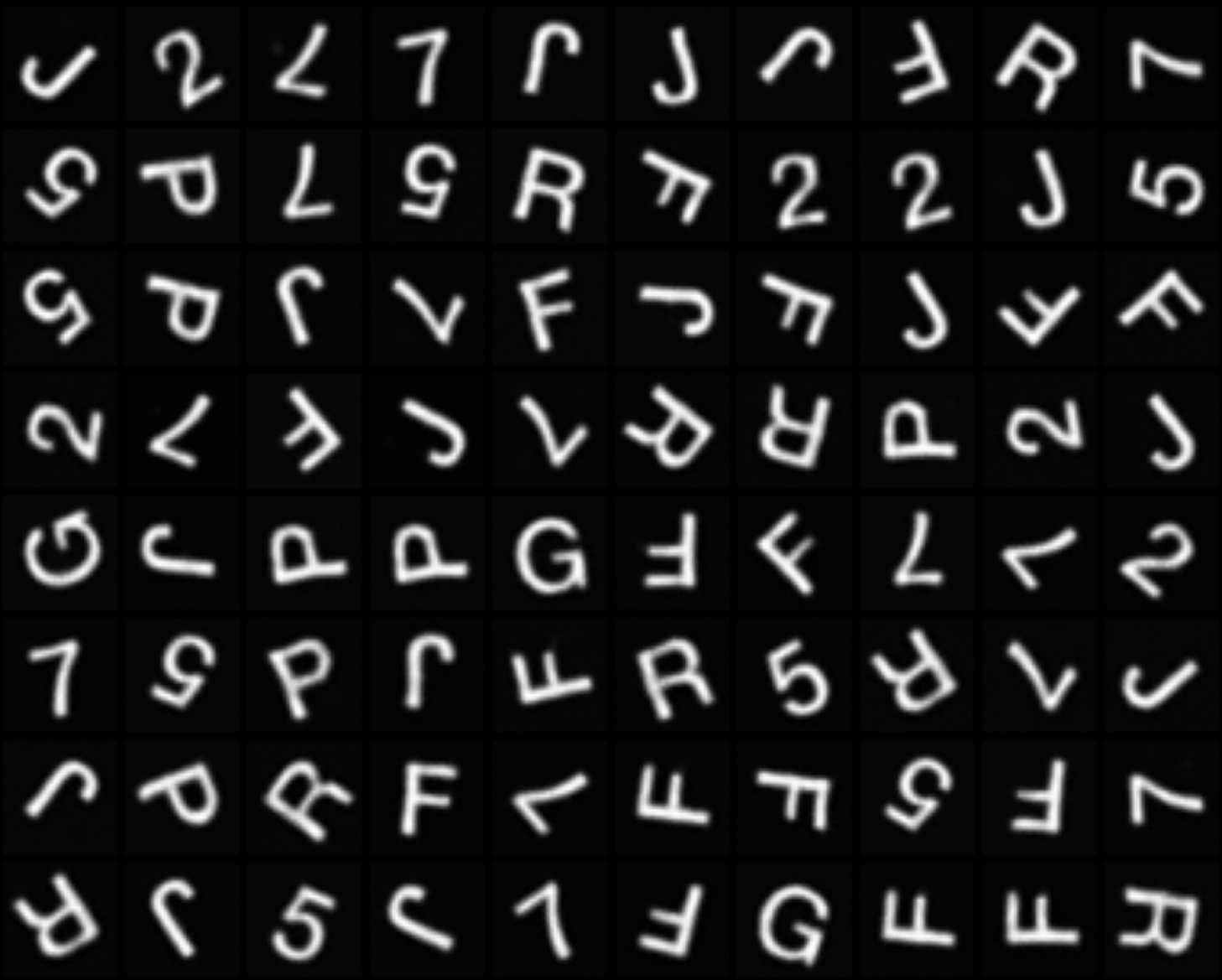}
  \caption{{\bf Samples from the EBM train on URC.} These samples are generated by applying Langevin dynamics to the energy function learned by the EBM.}
  \label{app:sample_EBM}
\end{figure}

The architecture of the energy function used in the EBM is shown in Table~\ref{fig:archi_alphanum_energy}, and the architecture of the interpolant network is provided in Table~\ref{fig:archi_alphanum_interpolant}. These architectures are used for both the URC and BRC datasets. The EBM was trained using the procedure described in Algorithm~\ref{algo:train-ebm}, and Fig.\ref{app:sample_EBM} shows samples generated by the EBM at the end of training. All EBM training hyperparameters match those described in Section\ref{app:EBM_algorithm}. For both the EBM and interpolant training, we use a batch size of 128. The interpolant network is optimized with Adam, using a learning rate of $1 \times 10^{-4}$.

%\subsection{Architecture and hyperparameters for the rotated typed letters}\label{app:ebm_alphanum}
\begin{table}[h]
\centering
\begin{minipage}{0.48\textwidth}
\centering
\begin{tabular}{|c|c|}
\hline
\textbf{Nb. Layers} & \textbf{Layer Type} \\
\hline
\multirow{2}{*}{1} & Linear (64, 128) \\
 & SiLU \\
 \hline
\multirow{2}{*}{1} & Linear (128, 512) \\
 & SiLU \\
\hline
\multirow{2}{*}{6} & Linear (512, 512) \\
 & SiLU \\
\hline
1 & Linear (512, 64) \\
\hline
 \multirow{4}{*}{1} & Three output heads: \\
  & \quad Linear (64, 1) for $f_1$ \\
  & \quad Linear (64, 1) for $f_2$ \\
  & \quad Linear (64, 1) for $f_3$ \\
\hline
 Output & $f_1(x) \cdot f_2(x) + f_3(x^2)$ \\
\hline
\end{tabular}
\caption{Archiecture of the EBM energy function on both URC and BRC datasets}
\label{fig:archi_alphanum_energy}
\end{minipage}%
\hfill
\begin{minipage}{0.48\textwidth}
\centering
\begin{tabular}{|c|c|}
\hline
\textbf{Nb. Layers} & \textbf{Layer Type} \\
\hline
\multirow{2}{*}{1} & Linear (64*3, 128) \\
 & SiLU \\
 \hline
\multirow{2}{*}{1} & Linear (128, 128) \\
 & SiLU \\
\hline
\multirow{2}{*}{1} & Linear 128, 128) \\
 & SiLU \\
 \hline
\multirow{2}{*}{1} & Linear 128, 128) \\
 & SiLU \\
 \hline
\multirow{2}{*}{1} & Linear 128, 128) \\
 & SiLU \\
 \hline
\multirow{2}{*}{1} & Linear 128, 64) \\
 & SiLU \\
\hline
\end{tabular}
\caption{Architecture of the interpolant network used on the URC and BRC dataset.}
\label{fig:archi_alphanum_interpolant}
\end{minipage}%
\end{table}

\subsection{Hyperparameters of the LAND and RBF metric}\label{app:LAND_RBF_param_alphanum}
\paragraph{LAND metric} We performed a hyperparameter search to tune the $\sigma$ parameter. We found that $\sigma = 0.4$ yielded the best performance. Parameters are similar for both the URC and BRC datasets.

\paragraph{RBF metric} We conducted a hyperparameter search to tune both the number of centroids $K$ and the scaling factor $\kappa$. The best results were obtained with $K = 300$ and $\kappa = 0.75$. Parameters are similar for both the URC and BRC datasets.

\subsection{Additional geodesics}\label{app:More_geodesics_rotated_char}

\begin{figure}[h!]
\begin{tikzpicture}
\draw [anchor=north west] (0\linewidth, 0.97\linewidth) node {\includegraphics[width=1\linewidth]{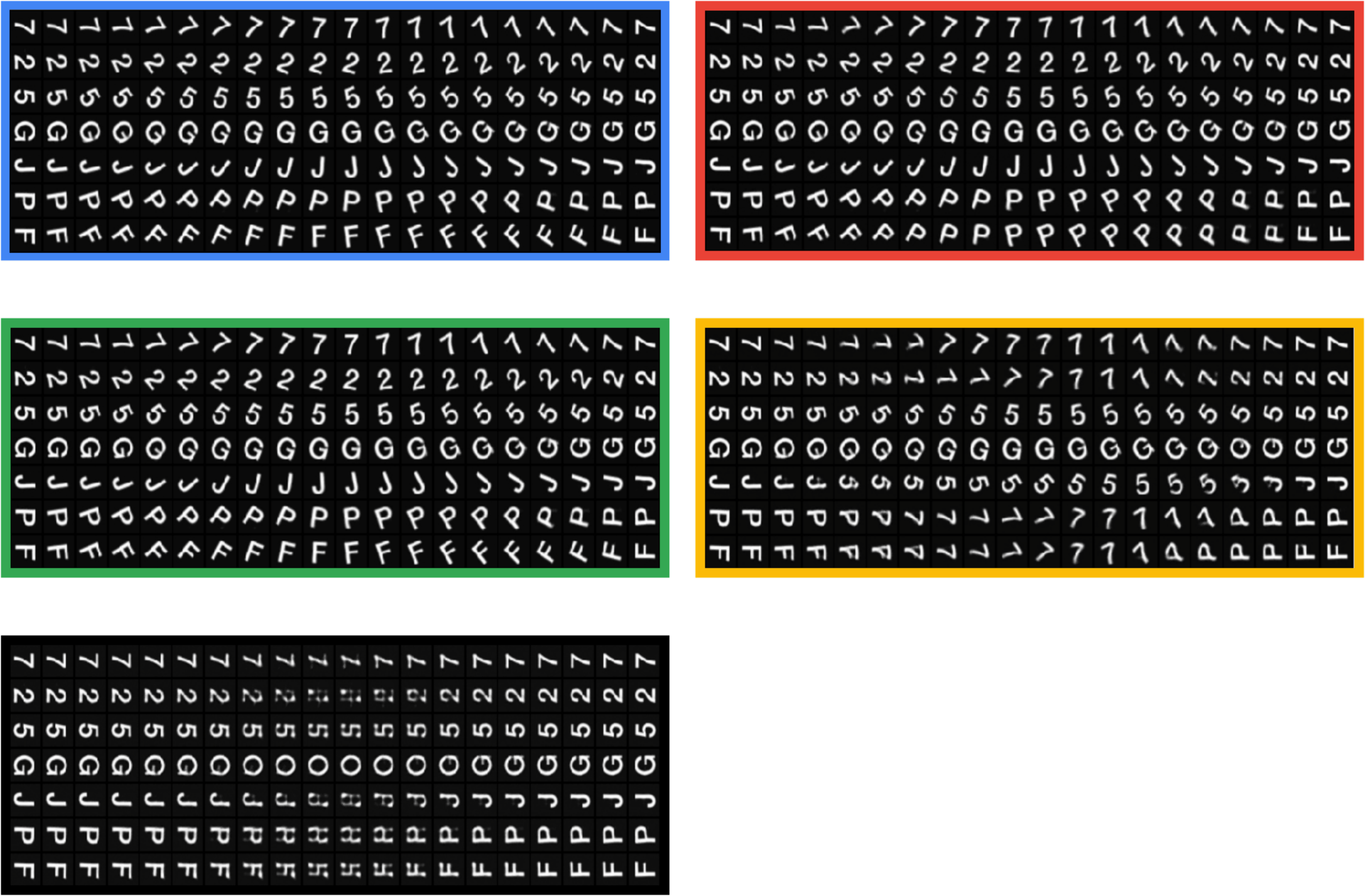}};

\begin{scope}
    \draw [anchor=north west,fill=white, align=left] (0.0\linewidth, 1\linewidth) node {{\bf a)} \logebm{G}} ;
    
    \draw [anchor=north west,fill=white, align=left] (0.5\linewidth, 1\linewidth) node {{\bf b)} \rbf{G}};

    \draw [anchor=north west,fill=white, align=left] (0.0\linewidth, 0.765\linewidth) node {{\bf c)} \invebm{G}} ;
    
    \draw [anchor=north west,fill=white, align=left] (0.5\linewidth, 0.765\linewidth) node {{\bf d)} \landm{G}};

    \draw [anchor=north west,fill=white, align=left] (0\linewidth, 0.53\linewidth) node {{\bf e)} Linear interpolation};
\end{scope}
\end{tikzpicture}
%\vspace{-7mm}
\caption{{\bf Geodesics on the URC dataset.} Geodesics are computed using four different metrics: {\bf a)} \logebm{G}, {\bf b)} \rbf{G}, {\bf c)} \invebm{G}, {\bf d)} \landm{G}. For comparison, a simple linear interpolation is shown in {\bf e)}. The trajectory are computed in the latent space of the autoencoder and projected into pixel space for visualization. Each trajectory is subsampled at 20 time steps for clarity.}
\label{app:additional_geo_alpha_num}
%\vspace{-4mm}
\end{figure}

\paragraph{URC dataset:} In Fig.~\ref{app:additional_geo_alpha_num} we show additional geodesics on the URC dataset.

\begin{figure}[h!]
\begin{tikzpicture}
\draw [anchor=north west] (0\linewidth, 0.97\linewidth) node {\includegraphics[width=1\linewidth]{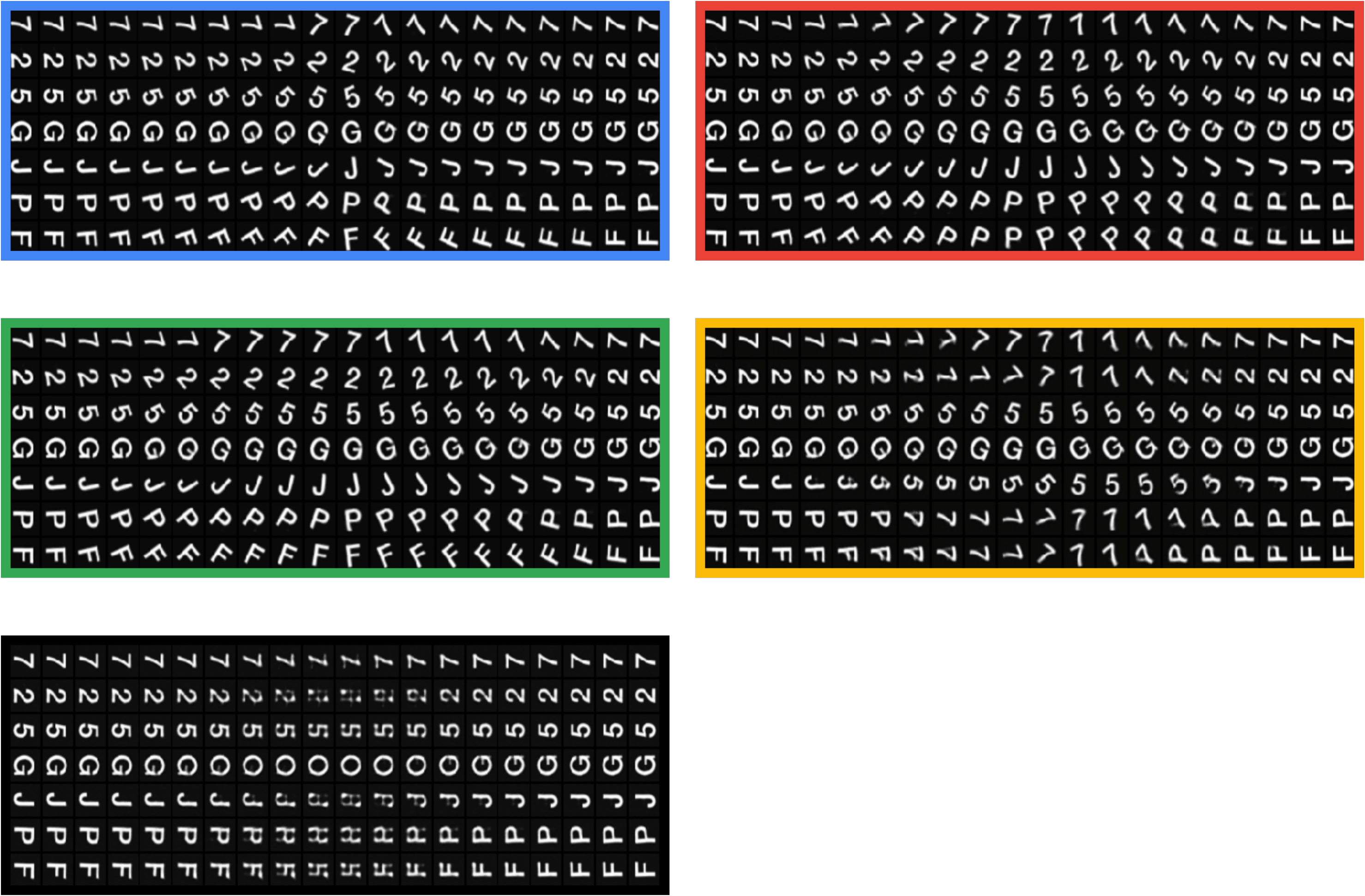}};

\begin{scope}
    \draw [anchor=north west,fill=white, align=left] (0.0\linewidth, 1\linewidth) node {{\bf a)} \logebm{G}} ;
    
    \draw [anchor=north west,fill=white, align=left] (0.5\linewidth, 1\linewidth) node {{\bf b)} \rbf{G}};

    \draw [anchor=north west,fill=white, align=left] (0.0\linewidth, 0.765\linewidth) node {{\bf c)} \invebm{G}} ;
    
    \draw [anchor=north west,fill=white, align=left] (0.5\linewidth, 0.765\linewidth) node {{\bf d)} \landm{G}};

    \draw [anchor=north west,fill=white, align=left] (0\linewidth, 0.53\linewidth) node {{\bf e)} Linear interpolation};
\end{scope}
\end{tikzpicture}
%\vspace{-7mm}
\caption{{\bf Geodesics on the BRC dataset.} Geodesics are computed using four different metrics: {\bf a)} \logebm{G}, {\bf b)} \rbf{G}, {\bf c)} \invebm{G}, {\bf d)} \landm{G}. For comparison, a simple linear interpolation is shown in {\bf e)}. The trajectory are computed in the latent space of the autoencoder and projected into pixel space for visualization. Each trajectory is subsampled at 20 time steps for clarity.}
\label{app:additional_geo_alpha_num_brc}
%\vspace{-4mm}
\end{figure}

\paragraph{BRC dataset:} In Fig.~\ref{app:additional_geo_alpha_num_brc} we show additional geodesics on the BRC dataset.

\subsection{Quantitative evaluation with error bars}\label{app:rot_2sig}

In Table.~\ref{tab:urc_metrics}, we report the same quantitative results as in Fig.~\ref{fig:fig3}, now including $2\text{-}\sigma$ error bars. The standard deviation $\sigma$ is computed over evaluation metrics, each averaged on a different set of randomly sampled trajectories (five sets in total).
\begin{table}[h!]
\centering
\begin{tabular}{c c c c }
\makecell[c]{Metric} & 
\makecell[c]{$\mathcal{D}$-RMSE\\
$(\downarrow)$} & 
\makecell[c]{$\gamma^{\star}$-RMSE \\
$(\downarrow)$} \\
\hline
\makecell[c]{linear\\interp.}
 & $2.96 \pm 0.42$ & $3.52 \pm 0.21$\\
\hdashline
\logebm{G} & $\mathbf{0.11}\pm 0.01$ & $\mathbf{0.40}\pm 0.03$\\
\invebm{G} & $0.14\pm 0.02$ & $0.44\pm 0.07$\\
\hdashline
\landm{G} & $0.66\pm 0.12$ & $2.39\pm 0.51$\\
\rbf{G} & $0.36\pm 0.06$ & $0.86\pm 0.17$\\
\end{tabular}
\caption{
{\bf Quantitative evaluation on the URC dataset with the $2\sigma$ error.}
Quantitative evaluation using two metrics: (i) $\mathcal{D}$-RMSE, which measures proximity to the dataset manifold, and (ii) $\gamma$-RMSE, which measures the deviation from an ideal smooth rotation. Values after the $\pm$ sign indicate the $2\text{-}\sigma$ error.
}
\label{tab:urc_metrics}

\end{table}

\newpage
\section{Experimental details on the Rotated Character Dataset}\label{app:afhq_experiment}

\subsection{Dataset}
\begin{figure}[h!]
  \centering
  \includegraphics[width=0.9\textwidth]{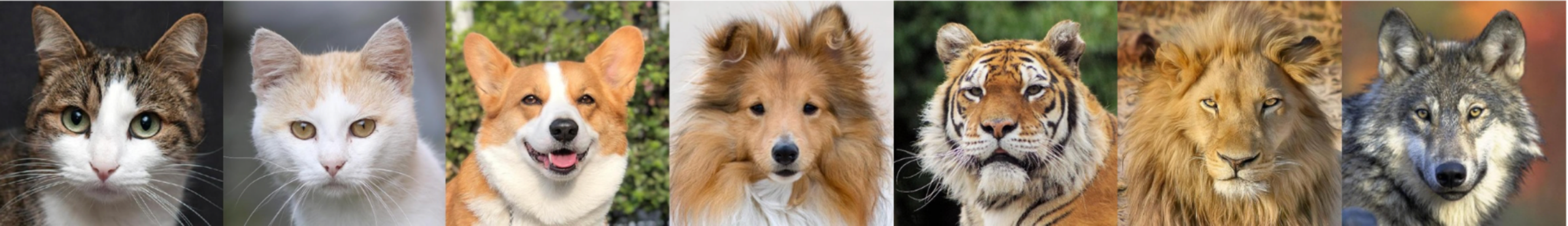}
  \caption{{\bf Samples from the AFHQ dataset~\citep{choi2020starganv2}}}
  \label{app:sample_afhq}
\end{figure}

In this section, we conduct experiments on the Animal Faces High-Quality (AFHQ) dataset introduced by~\cite{choi2020starganv2}. The full dataset contains 15,000 images across three categories: cats, dogs, and wild animals. For our experiments, we restrict the dataset to the cat and dog classes only, each comprising approximately 5,000 images. This choice avoids introducing curvature in the data manifold that could arise from the relatively small number of samples in the wild animal category. All images are cropped, aligned, and have a resolution of 512×512 pixels. AFHQ is widely used for image-to-image translation and style transfer, and its diversity in pose, breed, and appearance makes it well-suited for smooth interpolation tasks. See Fig.~\ref{app:sample_afhq} for example images from the AFHQ dataset.

For the experiments in this section, we compute geodesics in the latent space of a pretrained Variational Autoencoder (VAE). Specifically, we use the VAE from Stable Diffusion V1~\citep{rombach2022high}. The latent representations have a spatial size of $4 \times 16 \times 16$.

\subsection{Architecture of the energy function and the interpolant network on the AFHQ dataset}\label{app:ebm_afhq}

\paragraph{Energy Function:} The architecture used for the energy function is detailed in Table~\ref{tab:energy_function_architecture_afhq}. We set the number of input channels to $n_c = 4$, matching the dimensionality of the latent representation, and use $F = 256$ feature channels in the first convolutional layer. The network follows a simple sequence of downsampling convolutional layers, which we found to yield more stable training than ResNet-style architectures. The EBM is trained using Algorithm~\ref{algo:train-ebm}, with the same hyperparameters as in Section~\ref{app:EBM_algorithm}. To further improve training stability, we add a denoising score matching regularization term and use a cosine learning rate scheduler. 

\begin{table}[h]
\centering

\begin{tabular}{|c|c|}
\hline
\textbf{Nb. Layers} & \textbf{Layer Type} \\
\hline
\multirow{2}{*}{1} & Conv2d ($n_c$, $F$, 3, 1, 1) \\
 & SiLU \\
\hline
\multirow{2}{*}{1} & Conv2d ($F$, $F$, 3, 1, 1)\\
& SiLU \\
\hline
\multirow{2}{*}{1} & Conv2d ($F$, $2F$, 4, 2, 1) \\
& SiLU \\
\hline
\multirow{2}{*}{1} & Conv2d ($2F$, $2F$, 3, 1, 1) \\
& SiLU \\
\hline
\multirow{2}{*}{1}  & Conv2d ($2F$, $4F$, 4, 2, 1) \\
& SiLU \\
\hline
\multirow{2}{*}{1}  & Conv2d ($4F$, $4F$, 3, 1, 1) \\
& SiLU \\
\hline
\multirow{2}{*}{1}  & Conv2d ($4F$, $8F$, 4, 2, 1) \\
& SiLU \\
\hline
1  & Conv2d ($8F$, 1, 2, 1, 0): for $f_1$ \\
1  & Conv2d ($8F$, 1, 2, 1, 0): for $f_2$ \\
1  & Conv2d ($8F$, 1, 2, 1, 0): for $f_3$ \\
\hline
Output & $f_1(x) \cdot f_2(x) + f_3(x^2)$ \\
\hline
\end{tabular}
\caption{{\bf Architecture of the energy function}. $F$ denotes the base number of feature channels, and $n_c$ is the number of input channels. The final energy is computed using three parallel output heads. The notation Conv2d($n_c$, $n_f$, $k$, $s$, $p$) refers to a 2D convolutional layer with $n_c$ input channels, $n_f$ output channels, a kernel size of $k$, stride $s$, and padding $p$.}
\label{tab:energy_function_architecture_afhq}
\end{table}

\begin{figure}[h!]
  \centering
  \includegraphics[width=0.8\textwidth]{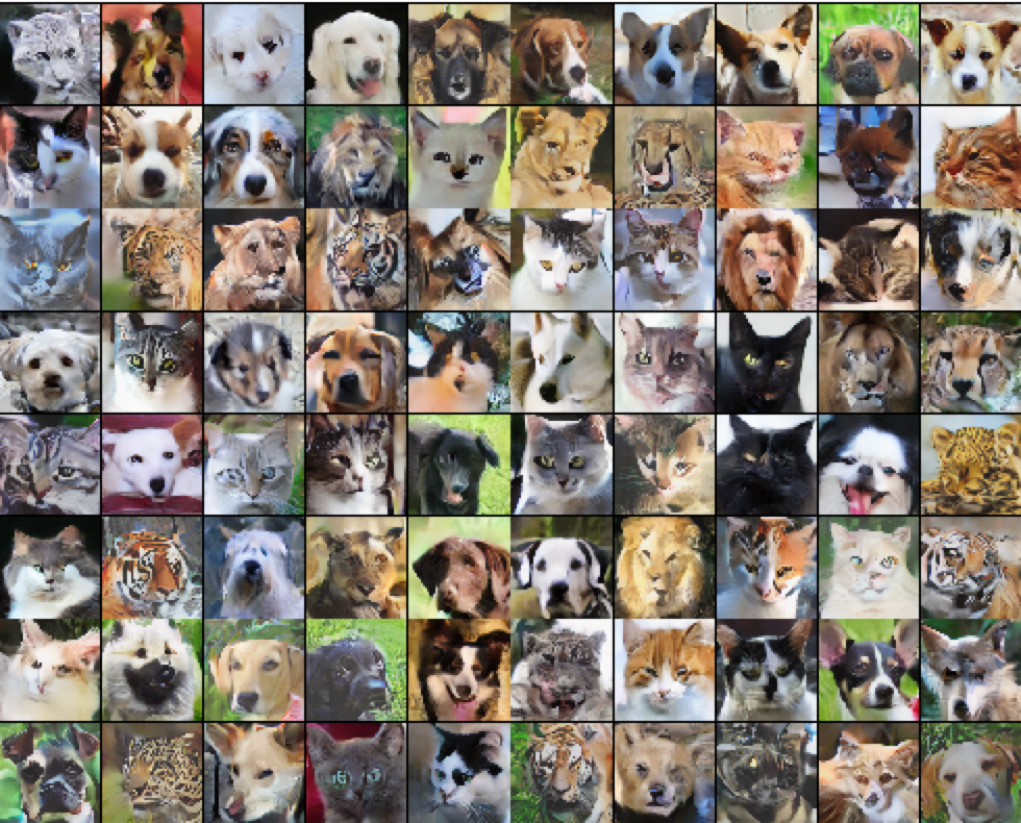}
  \caption{{\bf Samples from the EBM trained on the AFHQ dataset.} These samples are generated by applying Langevin dynamics to the energy function learned by the EBM.}
  \label{app:ebm_sample_afhq}
\end{figure}

In Fig.~\ref{app:ebm_sample_afhq}, we show randomly selected samples generated by the EBM after training. The Fréchet Inception Distance (FID) of the model is measured to be 9.89.

\paragraph{Interpolant Network:} We use the U-Net architecture from~\cite{dhariwal2021diffusion}, following the same hyperparameter settings.

\subsection{Hyperparameters of the LAND and RBF metric}
\paragraph{LAND metric} We performed a hyperparameter search to tune the $\sigma$ parameter. We found that $\sigma = 10$ yielded the best performance. 

\paragraph{RBF metric} We conducted a hyperparameter search to tune both the number of centroids $K$ and the scaling factor $\kappa$. The best results were obtained with $K = 1000$ and $\kappa = 3$. 

\subsection{FIDs with error bars}\label{app:FID_error}

In Table.~\ref{table:FID_err_tb}, we include $2\text{-}\sigma$ error bars. The standard deviation $\sigma$ is computed over different sets of randomly sampled trajectories (five sets in total).
\begin{table}[h!]
\centering
\begin{tabular}{cc}
    Metric & FID $(\downarrow)$\\
    \hline
    Linear interp. & $42.47 \pm 3.17$ \\
    Slerp interp. & $32.67 \pm 2.33$ \\
    \hdashline
    \logebm{G} & $20.79 \pm 2.17$ \\
    \invebm{G} & $\mathbf{16.47} \pm 1.04$ \\
    \hdashline
    \landm{G} & $39.17 \pm 3.63$ \\
    \rbf{G} & $37.98 \pm 2.46$ \\
  \end{tabular}
  \caption{{\bf FID along geodesics for different Riemannian metrics}. FID is computed at each trajectory point to assess on-manifold alignment.  Values after the $\pm$ sign indicate the $2\text{-}\sigma$ error.}
  \label{table:FID_err_tb}
\end{table}

\newpage
\subsection{Additional geodesics on AFHQ}\label{app:AFHQ_samples}

\subsubsection*{Riemanian metric: \invebm{G}}

\begin{figure}[h!]
\begin{tikzpicture}
\draw [anchor=north west] (0.0\linewidth, 0.97\linewidth) node {\includegraphics[width=1\linewidth]{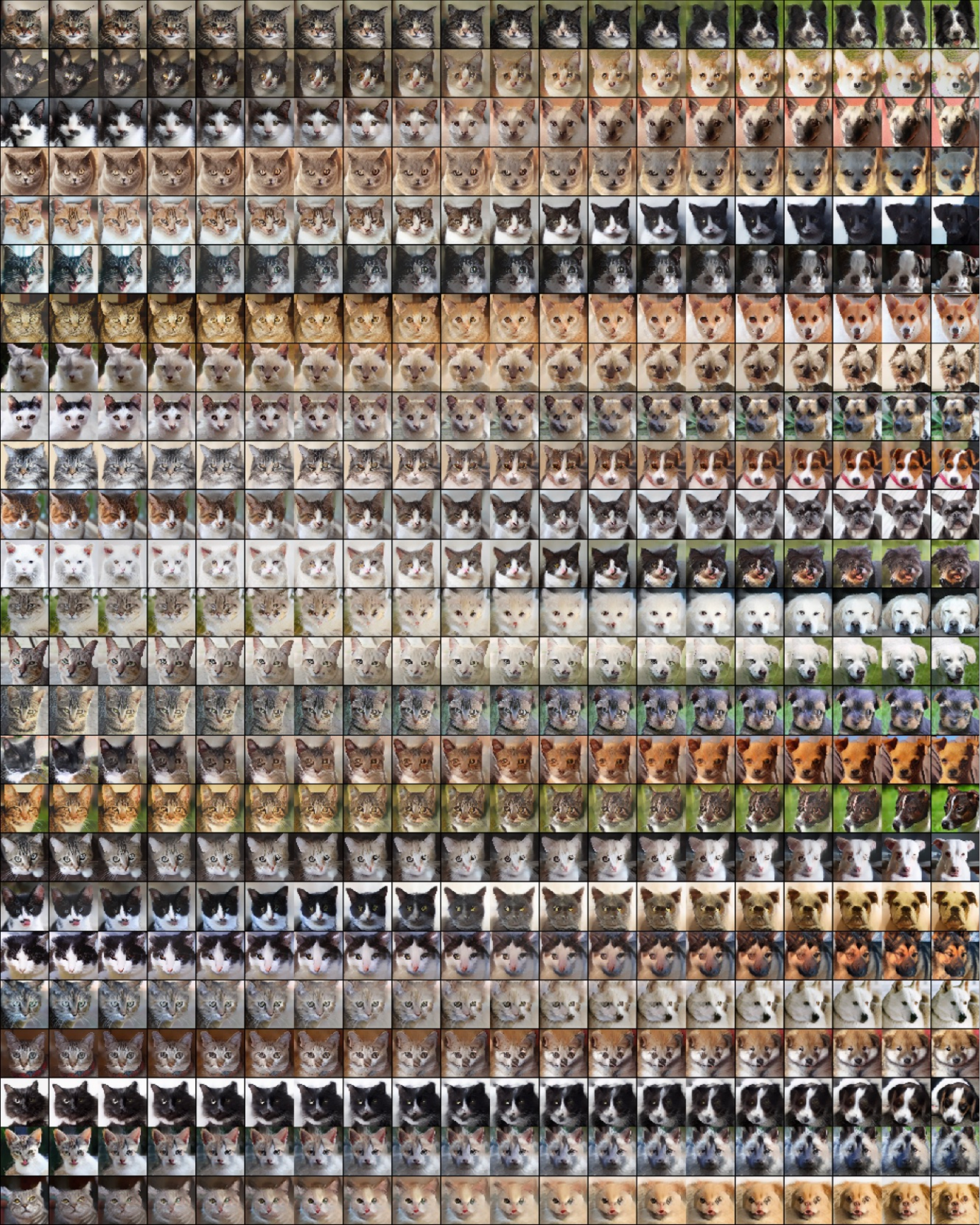}};
\end{tikzpicture}
\caption{{\bf Geodesics on the AFHQ dataset using \invebm{G}.} Each row shows a geodesic in latent space between a randomly sampled cat image (start point) and a dog image (end point). Columns correspond to time steps along each geodesic, from left (start) to right (end). Images are obtained by decoding the latent representations back into pixel space.}
\label{app:fig_sample_invpebm}
\end{figure}

\newpage

\subsubsection*{Riemanian metric: \logebm{G}}
\begin{figure}[h!]
\begin{tikzpicture}
\draw [anchor=north west] (0.0\linewidth, 0.97\linewidth) node {\includegraphics[width=1\linewidth]{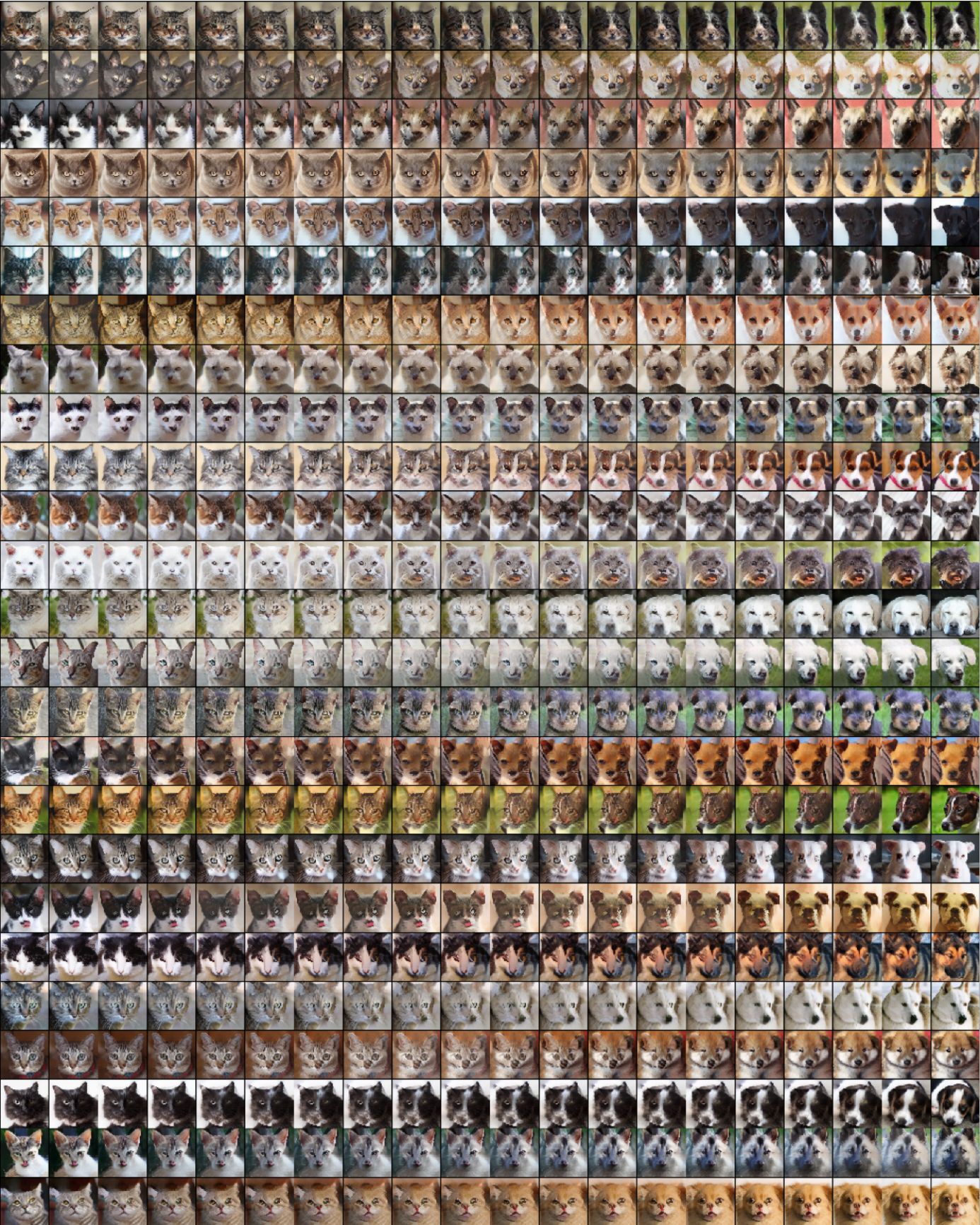}};
\end{tikzpicture}
\caption{{\bf Geodesics on the AFHQ dataset using \logebm{G}.} Each row shows a geodesic in latent space between a randomly sampled cat image (start point) and a dog image (end point). Columns correspond to time steps along each geodesic, from left (start) to right (end). Images are obtained by decoding the latent representations back into pixel space.}
\label{app:fig_sample_logebm}
\end{figure}

\newpage
\subsubsection*{Riemanian metric: \rbf{G}}
\begin{figure}[h!]
\begin{tikzpicture}
\draw [anchor=north west] (0.0\linewidth, 0.97\linewidth) node {\includegraphics[width=1\linewidth]{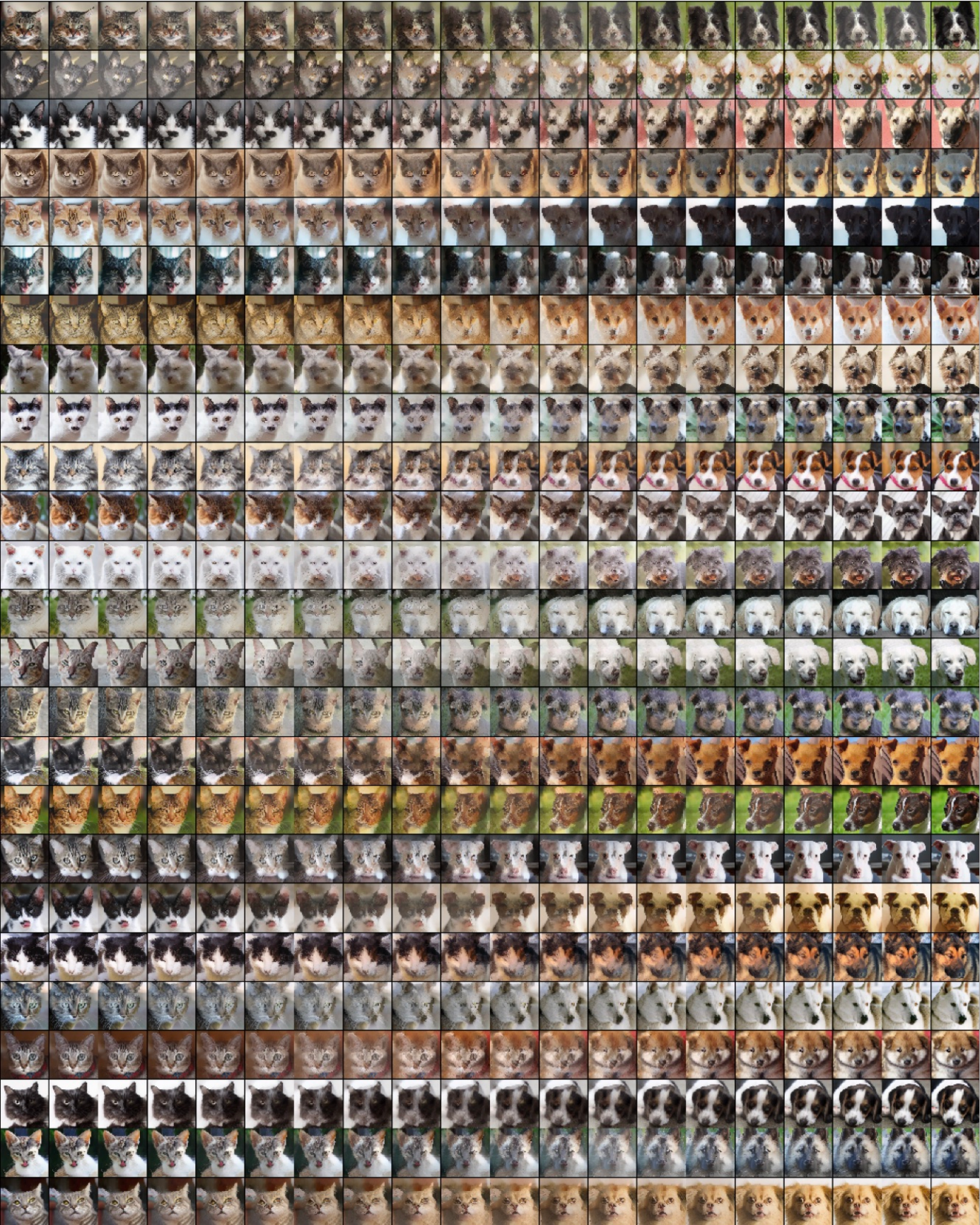}};
\end{tikzpicture}
\caption{{\bf Geodesics on the AFHQ dataset using \rbf{G}.} Each row shows a geodesic in latent space between a randomly sampled cat image (start point) and a dog image (end point). Columns correspond to time steps along each geodesic, from left (start) to right (end). Images are obtained by decoding the latent representations back into pixel space.}
\label{app:fig_sample_rbf}
\end{figure}

\newpage
\subsubsection*{Riemanian metric: \landm{G}}
\begin{figure}[h!]
\begin{tikzpicture}
\draw [anchor=north west] (0.0\linewidth, 0.97\linewidth) node {\includegraphics[width=1\linewidth]{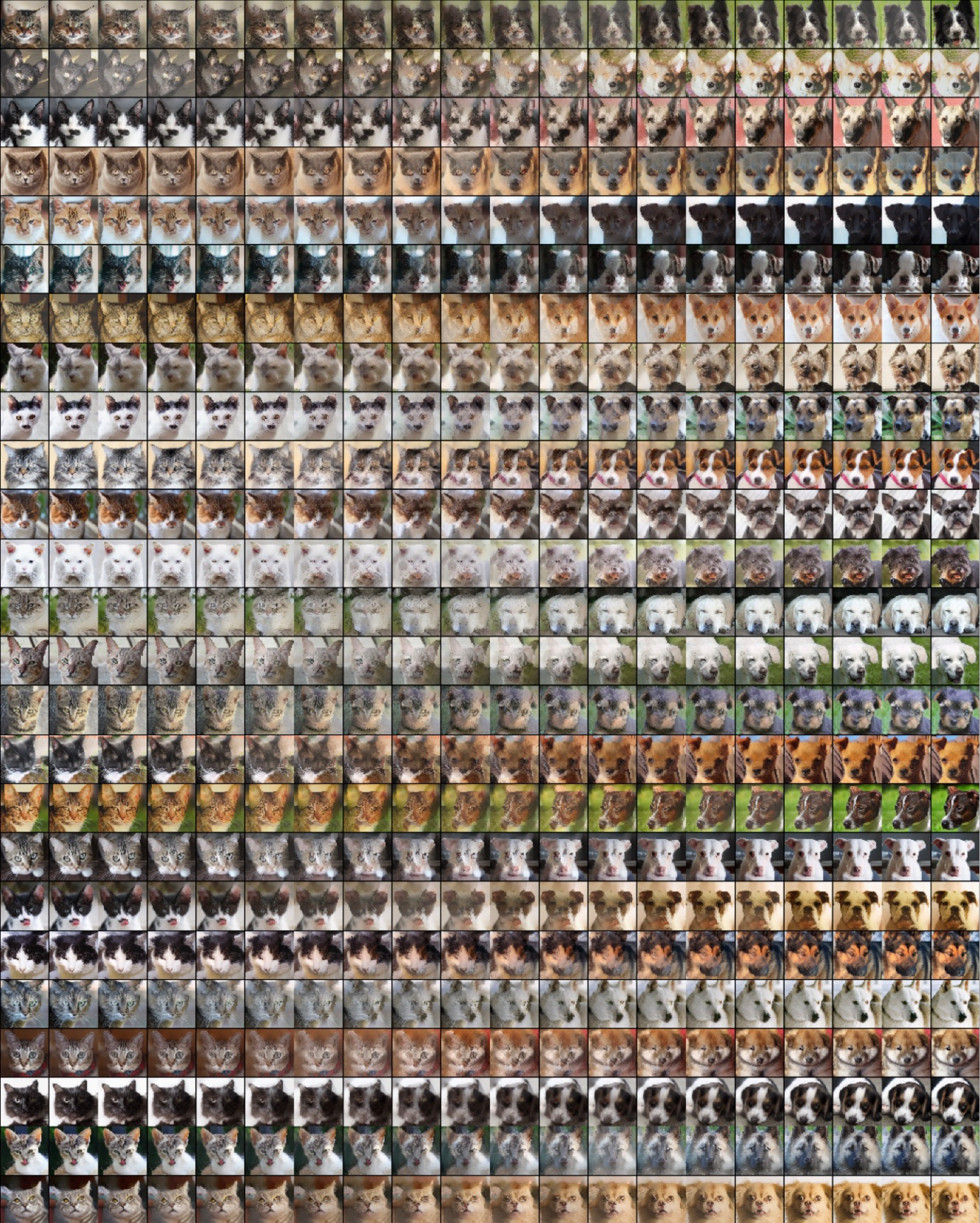}};
\end{tikzpicture}
\caption{{\bf Geodesics on the AFHQ dataset using \landm{G}.} Each row shows a geodesic in latent space between a randomly sampled cat image (start point) and a dog image (end point). Columns correspond to time steps along each geodesic, from left (start) to right (end). Images are obtained by decoding the latent representations back into pixel space.}
\label{app:fig_sample_land}
\end{figure}

\newpage
\subsubsection*{Linear interpolation}

\begin{figure}[h!]
\begin{tikzpicture}
\draw [anchor=north west] (0.0\linewidth, 0.97\linewidth) node {\includegraphics[width=1\linewidth]{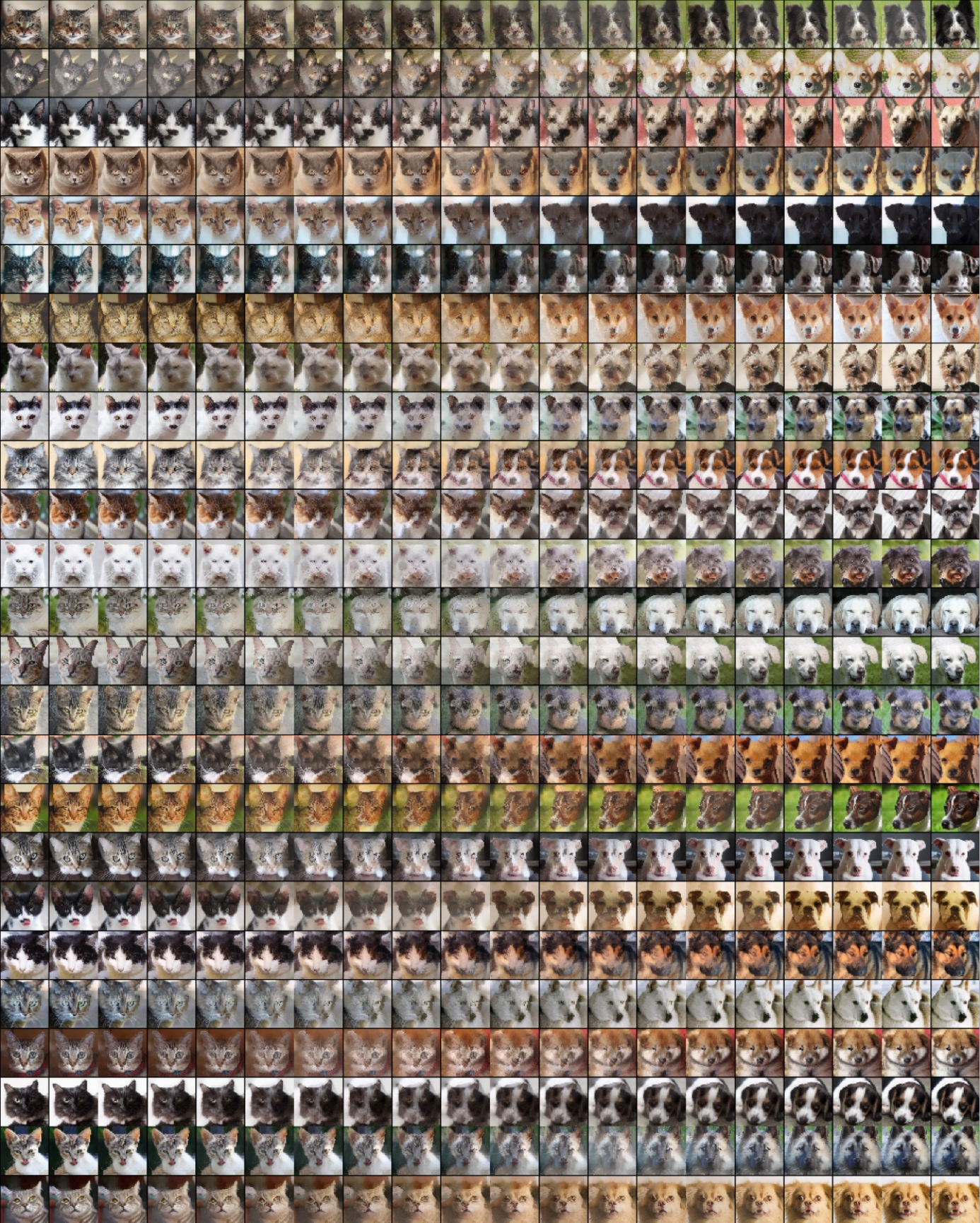}};
\end{tikzpicture}
\caption{{\bf Linear interpolation on the AFHQ dataset.} Each row shows an interpolation in latent space between a randomly sampled cat image (start point) and a dog image (end point). Columns correspond to time steps along each interpolation, from left (start) to right (end). Images are obtained by decoding the latent representations back into pixel space.}
\label{app:fig_sample_lin_interp}
\end{figure}

\newpage
\subsubsection*{Slerp interpolation}
\begin{figure}[h!]
\begin{tikzpicture}
\draw [anchor=north west] (0.0\linewidth, 0.97\linewidth) node {\includegraphics[width=1\linewidth]{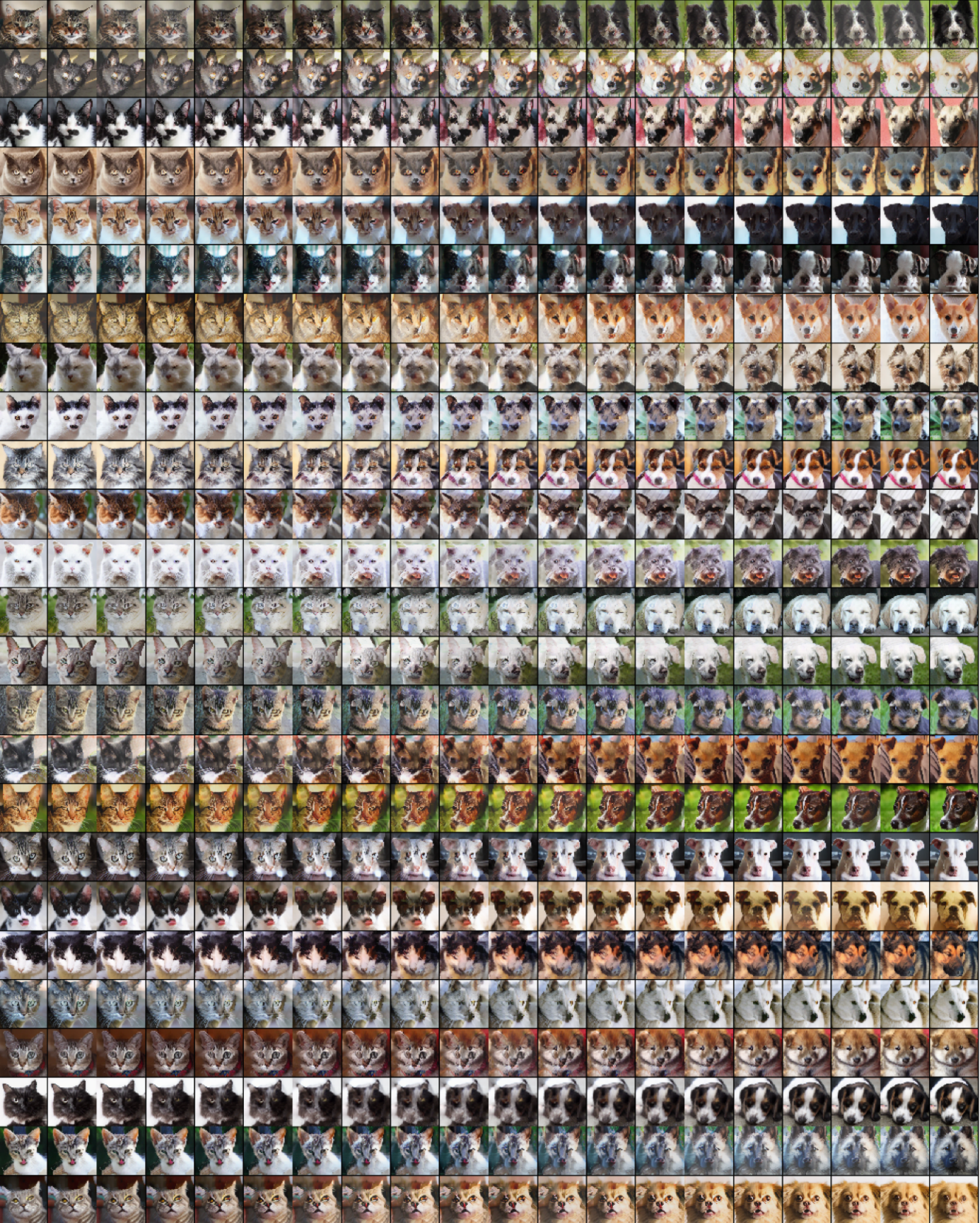}};
\end{tikzpicture}
\caption{{\bf Spherical interpolation (Slerp) on the AFHQ dataset.} Each row shows an interpolation in latent space between a randomly sampled cat image (start point) and a dog image (end point). Columns correspond to time steps along each interpolation, from left (start) to right (end). Images are obtained by decoding the latent representations back into pixel space.}
\label{app:fig_sample_slerp_interp}
\end{figure}

\newpage
\subsection{About the Spherical interpolation}\label{app:spherical_interpolation}

Given two points \(\mathbf{x}_0, \mathbf{x}_1 \in \mathbb{R}^D\) lying on the unit hypersphere (i.e., \(\|\mathbf{x}_0\| = \|\mathbf{x}_1\| = 1\)), the spherical interpolation between them is defined as:
\[
\mathrm{slerp}(t; \mathbf{x}_0, \mathbf{x}_1) = 
\frac{\sin((1 - t) \theta)}{\sin \theta} \mathbf{x}_0 
+ \frac{\sin(t \theta)}{\sin \theta} \mathbf{x}_1,
\quad t \in [0, 1],
\]
where \(\theta\) is the angle between \(\mathbf{x}_0\) and \(\mathbf{x}_1\), given by:
\[
\theta = \arccos\left( \frac{ \langle \mathbf{x}_0, \mathbf{x}_1 \rangle }{ \|\mathbf{x}_0\| \, \|\mathbf{x}_1\| } \right).
\]

In practice, when interpolating latent codes from a Variational Autoencoder (VAE), the latent vectors \(\mathbf{x}_0\) and \(\mathbf{x}_1\) are typically drawn from a standard normal prior and do not lie on the unit sphere. To apply slerp, we first normalize the vectors:
\[
\tilde{\mathbf{x}}_0 = \frac{\mathbf{x}_0}{\|\mathbf{x}_0\|}, \qquad
\tilde{\mathbf{x}}_1 = \frac{\mathbf{x}_1}{\|\mathbf{x}_1\|},
\]
and compute \(\theta\) as:
\[
\theta = \arccos\left( \langle \tilde{\mathbf{x}}_0, \tilde{\mathbf{x}}_1 \rangle \right).
\]

\begin{wrapfigure}{r}{0.5\textwidth}
  \centering
\includegraphics[width=0.5\textwidth]{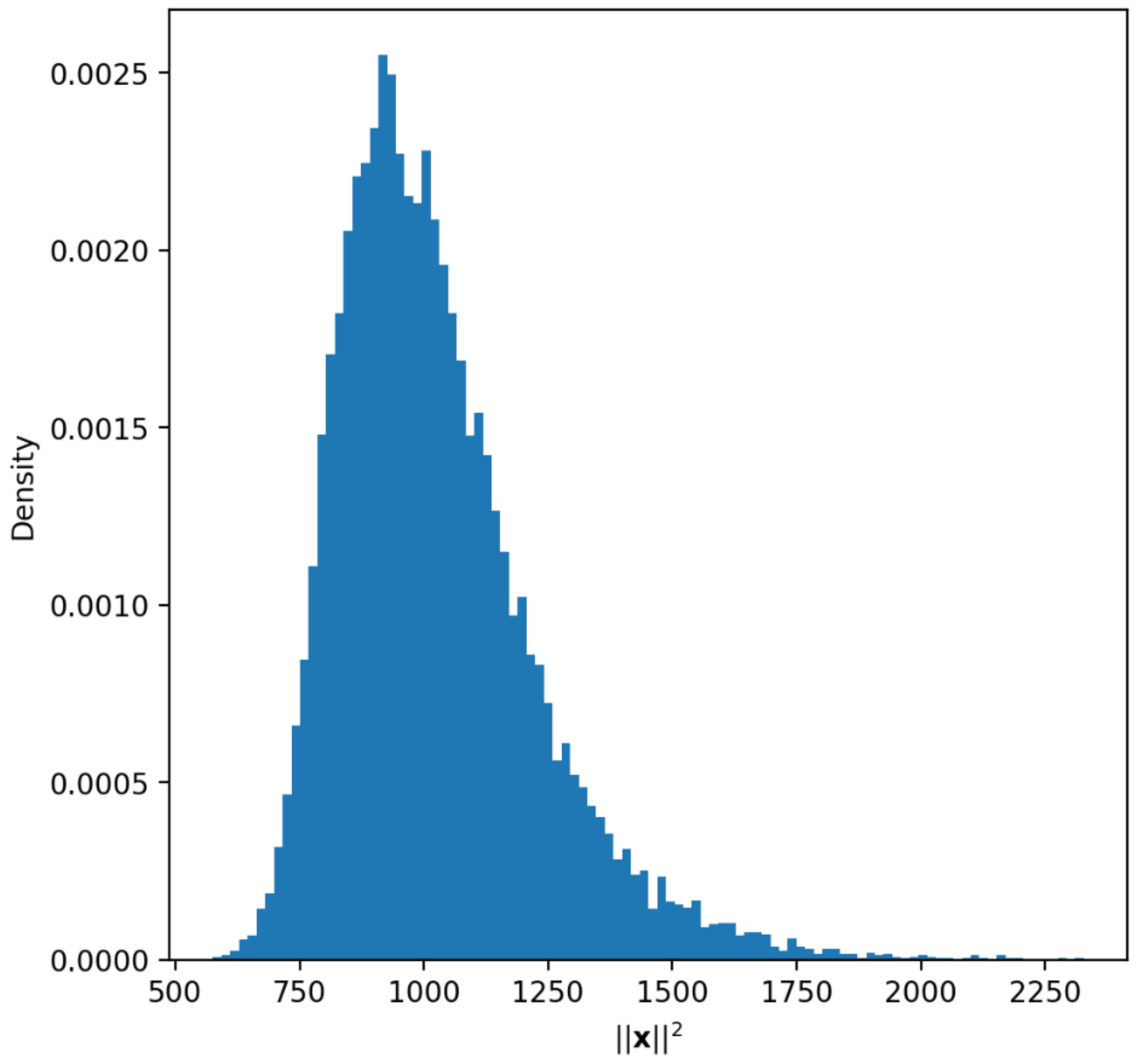}
  \caption{{\bf Distribution of $||\v{x}||^2$ on the AFHQ dataset}}
  \label{app:chi_square}
\end{wrapfigure}
This interpolation method, introduced by~\citep{slerpkarpathy2022}, has proven particularly effective for interpolating in the latent space of VAEs. The intuition behind its success is that it implicitly assumes the data manifold lies on a hypersphere. 
While this may seem restrictive, the assumption is reasonable in practice. In a VAE, each latent coordinate $x_i$ is drawn from a standard Normal distribution: $x_{i}\sim\mathcal{N}(0,1)$ ($1<i<D$). As a result, the squared norm, $\displaystyle ||\v{x}||^{2}=\sum_{i=1}^{D}x_{i}^{2}$ follows a chi-squared distribution with $D$ degree of freedom. This distribution is known to concentrate tightly around $D$, effectively placing most latent codes near the surface of a hypersphere. To validate this empirically, we visualize the distribution of $||\v{x}||^2$ for all latent codes of the AFHQ dataset (see Fig.~\ref{app:chi_square}). We observe that this distribution is concentrated on $D$---$D=1024$ for VAE with latent space of size $4\times16\times16$. 

To conclude, slerp interpolation is well-suited for VAE latent spaces because it aligns with their underlying geometric structure.

\newpage
\subsection{Physical interpretation}\label{app:physical_interpretation}
We refer the reader to~\cite{o1983semi} or~\cite{robbin2022introduction} for a detailed background on differential geometry.  
  
\paragraph{Geodesic equation.} Assume that the manifold $\mathcal{M}$ is the ambient $D$-dimensional Euclidean space ($\mathcal{M}=\mathbb{R}^{D}$). We equipped the manifold $\mathcal{M}$ with a conformal Riemannian metric $\displaystyle\v{G}(\v{x})=\frac{1}{p(\v{x})}\cdot\v{I}$, with $p$ the probability density of the data, and $\v{I}$ the identity matrix of $\mathbb{R}^{D\times D}$. Let $\boldsymbol{\gamma}(t)$ be a geodesic (i.e. $\gamma: [0,1]\rightarrow\mathbb{R}^{D}$). We denote the instantaneous speed of the geodesic at time $t$, $\dot{\boldsymbol{\gamma}}(t)$, and its acceleration $\ddot{\boldsymbol{\gamma}}(t)$. Said otherwise, $\dot{\boldsymbol{\gamma}}(t)$ and $\ddot{\boldsymbol{\gamma}}(t)$ denote $\frac{\partial\boldsymbol{\gamma}}{\partial t}(t)$ and $\frac{\partial^2\boldsymbol{\gamma}}{\partial t^2}(t)$ respectively.

The geodesic equation is the 2nd-order ODE written as:
\begin{equation}
\displaystyle \ddot{\boldsymbol{\gamma}}^{k}(t) + \sum_{i,j}\Gamma_{i,j}^{k}(\boldsymbol{\gamma}(t))\cdot\dot{\boldsymbol{\gamma}}^{i}(t)\cdot\dot{\boldsymbol{\gamma}}^{j}(t) = 0
\label{app:geodesics_eq}
\end{equation}
In Eq.\ref{app:geodesics_eq}, $\ddot{\boldsymbol{\gamma}}^{k}(t)$ and $\dot{\boldsymbol{\gamma}}^{k}(t)$ denotes the k-th coordinate of $\ddot{\boldsymbol{\gamma}}(t)$ and $\dot{\boldsymbol{\gamma}}(t)$, respectively (here $1<k<D$). $\Gamma_{i,j}^{k}$ are the Christoffel symbols, they are derived from the Riemannian metric and encode how it bends and curves the space. $\Gamma_{i,j}^{k}$ tells how much the change in direction in the $i$-th and $j$-th coordinate causes acceleration in the $k$-th coordinate ($1<i,j,k<D$). Said differently, $i$, $j$ refer to the coordinate direction along which the particule is moving, and $k$ refers to the coordinate direction where the motion causes effect (i.e. curvature induces acceleration).

\paragraph{Christoffel symbols for conformal metric.} The Christoffel symbols for a conform metric $\v{G}(\v{x}) = \lambda(\v{x})\cdot\v{I}$ (with $\lambda$ a scalar function):
\begin{equation}
\Gamma^k_{ij}(\v{x}) = \frac{1}{2\lambda(\v{x})} \left( \delta_{j,k} \, \partial_i \lambda(\v{x}) + \delta_{i,k} \, \partial_j \lambda(\v{x}) - \delta_{ij} \, \partial_k \lambda(\v{x}) \right)
%\Gamma_{i,j}^{k}(\v{x})=\frac{1}{2\lambda(\v{x})}
\label{app:christoffel}
\end{equation}
In Eq.~\ref{app:christoffel}, $\displaystyle\partial_i \lambda(\v{x}) = \frac{\partial \lambda(\v{x})}{\partial x^i}$ (i.e. the  partial derivative of \( \lambda (\v{x}) \) with respect to the \( i \)-th coordinate), and $\delta_{j,k}$ is the Kronecker symbol ( $\delta_{j,k}=1$ if $j=k$ and $\delta_{j,k}=0$ otherwise). If one plugs Eq.~\ref{app:christoffel} in the right hand side of Eq.~\ref{app:geodesics_eq}:
\begin{align}
\sum_{i,j} \Gamma^k_{ij}\big(\boldsymbol{\gamma}(t)\big) \, \dot{\boldsymbol{\gamma}}^i(t) \, \dot{\boldsymbol{\gamma}}^j(t)
&= \frac{1}{2\lambda\big(\boldsymbol{\gamma}(t)\big)} 
  \cdot \Bigg[ \sum_i \partial_i \lambda\big(\boldsymbol{\gamma}(t)\big) \, \dot{\boldsymbol{\gamma}}^i(t) \, \dot{\boldsymbol{\gamma}}^k(t) \nonumber \\
&\phantom{=} \hspace{3.1em}
  + \sum_j \partial_j \lambda\big(\boldsymbol{\gamma}(t)\big) \, \dot{\boldsymbol{\gamma}}^k(t) \, \dot{\boldsymbol{\gamma}}^j(t) \nonumber \\
&\phantom{=} \hspace{3.1em}
  - \sum_i \partial_k \lambda\big(\boldsymbol{\gamma}(t)\big) \, \dot{\boldsymbol{\gamma}}^i(t)^2 \Bigg] \nonumber \\
&= \frac{1}{2\lambda\big(\boldsymbol{\gamma}(t)\big)} 
  \cdot \Bigg[2\dot{\boldsymbol{\gamma}}^k(t)\big\langle \nabla \lambda\big(\boldsymbol{\gamma}(t)\big), \dot{\boldsymbol{\gamma}}(t)\big\rangle 
  - \partial_k \lambda\big(\boldsymbol{\gamma}(t)\big) \, \|\dot{\boldsymbol{\gamma}}(t)\|^2 \Bigg], 
\end{align}
where $\langle\cdot,\cdot\rangle$ and $\|\cdot\|$ are the usual \textit{Euclidean} inner product and norms, respectively.  

So Eq.~\ref{app:geodesics_eq}, becomes : 
\begin{equation}
\displaystyle \ddot{\boldsymbol{\gamma}}^{k}(t) = - \frac{\dot{\boldsymbol{\gamma}}^k(t)}{\lambda\big(\boldsymbol{\gamma}(t)\big)}\big\langle \nabla \lambda\big(\boldsymbol{\gamma}(t)\big), \dot{\boldsymbol{\gamma}}(t)\big\rangle + \frac{1}{2\lambda\big(\boldsymbol{\gamma}(t)\big)}\partial_k \lambda\big(\boldsymbol{\gamma}(t)\big) \, \|\dot{\boldsymbol{\gamma}}(t)\|^2
\label{app:eq_geo_2}
\end{equation}

\paragraph{Pulling everything together.}
If one plugs our definition of the Riemannian metric (i.e. $\displaystyle\lambda\big(\boldsymbol{\gamma}(t)\big)=\frac{1}{p\big(\boldsymbol{\gamma}(t)\big)}$, and therefore $\displaystyle\frac{\nabla{\lambda\big(\boldsymbol{\gamma}(t)\big)}}{\lambda\big(\boldsymbol{\gamma}(t)\big)}=-\nabla{\log p\big(\boldsymbol{\gamma}(t)\big)}$), Eq.~\ref{app:eq_geo_2} becomes:
\begin{equation}
\ddot{\boldsymbol{\gamma}}(t) 
= \left\langle \nabla_{\gamma} \log p(\boldsymbol{\gamma}(t)), \dot{\boldsymbol{\gamma}}(t) \right\rangle \cdot \dot{\boldsymbol{\gamma}}(t) 
\,-\, \frac{1}{2} \|\dot{\boldsymbol{\gamma}}(t)\|^2 \cdot \nabla_{\gamma} \log p(\boldsymbol{\gamma}(t))
\label{eq:Newton}
\end{equation}

% Let $s(x)=\nabla_x\log{p(x)}$ be the Stein score. Then we get:

% \begin{equation}
% \ddot{\boldsymbol{\gamma}}(t) 
% = \left\langle s(\boldsymbol{\gamma}(t)), \dot{\boldsymbol{\gamma}}(t) \right\rangle \cdot \dot{\boldsymbol{\gamma}}(t) 
% \,-\, \frac{1}{2} \|\dot{\boldsymbol{\gamma}}(t)\|^2 \cdot s(\boldsymbol{\gamma}(t)).
% \label{eq:NewtonStein}
% \end{equation}
Eq.~\ref{eq:Newton} is similar in form to Newton's second law. The acceleration of a particle (of unit mass) is governed by a velocity-dependent force built from the Stein Score (i.e. $\nabla_{\gamma} \log p(\boldsymbol{\gamma}(t))$). More speficically:
\begin{itemize}
    \item $\left\langle \nabla \log p(\boldsymbol{\gamma}(t)), \dot{\boldsymbol{\gamma}}(t) \right\rangle \cdot \dot{\boldsymbol{\gamma}}(t)$ describes a "force" aligned with the particle velocity direction. This term acts like an anisotropic drag or propulsion term: i) it speeds up the particle when it goes toward a high-density region and ii) it slows down the particle going the other way.
    \item $-\, \frac{1}{2} \|\dot{\boldsymbol{\gamma}}(t)\|^2 \cdot \nabla \log p(\boldsymbol{\gamma}(t))$ is a force in the direction of the stein score (pointing toward low density regions). It behaves like a repulsive force, pushing the particle toward areas with low probability. The faster the particle moves, the stronger the force.
\end{itemize}

The ``force'' seems to depend on the velocity $\dot{\boldsymbol{\gamma}}(t)$, which is typical of inertial forces (i.e, forces that depend on a given frame). This is an artifact from the affine parametrization of the geodesic, which ensures constant speed along the trajectory.

\paragraph{Newtonian formalism.} Note that the variable $t$ in previous equations is the geometrical ``time''. This variable $t$ stems from the affine parametrization (e.g. see Eq.~\ref{Eq:geodesic_parametrization}) and is not related to the physical time. 
% It can be handy to perform the change of variables $\frac{\partial t}{\partial s}(s)=\frac{1}{p(x(s))}$ to re-formulate the ODE in physical time $s$, closer to Newton equations. With this parametrization, the ``velocity'' is not constant anymore, as often observed in physics. We get $\dot{\boldsymbol{\gamma}}(t)=p(x(s(t)))\frac{\partial x}{\partial s}(s(t))$, where the parametrization $\gamma$ is understood as function of $t$, and the parametrization $x$ as function of $s$.  
To make Eq.~\ref{eq:Newton} compatible with the "physical" time, denoted $s$, one can consider the following change of variable:
\begin{equation}
\frac{\partial s}{\partial t}(t) = p(\boldsymbol{\gamma}(t)) \textnormal{\,\,\, or equivalently \,\,\,} \frac{\partial t}{\partial s}(s) = \frac{1}{p(\boldsymbol{\gamma}(t(s)))}
\label{app:physical_time}
\end{equation}
This change of variable implies that when moving through space according to arc-length s, the geometric time $t$ runs more slowly in low-density regions and faster in high-density ones. This change of variable is particularly handy to interpret Eq.~\ref{eq:Newton} as Newtonian motion. Let's therefore consider the following reparametrization: $\boldsymbol{\gamma}\big(t(s)\big) = \v{x}(s)$, where $\v{x}$ is the new trajectory parametrized by the physical time $s$. So:
\begin{align}
\dot{\boldsymbol{\gamma}}(t) 
&= \frac{\partial}{\partial t} \boldsymbol{\gamma}(t) 
= \frac{\partial}{\partial t} \mathbf{x}(s(t)) 
= \frac{\partial \mathbf{x}}{\partial s} \cdot \frac{\partial s}{\partial t} 
= \dot{\mathbf{x}}(s) \cdot p(\mathbf{x}(s))
\label{eq:dotgamma} \\
\ddot{\boldsymbol{\gamma}}(t) 
&= \frac{\partial}{\partial t} \left( \dot{\mathbf{x}}(s) \cdot p(\mathbf{x}(s)) \right) 
= \left( \frac{\partial}{\partial s} \left( \dot{\mathbf{x}}(s) \cdot p(\mathbf{x}(s)) \right) \right) \cdot \frac{\partial s}{\partial t} \nonumber \\
&= \bigg( \ddot{\mathbf{x}}(s) \cdot p(\mathbf{x}(s)) + \left\langle \nabla p(\mathbf{x}(s)), \dot{\mathbf{x}}(s) \right\rangle \cdot \dot{\mathbf{x}}(s) \bigg) \cdot p(\mathbf{x}(s)) \nonumber \\
&= p(\mathbf{x})^2 \ddot{\mathbf{x}} + p(\mathbf{x}) \left\langle \nabla p(\mathbf{x}), \dot{\mathbf{x}} \right\rangle \dot{\mathbf{x}}
\label{eq:ddotgamma}
\end{align}

Now plugging Eq.~\ref{eq:ddotgamma} and Eq.~\ref{eq:dotgamma} in Eq.~\ref{eq:Newton}:
\begin{equation}
\begin{aligned}
p(\mathbf{x})^2 \ddot{\mathbf{x}} 
&+ p(\mathbf{x}) \left\langle \nabla p(\mathbf{x}), \dot{\mathbf{x}} \right\rangle \dot{\mathbf{x}} \nonumber\\
&= \left\langle \nabla \log p(\mathbf{x}), \dot{\boldsymbol{\gamma}}(t) \right\rangle \cdot \dot{\boldsymbol{\gamma}}(t)
- \frac{1}{2} \left\| \dot{\boldsymbol{\gamma}}(t) \right\|^2 \cdot \nabla \log p(\mathbf{x}) \nonumber\\
&= \left\langle \nabla \log p(\mathbf{x}), p(\mathbf{x}) \dot{\mathbf{x}} \right\rangle \cdot \left( p(\mathbf{x}) \dot{\mathbf{x}} \right)
- \frac{1}{2} \left\| p(\mathbf{x}) \dot{\mathbf{x}} \right\|^2 \cdot \nabla \log p(\mathbf{x}) \nonumber\\
&= p(\mathbf{x})^2 \left\langle \nabla \log p(\mathbf{x}), \dot{\mathbf{x}} \right\rangle \dot{\mathbf{x}}
- \frac{1}{2} p(\mathbf{x})^2 \|\dot{\mathbf{x}}\|^2 \cdot \nabla \log p(\mathbf{x}) \nonumber\\
\Rightarrow \quad \ddot{\mathbf{x}} 
&= -\frac{1}{2} \|\dot{\mathbf{x}}\|^2 
\underbrace{\nabla \log p(\mathbf{x})}_{\text{Stein score}}.
\label{eq:final_newton}
\end{aligned}
\end{equation}

This equation can be interpreted through Newton's second law: it describes the motion of a particle $\v{x}$ following a geodesic in the Riemanannian manifold $\big(\mathcal{M}, \frac{1}{p(\v{x})}\big)$, where $p(\v{x})$ denotes the data density. The particle experiences a force $-\frac{1}{2} \|\dot{\mathbf{x}}\|^2 
\nabla \log p(\mathbf{x})$, pushing away from regions of high probability. The term $||\v{x}||^2$ modulates the forces magnitude and plays a role analogous to momentum, strengthening the pull when the particle moves quickly. While this is not a literal physical system---here the particle is a data point, and has no mass, it provides a useful analogy for understanding the dynamics of trajectories shaped by data geometry.

\newpage
\section{Limitations}\label{app:limitations}
While our approach provides a promising framework for deriving Riemannian metrics from EBMs, several limitations should be acknowledged: 
\begin{itemize}
    \item First, we restrict our study to conformal metrics, which uniformly scale the identity matrix and thus cannot capture directional (anisotropic) structure in the data manifold. While this simplifies optimization, it limits expressivity in settings where geometry varies across directions—something more expressive, score-based metrics may help resolve.
    \item Second, our method relies on pretrained EBMs that assign meaningful energy values across the entire space. Training such models is challenging in high-dimensional settings due to the computational cost of sampling (e.g., Langevin dynamics), and performance can degrade if the energy landscape is poorly shaped or overfitted.
    \item Third, although we demonstrate improvements over strong baselines, our evaluation of geodesic quality remains largely indirect—relying on alignment with proxy measures (e.g., density, rotation smoothness, FID). In complex datasets like natural images, the absence of ground-truth geometry makes rigorous evaluation difficult.
    \item Fourth, our approach assumes that the data distribution is adequately captured by the EBM, yet in practice, misestimation of density—especially in underrepresented regions—may distort the metric and lead to suboptimal paths.
    \item Finally, while we demonstrate promising results on several datasets, our experiments are constrained to pretrained generative models and fixed feature spaces (e.g., VAE latents), and generalizing to end-to-end learnable architectures remains unexplored.
\end{itemize}
  Future work may address these limitations by developing scalable score-based metrics, improving EBM training stability, integrating richer evaluation protocols, and extending the framework to broader model classes and learning settings.

\section{Broader Impact}\label{app:broader_impact}
This work advances our understanding of data geometry by connecting generative modeling and Riemannian geometry, with potential implications across machine learning, neuroscience, and cognitive science. By enabling principled geodesic computation in high-dimensional spaces, our approach could support safer interpolation in generative models, improve motion planning in robotics, or inform models of human cognition. However, care should be taken when applying such methods to sensitive domains, as learned energy landscapes may inherit biases present in training data.

\section{Computational ressources}
\label{app:computational_resources}
All experiments were conducted on NVIDIA RTX 3090 GPUs (32 GB memory). Training on the toy dataset was fast, with both the EBM and interpolant completing in a few minutes. For the Rotated Characters dataset, EBM training took 8 GPU hours and the interpolant 30 minutes. On the AFHQ dataset, training required 6 GPU days for the EBM and 24 GPU hours for the interpolant. Including extensive hyperparameter searches and trial-and-error development, the total compute usage amounted to approximately 123,000 GPU hours.

\newpage
\section*{NeurIPS Paper Checklist}

\begin{enumerate}

\item {\bf Claims}
    \item[] Question: Do the main claims made in the abstract and introduction accurately reflect the paper's contributions and scope?
    \item[] Answer: \answerYes{}%, \answerNo{}, or \answerNA{}.
    \item[] Justification: Our main claim is that EBM-derived metrics stay closer to the data manifold and better capture the geometry of the data compared to alternative metrics. In all experimental settings this claim is verified (see Fig.~\ref{fig:fig1}, Fig.~\ref{fig:fig3}, Table~\ref{table:FID_AFHQ}).
    \item[] Guidelines:
    \begin{itemize}
        \item The answer NA means that the abstract and introduction do not include the claims made in the paper.
        \item The abstract and/or introduction should clearly state the claims made, including the contributions made in the paper and important assumptions and limitations. A No or NA answer to this question will not be perceived well by the reviewers. 
        \item The claims made should match theoretical and experimental results, and reflect how much the results can be expected to generalize to other settings. 
        \item It is fine to include aspirational goals as motivation as long as it is clear that these goals are not attained by the paper. 
    \end{itemize}

\item {\bf Limitations}
    \item[] Question: Does the paper discuss the limitations of the work performed by the authors?
    \item[] Answer: \answerYes{}%, \answerNo{}, or \answerNA{}.
    \item[] Justification: We have briefly discussed limitations in the conclusion section of the main article. But we have included an addtional a full section in the supplementary information (see Supp.~\ref{app:limitations}) to expand those limitations.
    \item[] Guidelines:
    \begin{itemize}
        \item The answer NA means that the paper has no limitation while the answer No means that the paper has limitations, but those are not discussed in the paper. 
        \item The authors are encouraged to create a separate "Limitations" section in their paper.
        \item The paper should point out any strong assumptions and how robust the results are to violations of these assumptions (e.g., independence assumptions, noiseless settings, model well-specification, asymptotic approximations only holding locally). The authors should reflect on how these assumptions might be violated in practice and what the implications would be.
        \item The authors should reflect on the scope of the claims made, e.g., if the approach was only tested on a few datasets or with a few runs. In general, empirical results often depend on implicit assumptions, which should be articulated.
        \item The authors should reflect on the factors that influence the performance of the approach. For example, a facial recognition algorithm may perform poorly when image resolution is low or images are taken in low lighting. Or a speech-to-text system might not be used reliably to provide closed captions for online lectures because it fails to handle technical jargon.
        \item The authors should discuss the computational efficiency of the proposed algorithms and how they scale with dataset size.
        \item If applicable, the authors should discuss possible limitations of their approach to address problems of privacy and fairness.
        \item While the authors might fear that complete honesty about limitations might be used by reviewers as grounds for rejection, a worse outcome might be that reviewers discover limitations that aren't acknowledged in the paper. The authors should use their best judgment and recognize that individual actions in favor of transparency play an important role in developing norms that preserve the integrity of the community. Reviewers will be specifically instructed to not penalize honesty concerning limitations.
    \end{itemize}

\item {\bf Theory Assumptions and Proofs}
    \item[] Question: For each theoretical result, does the paper provide the full set of assumptions and a complete (and correct) proof?
    \item[] Answer: \answerNA{}.
    \item[] Justification: This paper does not present new theoretical results, but it builds on and leverages existing theoretical insights from prior work.
    \item[] Guidelines:
    \begin{itemize}
        \item The answer NA means that the paper does not include theoretical results. 
        \item All the theorems, formulas, and proofs in the paper should be numbered and cross-referenced.
        \item All assumptions should be clearly stated or referenced in the statement of any theorems.
        \item The proofs can either appear in the main paper or the supplemental material, but if they appear in the supplemental material, the authors are encouraged to provide a short proof sketch to provide intuition. 
        \item Inversely, any informal proof provided in the core of the paper should be complemented by formal proofs provided in appendix or supplemental material.
        \item Theorems and Lemmas that the proof relies upon should be properly referenced. 
    \end{itemize}

    \item {\bf Experimental Result Reproducibility}
    \item[] Question: Does the paper fully disclose all the information needed to reproduce the main experimental results of the paper to the extent that it affects the main claims and/or conclusions of the paper (regardless of whether the code and data are provided or not)?
    \item[] Answer: \answerYes{}%, \answerNo{}, or \answerNA{}.
    \item[] Justification: Due to space constraints, we could not include all experimental details in the main paper. However, the supplementary material provides a thorough description of each experiment, including neural network architectures, all hyperparameters, additional samples, and the pseudo-codes for the main algorithms we used.
    \item[] Guidelines:
    \begin{itemize}
        \item The answer NA means that the paper does not include experiments.
        \item If the paper includes experiments, a No answer to this question will not be perceived well by the reviewers: Making the paper reproducible is important, regardless of whether the code and data are provided or not.
        \item If the contribution is a dataset and/or model, the authors should describe the steps taken to make their results reproducible or verifiable. 
        \item Depending on the contribution, reproducibility can be accomplished in various ways. For example, if the contribution is a novel architecture, describing the architecture fully might suffice, or if the contribution is a specific model and empirical evaluation, it may be necessary to either make it possible for others to replicate the model with the same dataset, or provide access to the model. In general. releasing code and data is often one good way to accomplish this, but reproducibility can also be provided via detailed instructions for how to replicate the results, access to a hosted model (e.g., in the case of a large language model), releasing of a model checkpoint, or other means that are appropriate to the research performed.
        \item While NeurIPS does not require releasing code, the conference does require all submissions to provide some reasonable avenue for reproducibility, which may depend on the nature of the contribution. For example
        \begin{enumerate}
            \item If the contribution is primarily a new algorithm, the paper should make it clear how to reproduce that algorithm.
            \item If the contribution is primarily a new model architecture, the paper should describe the architecture clearly and fully.
            \item If the contribution is a new model (e.g., a large language model), then there should either be a way to access this model for reproducing the results or a way to reproduce the model (e.g., with an open-source dataset or instructions for how to construct the dataset).
            \item We recognize that reproducibility may be tricky in some cases, in which case authors are welcome to describe the particular way they provide for reproducibility. In the case of closed-source models, it may be that access to the model is limited in some way (e.g., to registered users), but it should be possible for other researchers to have some path to reproducing or verifying the results.
        \end{enumerate}
    \end{itemize}

\item {\bf Open access to data and code}
    \item[] Question: Does the paper provide open access to the data and code, with sufficient instructions to faithfully reproduce the main experimental results, as described in supplemental material?
    \item[] Answer: \answerYes{}%, \answerNo{}, or \answerNA{}.
    \item[] Justification: All datasets we used are open access. In addition, upon acceptance we will release the github code to reproduce all the experiments.
    \item[] Guidelines:
    \begin{itemize}
        \item The answer NA means that paper does not include experiments requiring code.
        \item Please see the NeurIPS code and data submission guidelines (\url{https://nips.cc/public/guides/CodeSubmissionPolicy}) for more details.
        \item While we encourage the release of code and data, we understand that this might not be possible, so “No” is an acceptable answer. Papers cannot be rejected simply for not including code, unless this is central to the contribution (e.g., for a new open-source benchmark).
        \item The instructions should contain the exact command and environment needed to run to reproduce the results. See the NeurIPS code and data submission guidelines (\url{https://nips.cc/public/guides/CodeSubmissionPolicy}) for more details.
        \item The authors should provide instructions on data access and preparation, including how to access the raw data, preprocessed data, intermediate data, and generated data, etc.
        \item The authors should provide scripts to reproduce all experimental results for the new proposed method and baselines. If only a subset of experiments are reproducible, they should state which ones are omitted from the script and why.
        \item At submission time, to preserve anonymity, the authors should release anonymized versions (if applicable).
        \item Providing as much information as possible in supplemental material (appended to the paper) is recommended, but including URLs to data and code is permitted.
    \end{itemize}

\item {\bf Experimental Setting/Details}
    \item[] Question: Does the paper specify all the training and test details (e.g., data splits, hyperparameters, how they were chosen, type of optimizer, etc.) necessary to understand the results?
    \item[] Answer: \answerYes{}%, \answerNo{}, or \answerNA{}.
    \item[] In the supplementary materials (see Supp.~\ref{app:circular_dataset}, Supp.~\ref{app:rotated_char} and Supp.~\ref{app:afhq_experiment}), we have extensively reported experimental details about the datasets, the type of optimizers we used, and the hyperparameters.
    \item[] Guidelines:
    \begin{itemize}
        \item The answer NA means that the paper does not include experiments.
        \item The experimental setting should be presented in the core of the paper to a level of detail that is necessary to appreciate the results and make sense of them.
        \item The full details can be provided either with the code, in appendix, or as supplemental material.
    \end{itemize}

\item {\bf Experiment Statistical Significance}
    \item[] Question: Does the paper report error bars suitably and correctly defined or other appropriate information about the statistical significance of the experiments?
    \item[] Answer: \answerYes{}
    \item[] Justification: We reported the 2$\sigma$ error bar for all quantitative metrics on the Supplementary information (see 
    Supp.~\ref{app:toy_2sig}, Supp.~\ref{app:afhq_experiment} and Supp.~\ref{app:FID_error}).
    \item[] Guidelines:
    \begin{itemize}
        \item The answer NA means that the paper does not include experiments.
        \item The authors should answer "Yes" if the results are accompanied by error bars, confidence intervals, or statistical significance tests, at least for the experiments that support the main claims of the paper.
        \item The factors of variability that the error bars are capturing should be clearly stated (for example, train/test split, initialization, random drawing of some parameter, or overall run with given experimental conditions).
        \item The method for calculating the error bars should be explained (closed form formula, call to a library function, bootstrap, etc.)
        \item The assumptions made should be given (e.g., Normally distributed errors).
        \item It should be clear whether the error bar is the standard deviation or the standard error of the mean.
        \item It is OK to report 1-sigma error bars, but one should state it. The authors should preferably report a 2-sigma error bar than state that they have a 96\% CI, if the hypothesis of Normality of errors is not verified.
        \item For asymmetric distributions, the authors should be careful not to show in tables or figures symmetric error bars that would yield results that are out of range (e.g., negative error rates).
        \item If error bars are reported in tables or plots, The authors should explain in the text how they were calculated and reference the corresponding figures or tables in the text.
    \end{itemize}

\item {\bf Experiments Compute Resources}
    \item[] Question: For each experiment, does the paper provide sufficient information on the computer resources (type of computing workers, memory, time of execution) needed to reproduce the experiments?
    \item[] Answer: \answerYes{}
    \item[] Justification: We have included a section describing the computational resources we use for all experiments in the supplementary information (see Supp.~\ref{app:computational_resources}).
    \item[] Guidelines:
    \begin{itemize}
        \item The answer NA means that the paper does not include experiments.
        \item The paper should indicate the type of compute workers CPU or GPU, internal cluster, or cloud provider, including relevant memory and storage.
        \item The paper should provide the amount of compute required for each of the individual experimental runs as well as estimate the total compute. 
        \item The paper should disclose whether the full research project required more compute than the experiments reported in the paper (e.g., preliminary or failed experiments that didn't make it into the paper). 
    \end{itemize}
    
\item {\bf Code Of Ethics}
    \item[] Question: Does the research conducted in the paper conform, in every respect, with the NeurIPS Code of Ethics \url{https://neurips.cc/public/EthicsGuidelines}?
    \item[] Answer: \answerYes{}%, \answerNo{}, or \answerNA{}.
    \item[] Justification: All the research conducted in this article conforms to the Neurips Code of Ethics
    \item[] Guidelines:
    \begin{itemize}
        \item The answer NA means that the authors have not reviewed the NeurIPS Code of Ethics.
        \item If the authors answer No, they should explain the special circumstances that require a deviation from the Code of Ethics.
        \item The authors should make sure to preserve anonymity (e.g., if there is a special consideration due to laws or regulations in their jurisdiction).
    \end{itemize}

\item {\bf Broader Impacts}
    \item[] Question: Does the paper discuss both potential positive societal impacts and negative societal impacts of the work performed?
    \item[] Answer: \answerYes{}%, \answerNo{}, or \answerNA{}.
    \item[] Justification: We have discussed the broader impact of our research in Supp.~\ref{app:broader_impact}
    \item[] Guidelines:
    \begin{itemize}
        \item The answer NA means that there is no societal impact of the work performed.
        \item If the authors answer NA or No, they should explain why their work has no societal impact or why the paper does not address societal impact.
        \item Examples of negative societal impacts include potential malicious or unintended uses (e.g., disinformation, generating fake profiles, surveillance), fairness considerations (e.g., deployment of technologies that could make decisions that unfairly impact specific groups), privacy considerations, and security considerations.
        \item The conference expects that many papers will be foundational research and not tied to particular applications, let alone deployments. However, if there is a direct path to any negative applications, the authors should point it out. For example, it is legitimate to point out that an improvement in the quality of generative models could be used to generate deepfakes for disinformation. On the other hand, it is not needed to point out that a generic algorithm for optimizing neural networks could enable people to train models that generate Deepfakes faster.
        \item The authors should consider possible harms that could arise when the technology is being used as intended and functioning correctly, harms that could arise when the technology is being used as intended but gives incorrect results, and harms following from (intentional or unintentional) misuse of the technology.
        \item If there are negative societal impacts, the authors could also discuss possible mitigation strategies (e.g., gated release of models, providing defenses in addition to attacks, mechanisms for monitoring misuse, mechanisms to monitor how a system learns from feedback over time, improving the efficiency and accessibility of ML).
    \end{itemize}
    
\item {\bf Safeguards}
    \item[] Question: Does the paper describe safeguards that have been put in place for responsible release of data or models that have a high risk for misuse (e.g., pretrained language models, image generators, or scraped datasets)?
    \item[] Answer: \answerNA{}.
    \item[] Justification: We don’t think our work poses a significant risk.
    \item[] Guidelines:
    \begin{itemize}
        \item The answer NA means that the paper poses no such risks.
        \item Released models that have a high risk for misuse or dual-use should be released with necessary safeguards to allow for controlled use of the model, for example by requiring that users adhere to usage guidelines or restrictions to access the model or implementing safety filters. 
        \item Datasets that have been scraped from the Internet could pose safety risks. The authors should describe how they avoided releasing unsafe images.
        \item We recognize that providing effective safeguards is challenging, and many papers do not require this, but we encourage authors to take this into account and make a best faith effort.
    \end{itemize}

\item {\bf Licenses for existing assets}
    \item[] Question: Are the creators or original owners of assets (e.g., code, data, models), used in the paper, properly credited and are the license and terms of use explicitly mentioned and properly respected?
    \item[] Answer: \answerYes{}
    \item[] Justification: In terms of datasets, we use the Mixture of Gaussians (not under license), the AFHQ dataset (under CC BY 4.0 Licence), and the rotated letter (based and sklearn letters). In addition, we use the interpolant training algorithms, and the contrastive divergence to train EBMS. All the creators of these assets have been credited by citing the corresponding articles.
    \item[] Guidelines:
    \begin{itemize}
        \item The answer NA means that the paper does not use existing assets.
        \item The authors should cite the original paper that produced the code package or dataset.
        \item The authors should state which version of the asset is used and, if possible, include a URL.
        \item The name of the license (e.g., CC-BY 4.0) should be included for each asset.
        \item For scraped data from a particular source (e.g., website), the copyright and terms of service of that source should be provided.
        \item If assets are released, the license, copyright information, and terms of use in the package should be provided. For popular datasets, \url{paperswithcode.com/datasets} has curated licenses for some datasets. Their licensing guide can help determine the license of a dataset.
        \item For existing datasets that are re-packaged, both the original license and the license of the derived asset (if it has changed) should be provided.
        \item If this information is not available online, the authors are encouraged to reach out to the asset's creators.
    \end{itemize}

\item {\bf New Assets}
    \item[] Question: Are new assets introduced in the paper well documented and is the documentation provided alongside the assets?
    \item[] Answer: \answerYes{}.
    \item[] Justification: Our only new asset is the code used to run all experiments, which will be released publicly under the MIT license upon acceptance. All other assets are fully documented in this article.
    \item[] Guidelines:
    \begin{itemize}
        \item The answer NA means that the paper does not release new assets.
        \item Researchers should communicate the details of the dataset/code/model as part of their submissions via structured templates. This includes details about training, license, limitations, etc. 
        \item The paper should discuss whether and how consent was obtained from people whose asset is used.
        \item At submission time, remember to anonymize your assets (if applicable). You can either create an anonymized URL or include an anonymized zip file.
    \end{itemize}

\item {\bf Crowdsourcing and Research with Human Subjects}
    \item[] Question: For crowdsourcing experiments and research with human subjects, does the paper include the full text of instructions given to participants and screenshots, if applicable, as well as details about compensation (if any)? 
    \item[] Answer: \answerNA{}. % Replace by \answerYes{}, \answerNo{}, or \answerNA{}.
    \item[] Justification: No human experiments or crowdsourcing are involved in this article.
    \item[] Guidelines:
    \begin{itemize}
        \item The answer NA means that the paper does not involve crowdsourcing nor research with human subjects.
        \item Including this information in the supplemental material is fine, but if the main contribution of the paper involves human subjects, then as much detail as possible should be included in the main paper. 
        \item According to the NeurIPS Code of Ethics, workers involved in data collection, curation, or other labor should be paid at least the minimum wage in the country of the data collector. 
    \end{itemize}

\item {\bf Institutional Review Board (IRB) Approvals or Equivalent for Research with Human Subjects}
    \item[] Question: Does the paper describe potential risks incurred by study participants, whether such risks were disclosed to the subjects, and whether Institutional Review Board (IRB) approvals (or an equivalent approval/review based on the requirements of your country or institution) were obtained?
    \item[] Answer: \answerNA{}.
    \item[] Justification: No human experiments or crowdsourcing are involved in this article.
    \item[] Guidelines:
    \begin{itemize}
        \item The answer NA means that the paper does not involve crowdsourcing nor research with human subjects.
        \item Depending on the country in which research is conducted, IRB approval (or equivalent) may be required for any human subjects research. If you obtained IRB approval, you should clearly state this in the paper. 
        \item We recognize that the procedures for this may vary significantly between institutions and locations, and we expect authors to adhere to the NeurIPS Code of Ethics and the guidelines for their institution. 
        \item For initial submissions, do not include any information that would break anonymity (if applicable), such as the institution conducting the review.
    \end{itemize}

\end{enumerate}

\end{document}